\documentclass[10pt,twocolumn,letterpaper]{article}

\usepackage{cvpr}              

\usepackage[dvipsnames]{xcolor}

\usepackage{mathtools}
\usepackage{amsthm}

\usepackage{multirow}
\usepackage{makecell}
\usepackage{xcolor}

\usepackage{algorithm}
\usepackage{booktabs}

\usepackage{makecell}

\usepackage[font=small,aboveskip=1pt,belowskip=-8pt]{caption} 
\usepackage{xcolor}
\usepackage{adjustbox}
\usepackage{tabularray}
\UseTblrLibrary{booktabs}
\UseTblrLibrary{diagbox}

\usepackage[accsupp]{axessibility} 

\usepackage[frozencache=true,cachedir=cache-minted]{minted} 

\definecolor{codebg}{HTML}{F5F5F5}

\definecolor{cvprblue}{rgb}{0.21,0.49,0.74}
\usepackage[pagebackref,breaklinks,colorlinks,citecolor=cvprblue]{hyperref}

\usepackage[capitalize]{cleveref}
\crefname{section}{Sec.}{Secs.}
\Crefname{section}{Section}{Sections}
\Crefname{table}{Table}{Tables}
\crefname{table}{Tab.}{Tabs.}

\newcommand{\specialcelll}[2][l]{
  \begin{tabular}[#1]{@{}l@{}}#2\end{tabular}}
\newlength\savewidth

\newcommand{\tablestyle}[2]{\setlength{\tabcolsep}{#1}\renewcommand{\arraystretch}{#2}\centering\footnotesize}
\makeatletter\renewcommand\paragraph{\@startsection{paragraph}{4}{\z@}
  {.5em \@plus1ex \@minus.2ex}{-.5em}{\normalfont\normalsize\bfseries}}\makeatother
\newdimen\abovecrulesep
\newdimen\belowcrulesep
\makeatletter
\patchcmd{\@@@cmidrule}{\aboverulesep}{\abovecrulesep}{}{}
\patchcmd{\@@@midrule}{\aboverulesep}{\abovecrulesep}{}{}
\patchcmd{\@xcmidrule}{\belowrulesep}{\belowcrulesep}{}{}
\makeatother

\newcommand{\benchmark}{\textsc{LayoutBench}}

\newcommand{\benchmarkreal}{\textsc{LayoutBench-COCO}}
\newcommand{\clevr}{CLEVR}

\newcommand{\stable}{Stable Diffusion}
\newcommand{\method}{\textsc{IterInpaint}}
\newcommand{\reco}{ReCo}
\newcommand{\ldm}{LDM}

\newcommand{\gligen}{GLIGEN}
\newcommand{\controlnet}{ControlNet}

\def\paperID{6} 
\def\confName{CVPR}
\def\confYear{2024}

\begin{document}

\title{
Diagnostic Benchmark and Iterative Inpainting\\ for Layout-Guided Image Generation
}

\author{
  Jaemin Cho$^1$ \quad
  Linjie Li$^2$ \quad
  Zhengyuan Yang$^2$ \quad
  Zhe Gan$^2$ \quad 
  Lijuan Wang$^2$ \quad
  Mohit Bansal$^1$ \\
  UNC Chapel Hill$^1$ \quad Microsoft Research$^2$\\
  {\tt\small \{jmincho, mbansal\}@cs.unc.edu} \quad 
  {\tt\small \{lindsey.li, zhengyang, zhe.gan, lijuanw\}@microsoft.com} 
  \\  
    {\tt \normalsize \href{https://layoutbench.github.io}{https://layoutbench.github.io}}
}

\twocolumn[{
\renewcommand\twocolumn[1][]{#1}
\maketitle
\begin{center}
    \centering
    \captionsetup{type=figure}
    \vspace{-8mm}      
    \includegraphics[width=0.75\linewidth]{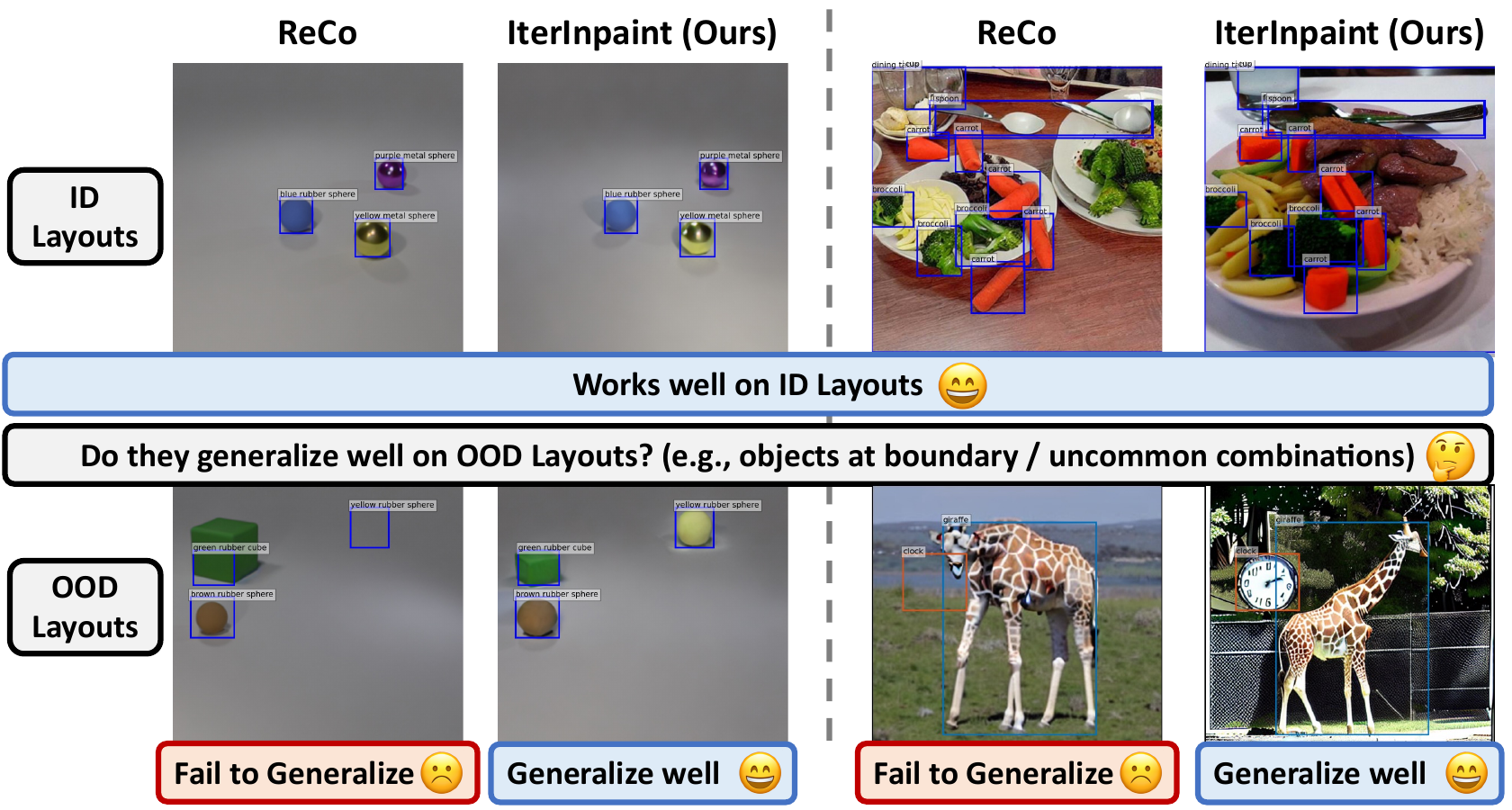}
    \captionof{figure}{
    We propose 
    \textbf{\benchmark{}} (\cref{sec:benchmark}), a diagnostic benchmark for layout-guided image generation models with \textbf{out-of-distribution (OOD)} layouts
    in four skills: \textit{number}, \textit{position}, \textit{size}, and \textit{shape}.
    Existing models such as \reco{}~\cite{Yang2022ReCo} fail on OOD layouts by misplacing objects.
    Next, we introduce \textbf{\method{}} (\cref{sec:method}), a new baseline model with a better generalization on OOD layouts.   
    }
    \vspace{10pt}
    \label{fig:teaser}
\end{center}
}]

\maketitle

\begin{abstract}

Spatial control is a core capability in controllable image generation.
Advancements in layout-guided image generation have shown promising results on in-distribution (ID) datasets with similar spatial configurations.
However, it is unclear how these models perform when facing 
out-of-distribution (OOD) samples with arbitrary, unseen layouts.
In this paper, we propose~\textbf{\benchmark{}}, a diagnostic benchmark for layout-guided image generation
that examines four categories of spatial control skills: number, position, size, and shape.
We benchmark two recent representative layout-guided image generation methods and observe that
the good ID layout control may not generalize well to
arbitrary layouts in the wild (e.g., objects at the boundary).
Next, we propose~\textbf{\method{}}, a new baseline that generates foreground and background regions step-by-step via inpainting, demonstrating stronger generalizability than existing models on OOD layouts in \benchmark{}.
We perform quantitative and qualitative evaluation and fine-grained analysis on the four \benchmark{} skills to pinpoint the weaknesses of existing models.
We show comprehensive ablation studies on \method{}, including training task ratio, crop\&paste vs. repaint,
and generation order.
Lastly, we evaluate the zero-shot performance of different pretrained layout-guided image generation models on \benchmarkreal{}, our new benchmark for OOD layouts with real objects, where our \method{} consistently outperforms SOTA baselines in all four splits.
\end{abstract}

\section{Introduction}
\label{sec:intro}

With the advance of image generation systems that can synthesize diverse and realistic images,
there is an increasing demand for controllable image generation systems that can precisely follow arbitrary spatial configurations defined by users.
For this reason, recent work has focused on the task of layout-to-image generation~\cite{zhao2019image,sun2019image,li2020bachgan,li2021image,frolov2021attrlostgan,yang2022modeling,Rombach_2022_CVPR,fan2022frido}, which aims to generate images conditioned on multiple object bounding boxes and their paired object labels.
Recent layout-guided text-to-image generation models~\cite{Yang2022ReCo,Avrahami2022SpaText,li2023gligen} extend predefined object labels with open-ended regional captions, facilitating the models to generate open-set entities with the queried spatial configurations.
With the recent advances in large-scale image generation models~\cite{Gafni2022MakeAScene,Yu2022Parti,Rombach_2022_CVPR,Ramesh2022DALLE2,Saharia2022Imagen},
newer layout-guided models~\cite{Yang2022ReCo,Avrahami2022SpaText,li2023gligen} have shown promise
in generating high-fidelity images following spatial configurations.

However, most experiments in these previous works are conducted in the in-distribution (ID) setting, where the queried spatial configuration shares a similar layout as the ones in the training samples.
Hence, a natural question arises: how well do these image generation methods perform in real-world scenarios with arbitrary, unseen out-of-distribution (OOD) layouts (\eg, many more or larger/smaller or unusually positioned/shaped regions as compared to the training samples)?
Recent studies~\cite{Yang2022ReCo,li2023gligen}
use qualitative and human evaluations to interpret the model's generation capabilities on arbitrary spatial configurations. 
However, those studies focus on the method development and do not provide systematic benchmarks for spatial control in image generation.
In this study, we aim to develop a benchmark to understand the status quo of image generation with arbitrary spatial configurations and further develop an iterative inpainting-based model to improve the OOD layout generalization. 

To this end, we first propose~\textbf{\benchmark{}} (\cref{sec:benchmark}), a diagnostic benchmark featuring three properties as follows.
\textbf{(1)} We define four categories for spatial control: number, position, size, and shape.
\benchmark{} systematically designs the out-of-distribution (OOD) testing queries for each skill, allowing easy comparison between different spatial configurations.
\textbf{(2)} We evaluate images by layout accuracy in average precision (AP) to reflect the controllable generation quality.
With the release of a well-performing category-balanced object detector that localizes generated objects, \benchmark{} allows fair and easy comparison with prior works.
\textbf{(3)} We choose to develop the benchmark based on the \clevr{} simulator~\cite{johnson2017clevr} to disentangle the factor of image generation quality from the interested spatial controllability.
By simplifying the benchmark with simulated objects, \benchmark{} can better reflect the true spatial control capabilities and avoid blindly favoring large-scale generation models that generate images with better visual qualities but do not understand spatial configurations.

Based on~\benchmark{}, we systematically evaluate two recent representative layout-guided image generation methods:
\ldm{}~\cite{Rombach_2022_CVPR}
and
\reco{}~\cite{Yang2022ReCo}, where both models are initialized by the \stable{} checkpoint.
We perform quantitative and qualitative analyses and fine-grained split analyses on the four \benchmark{} skills to pinpoint the weaknesses of different models.
As depicted in \cref{fig:teaser},
we find that both models fail on OOD layouts of \benchmark{}, while they perform reasonably well on 
ID layouts.

Inspired by the OOD failures of existing models revealed by our \benchmark{} benchmark,
we next propose \textbf{\method{}}, a new baseline for layout-guided image generation (\cref{sec:method}).
Unlike existing methods that condition all the region configurations at a single generation step,
\method{} decomposes image generation into multiple inpainting steps and iteratively updates each region at a time.
By focusing on updating a single region at each time, the model can tackle unseen, complex layouts more robustly than existing methods.
In experiments (\cref{sec:experiments}), \method{} shows significantly better layout accuracy on OOD layouts
and similar or better layout accuracy on ID layouts than prior works.
We also provide comprehensive ablation studies on \method{}, including training task ratio, crop\&paste \vs repaint-based update, and generation order.
Lastly, we evaluate zero-shot performance of different pretrained layout-guided image generation models on \benchmarkreal{}, our new OOD layouts with real objects, where our \method{} outperforms other SOTA models in all four splits.

Our contributions are summarized as follows:
\textbf{(1)} \benchmark{}, a diagnostic benchmark for arbitrary spatial control capabilities of layout-guided image generation models in four criteria: number, position, size, and shape, where existing models often struggle (\cref{sec:benchmark});
\textbf{(2)} \method{}, a new baseline for layout-guided image generation that generates foreground and background in a step-by-step manner, which shows better generalization on OOD layout than prior works (\cref{sec:method});
and 
\textbf{(3)} detailed qualitative/quantitative/sub-split evaluation of spatial control skills of different layout-guided image generation models, comprehensive ablation studies of \method{} design choices,
and zero-shot evaluation of pretrained layout-guided image generation models on \benchmarkreal{}
(\cref{sec:experiments}).

\begin{figure*}[t]
    \centering
    \includegraphics[width=0.9\textwidth]{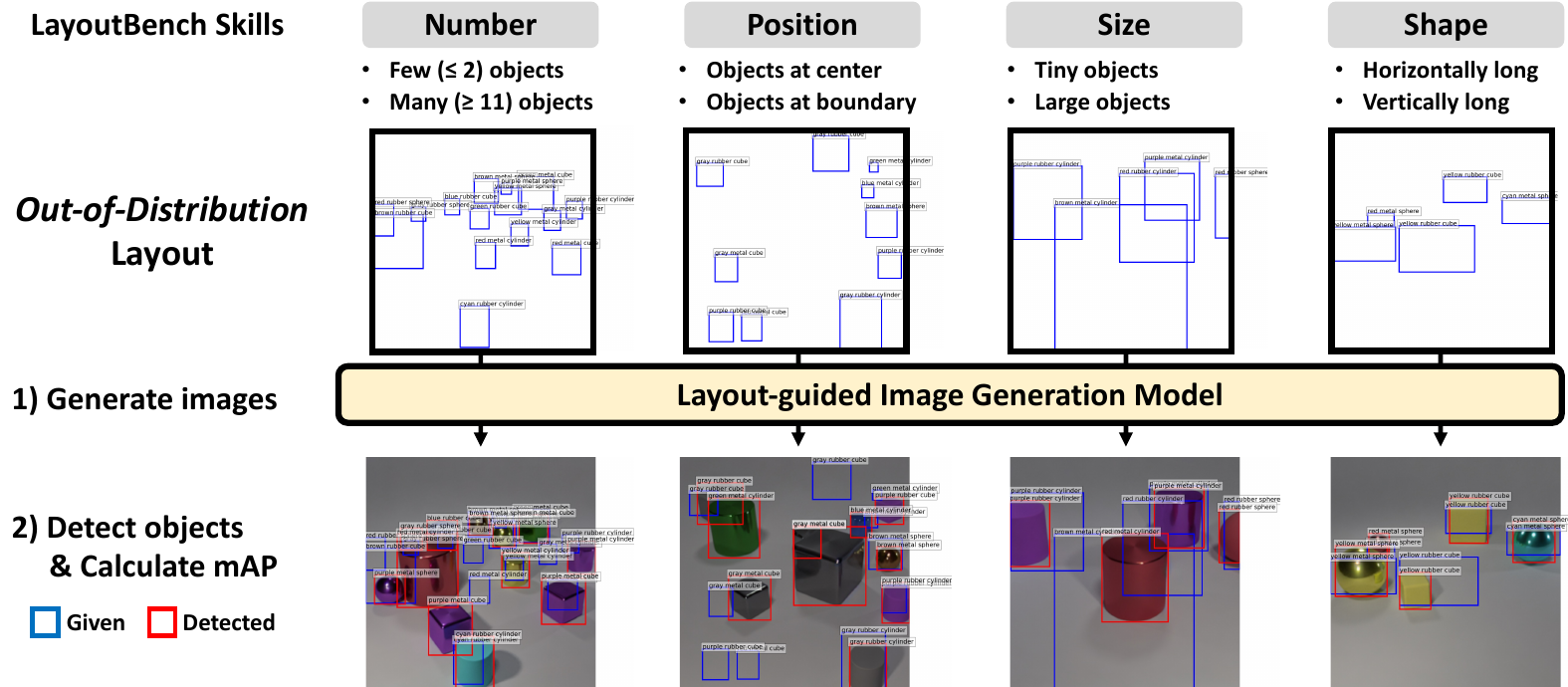}
    \caption{
    In \benchmark{}, we measure 4 spatial control skills (number, position, size, shape) for layout-guided image generation.
    First, 1) we query the image generation models with OOD layouts.
    Then, 2) we detect the objects from the generated images, and calculate the layout accuracy in average precision (AP).
    In each image, the ground-truth boxes are shown in \textcolor{blue}{blue} and the objects detected are shown in \textcolor{red}{red}.
    The images are generated by \reco{}~\cite{Yang2022ReCo} trained on \clevr{}~\cite{johnson2017clevr}, where it often misplaces (\ie, many \textcolor{red}{red} boxes outside of \textcolor{blue}{blue} boxes) or misses objects (\ie, many \textcolor{blue}{blue} boxes are missed) on OOD layouts from \benchmark{}.
    }
    \label{fig:task_overview}
\end{figure*}

\section{Related Work}
\label{sec:related_work}

\paragraph{Text-to-Image Generation Models.} 
In the text-to-image generation task, models generate images from natural language descriptions.
Early deep learning-based models~\cite{Mansimov2016, Reed2016,Zhang2017,Xu2018e} were based on Generative Adversarial Networks (GANs)~\cite{Goodfellow2014}.
Recently, multimodal language models and diffusion models have been widely used for this task.
X-LXMERT \cite{Cho2020} and DALL-E \cite{Ramesh2021} introduce multimodal language models that take text as input and generate discrete image codes iteratively, where a vector-quantized autoencoder learns the mapping between image codes and pixels.
\ldm{}~\cite{Rombach_2022_CVPR} and GLIDE~\cite{Nichol2022} propose text-conditional diffusion models~\cite{Sohl-Dickstein2015, Ho2020} that iteratively update noisy images to clean images.
Recent multimodal language models (\eg, Parti~\cite{Yu2022Parti} and MUSE~\cite{Chang2023Muse}) and diffusion models (\eg, \stable{}~\cite{Rombach_2022_CVPR}, DALL-E 2~\cite{Ramesh2022DALLE2}, and Imagen~\cite{Saharia2022Imagen}) deliver high level of photorealism in zero-shot generation.

\paragraph{Layout-to-Image Generation Models.}
In the layout-to-image generation task, models generate images from layouts (\eg{}, bounding boxes with paired text descriptions).
Early models adopt GAN framework~\cite{Zhao2019,Sun2019e,Sylvain2021}, where an adversarially trained convolutional generator is conditioned on layout input.
Mirroring the success of the text-to-image models, recent layout-to-image models adopt
multimodal language model (\eg{}, Make-A-Scene~\cite{Gafni2022MakeAScene})
and diffusion models (\eg{}, \ldm{}~\cite{Rombach_2022_CVPR}, \reco{}~\cite{Yang2022ReCo}, SpaText~\cite{Avrahami2022SpaText}, GLIGEN~\cite{li2023gligen}, Universal Guided Diffusion~\cite{bansal2023universal}).
While the prior works encode and decode all region inputs in a single step,
\method{} decomposes image generation into multiple steps by focusing on generating one region at one time, showing better generalization on unseen OOD layouts.

\paragraph{Evaluation for Layout-to-Image Generation.}
The layout-to-image generation community has adopted two types of metrics used in text-to-image generation tasks: image quality and image-layout alignment.
For image quality, Inception Score (IS)~\cite{Salimans2016a} and Fréchet Inception Distance (FID)~\cite{Heusel2017} are commonly used.
These metrics use a classifier pretrained on ImageNet~\cite{Deng2009} that mostly contains single-object images.
Therefore, they are not well suited for evaluating images with more complex scenes~\cite{Frolov2021}.
To measure image-layout alignment,
calculating FID on box crops (SceneFID)~\cite{Sylvain2021} and object classification accuracy on box crops~\cite{Zhao2019} have been proposed.
All these metrics summarize the performance of layout-to-image models in a single number, which does not reveal the skills in which the model is good versus the model is bad.
In contrast to the existing metrics which do not pinpoint the model weakness, our \benchmark{} measures four spatial layout control (number, position, size, and shape), to provide a more fine-grained analysis of region control capabilities.

\section{\benchmark{}}
\label{sec:benchmark}

We introduce \textbf{\benchmark{}}, a diagnostic benchmark for layout-guided image generation, with a focus on
four spatial control skills.
In the following,
we discuss
dataset (\cref{sec:dataset}),
layout accuracy (\cref{sec:layout_accuracy}),
and the poor generalizability of existing methods~\cite{Rombach_2022_CVPR,Yang2022ReCo} on \benchmark{}, which motivates us to propose \method{} (\cref{sec:method}).

\begin{table*}[t]
\centering
\tablestyle{2pt}{1.1}      
      \resizebox{\linewidth}{!}{
      \begin{tabular}{c c c c c c c c c}
          \toprule
          ID layout & \multicolumn{8}{c}{OOD layout (\benchmark{})}\\
          \cmidrule(lr){1-1} \cmidrule(lr){2-9}
          \multirow{2}{*}{\clevr{}} & \multicolumn{2}{c} {Skill 1: Number} & \multicolumn{2}{c} {Skill 2: Position} & \multicolumn{2}{c} {Skill 3: Size} & \multicolumn{2}{c} {Skill 4: Shape}\\
          \cmidrule(lr){2-3} \cmidrule(lr){4-5} \cmidrule(lr){6-7} \cmidrule(lr){8-9}
          & few & many & center & boundary & tiny & large & horizontal & vertical \\
          \midrule
          \adjustbox{valign=c}{\includegraphics[height=0.1\textwidth]{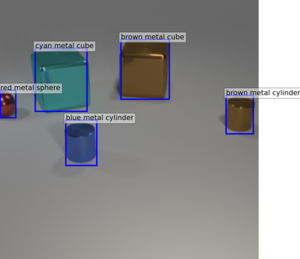}} &
          \adjustbox{valign=c}{\includegraphics[height=0.1\textwidth]{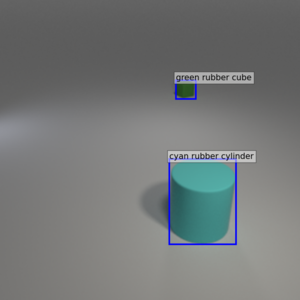}} &
          \adjustbox{valign=c}{\includegraphics[height=0.1\textwidth]{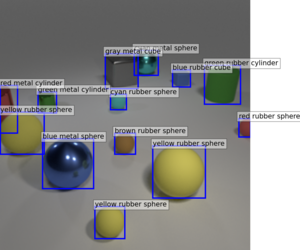}} &
          \adjustbox{valign=c}{\includegraphics[height=0.1\textwidth]{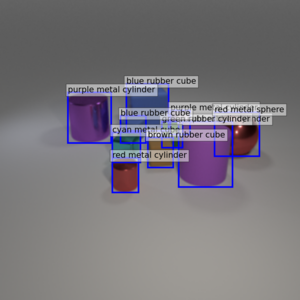}} &
          \adjustbox{valign=c}{\includegraphics[height=0.1\textwidth]{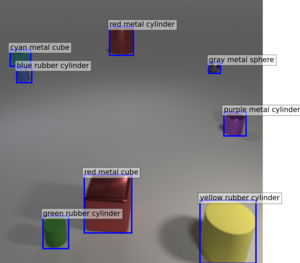}} &
          \adjustbox{valign=c}{\includegraphics[height=0.1\textwidth]{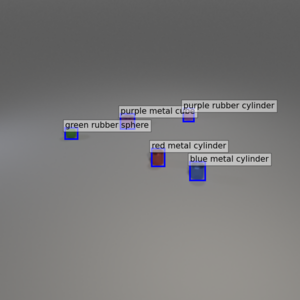}} &
          \adjustbox{valign=c}{\includegraphics[height=0.1\textwidth]{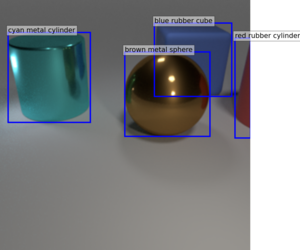}} &
          \adjustbox{valign=c}{\includegraphics[height=0.1\textwidth]{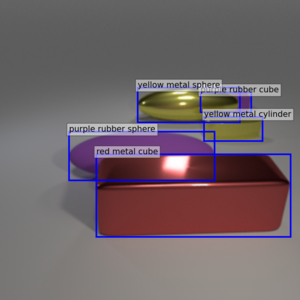}} &
          \adjustbox{valign=c}{\includegraphics[height=0.1\textwidth]{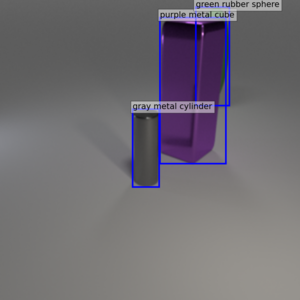}} \\
          \bottomrule
        \end{tabular}
        }
\caption{
Example images with ID (\clevr{}) and OOD (\benchmark{}) layouts. GT boxes are shown in \textcolor{blue}{blue}. 
}
\label{tab:clevr_layoutbench}
\end{table*}

\subsection{Dataset}
\label{sec:dataset}

As illustrated in \cref{fig:task_overview},
\benchmark{} evaluates spatial control capability in 4 skills (number, position, size, shape), where each skill consists of 2 different OOD layout splits, \ie, in total 8 tasks = 4 skills $\times$ 2 splits.
To disentangle the spatial control from other aspects in image generation, such as generating diverse objects,
\benchmark{} keeps the same object configurations as 
\clevr{}~\cite{Johnson2017}, whose objects have 3 shapes, 2 materials, and 8 colors (48 combinations in total).
Images in \benchmark{} are collected in two steps:
(1) sample scenes for each skill, where a scene is defined by the objects and their positions,
(2) render images from the scenes with Blender~\cite{blender} simulator and obtain bounding box layouts.
In total, we collect 8K images for \benchmark{} evaluation, with 1K images per task.
In \cref{tab:clevr_layoutbench}, we show example images with ID and OOD layouts.
We explain the scene configurations below.

\paragraph{In-distribution: \clevr{}.} 
All scenes have 3$\sim$10 objects. These objects are positioned evenly on the canvas, without much occlusion between them.
In terms of size, the rendered objects are in one of two scales $\{3.5, 7\}$.
For shape, the bounding box for each object is an almost perfect square.
For each of the skills below, we only alter the configurations specific to that skill, while keeping the remaining configurations the same as \clevr{}.

\paragraph{Skill 1: Number.}
This skill involves generating images with a specified number of objects.
In contrast to the ID \clevr{} images with 3$\sim$10 objects, we evaluate models on two OOD splits:
(1) \textbf{few}: images with 0$\sim$2 objects;
(2) \textbf{many}: images with 11$\sim$16 objects.

\paragraph{Skill 2: Position.} This skill involves generating images with objects placed at specific positions.
Different from ID \clevr{} images featuring evenly distributed object position without much occlusion between objects,
we design two OOD splits:
(1) \textbf{center}: objects are placed at the center, thus leading to more occlusions;
(2) \textbf{boundary}: objects are only placed on boundaries (top/bottom/left/right). 

\paragraph{Skill 3: Size.}
This skill involves generating images with objects of a specified size.
We construct two OOD splits:
(1) \textbf{tiny}: objects with scale 2;
(2) \textbf{large}: objects with scale $\{9, 11, 13, 15\}$. In comparison, the objects in \clevr{} images have only two scales $\{3.5, 7\}$.
We use 3$\sim$5 objects for this skill, as we find that using more than this number of large objects can often obstruct the object visibilities.

\begin{figure*}[t]
    \centering
    \includegraphics[width=.95\textwidth]{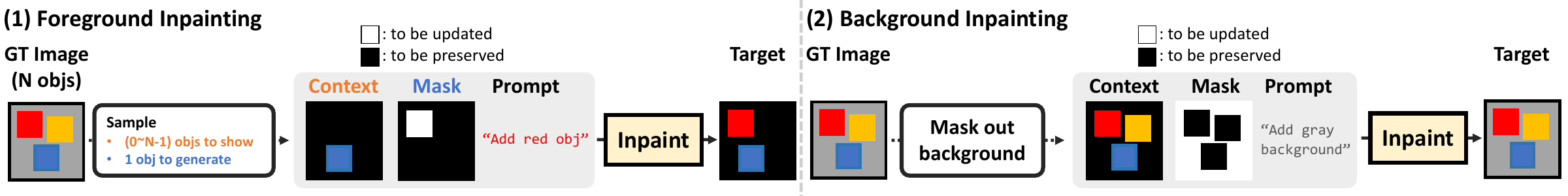}
    \caption{\textbf{\method{} Training.} Our model is trained with (1) foreground and (2) background inpainting tasks  (\cref{sec:iterative_inpainting}).
    }
    \label{fig:training}
\end{figure*}

\paragraph{Skill 4: Shape.}
This skill involves generating images with objects of a specified aspect ratio.
As the objects in \clevr{} images mostly have square aspect ratios,
we evaluate models with two OOD splits:
(1) \textbf{horizontal}: objects in which one of the horizontal (x/y) axes are 2 or 3 times longer than the other axis, leading to object bounding boxes with an aspect ratio (width:height) of 2:1 or 3:1;
(2) \textbf{vertical}: objects whose vertical (z) axis are 2 or 3 times longer than horizontal (x/y) axes, resulting in object bounding boxes with an aspect ratio of 1:2 or 1:3.
We use 3$\sim$5 objects for this skill, as we find that using more than this number of objects can often obstruct the object visibilities.

\subsection{Layout Accuracy Evaluation}
\label{sec:layout_accuracy}

As illustrated in \cref{fig:task_overview},
we evaluate models with four spatial control skills: number, position, size, and shape.
Since existing metrics FID and SceneFID measure how the overall distribution of Inception v3~\cite{Szegedy2016} mean-pool features of generated images/patches is similar to the feature distribution of ground-truth images/patches,
they are less effective in measuring how accurately each generated image follows the input layout~\cite{Frolov2021}. 
Following previous analyses~\cite{Hinz2020,Cho2022DallEval},
we evaluate the skills based on how well an object detector can detect the object described in the input layout.
To better capture the objects with uncommon sizes, positions, and aspect ratios \, etc,
we train DETR~\cite{Carion2020} on separately generated 5K training images for each of 8 tasks, with 40K total images.
We initialize DETR parameters from the official checkpoint with ResNet101~\cite{He2016} backbone pretrained on the COCO~\cite{Lin2014COCO} train 2017 split.
Following object detection literature~\cite{Ren2015,Carion2020,Zhu2020DeformableDETR}, we report average precision (AP).

We evaluate two recent layout-guided image generation models: \ldm{}~\cite{Rombach_2022_CVPR} and \reco{}~\cite{Yang2022ReCo}, trained on \clevr{}.
As shown in \cref{fig:teaser}, they fail on \benchmark{} by ignoring objects, misplacing objects, or placing wrong objects. We closely examine the experiment results in \cref{sec:experiments}.

\begin{figure*}[t]
    \centering
    \begin{minipage}{0.5\textwidth}
    \centerline{\includegraphics[width=\linewidth]{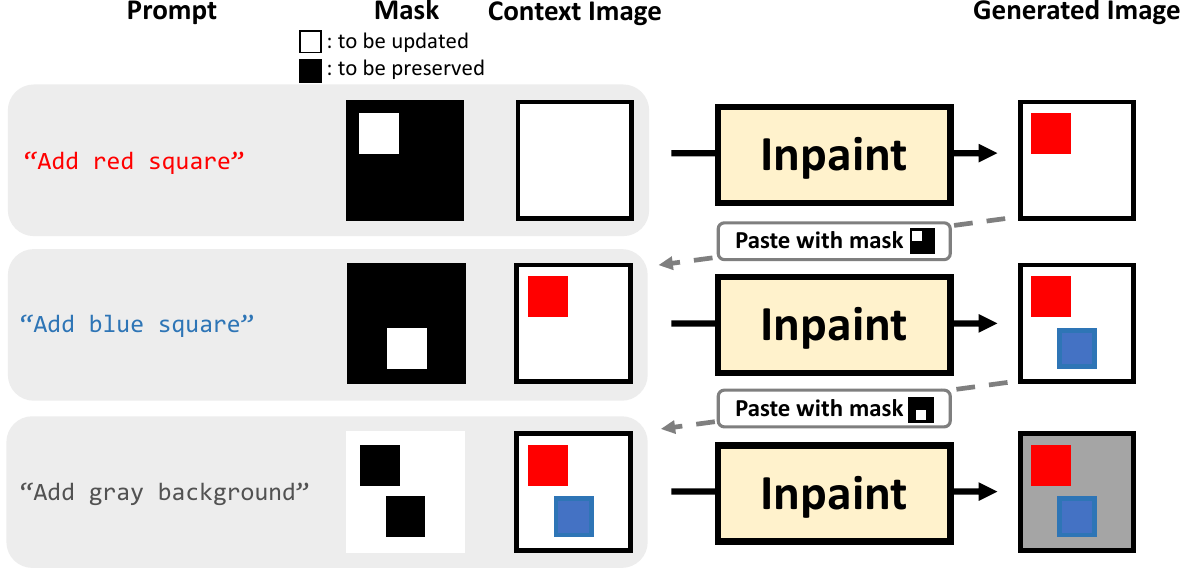}}
    \end{minipage}
    \hfill
    \begin{minipage}{0.45\textwidth}
    \begin{minted}[fontsize=\scriptsize, bgcolor=codebg]{python}
def iterinpaint(caps, boxes):
    # Empty image
    img = Image.new((W,H))
    # Add foreground
    for cap, box in zip(caps, boxes):
        prompt = f"Add {cap}"
        mask = get_obj_mask(box)
        gen = inpaint(img, prompt, mask)
        img = (1-mask) * img + mask * gen
    # Add background
    prompt = "Add gray background"
    mask = get_bg_mask(boxes)
    gen = inpaint(img, prompt, mask)
    return gen
    \end{minted}
    \end{minipage}
    \caption{\textbf{\method{} Inference.}
    Illustration (left) and Python pseudocode (right) of layout-guided image generation with \method{} (\cref{sec:iterative_inpainting}).
    At each iteration, the inpainting model takes the prompt, mask, and previous image as inputs and generates a new image.
    }
    \label{fig:inference}
\end{figure*}

\section{\method{}}
\label{sec:method}

To improve the generalizability of OOD layouts, we propose \method{}, a new layout-guided image generation method based on \textbf{iter}ative \textbf{inpaint}ing.
Unlike previous methods~\cite{Rombach_2022_CVPR,Yang2022ReCo} that generate all objects simultaneously in a single step, \method{} decomposes the image generation process into multiple steps and uses a text-guided inpainting model to update foreground and background regions step-by-step.
In what follows,
we briefly recap \stable{}, which we build \method{} on (\cref{sec:sd}),
describe how we extend the \stable{} for layout-guided inpainting (\cref{sec:extend_to_inpaint}),
and introduce iterative foreground/background inpainting (\cref{sec:iterative_inpainting}).

\subsection{Preliminaries: Stable Diffusion}
\label{sec:sd}

We implement the \method{} model by extending \stable{}, a public text-to-image model based on Latent Diffusion~\cite{Rombach_2022_CVPR}.
\stable{} consists of
(1) a CLIP ViT-L/14~\cite{Radford2021CLIP} text encoder $\texttt{CLIP}_{text} (s)$ that encodes a prompt $s$ into a 512-dimensional vector,
(2) an autoencoder $(E(x), D(z))$ with downsampling factor of 8, which embeds an image $x$ into a 4-dimensional latent space $z_0 \in \mathcal{R}^{(4,H,W)}$, and 
(3) a diffusion U-Net $\epsilon_\theta$ that performs denoising steps in the latent space given timestamp $t$ and CLIP text encoding (conditioned via cross-attention).
The model is trained with the following objective:
$L_{LDM} = \mathbb{E}_{s, z_0, t, \epsilon} [ || \epsilon - \epsilon_\theta (z_{t}, t, \texttt{CLIP}_{text} (s))  ||^2_2 ]$.

\subsection{Extending \stable{} for Inpainting}
\label{sec:extend_to_inpaint}

Our \method{} method decomposes complex scene generation into multiple steps, where each step is a text-guided inpainting~\cite{Zhang2020TextInpaint} process.
Concretely, a model completes an image region, given a context image, a binary mask indicating the region, and a text description of the region.
To enable inpainting, we extend the U-Net of \stable{} to take the mask and a context image as additional inputs.
We use a binary mask of the same size as the image, indicating the region to be updated (1: to be updated; 0: to be preserved).
To encode the mask and context image, we add 5 additional channels to the U-Net's first convolutional layer,
where the first dimension is used to encode the spatially downsampled mask $m \in [0,1]^{(H,W)}$,
and the remaining 4 dimensions are for encoding the latent vector of the context image $z^{ctx}_0 = E(x^{ctx}) \in \mathcal{R}^{(4,H,W)}$.
The resulting layout-guided inpainting model takes
a context image $x^{ctx}$,
a text prompt $s$,
and a binary mask $m$,
as input
and generates an image $x^{gen} = \texttt{inpaint}(x^{ctx}, s, m)$.
Next, we describe the training and inference process of iterative inpainting.

\subsection{Iterative Inpainting}
\label{sec:iterative_inpainting}

\method{} decomposes the image generation process into two phases:
(1) step-by-step generation of each bounding box/mask (foreground), and (2) filling the rest of the images (background).
This decomposition would make each generation step easier by allowing the model to focus on generating a single foreground object or background.

During training, as shown in \cref{fig:training}, we use a single objective to cover both foreground/background 
inpainting by giving the model a different context image and mask:
(1) foreground inpainting - we sample context objects (from N GT objects) to show, then sample an object to generate;
(2) background inpainting - we mask out all objects, and generate the background.
We explore different ratios to sample the two training tasks in \cref{sec:ablation}, and find a 30\% and 70\% ratio for foreground and background inpainting tasks gives the best performance.
We train our model with the modified latent diffusion objective~\cite{Rombach_2022_CVPR},
$L_{IterInpaint} = \mathbb{E}_{s, z_0, t, \epsilon} [ || \epsilon - \epsilon_\theta (z_{t}, t, \texttt{CLIP}_{text} (s), m, z_0^{ctx})  ||^2_2 ]$, where U-Net is additionally conditioned on the mask $m$ and the previous image $z_0^{ctx}$.

\begin{table*}
\centering
\tablestyle{5.5pt}{1.0}

    \begin{tabular}{l c cc cc cc cc c}
        \toprule
        \multirow{3}{*}{Method} & {\clevr{}} & \multicolumn{9}{c}{\benchmark{}} \\
        \cmidrule(lr){2-2} \cmidrule(lr){3-11} 
        
        & \multirow{2}{*}{val} & \multicolumn{2}{c}{Number} & \multicolumn{2}{c}{Position}& \multicolumn{2}{c}{Size} & \multicolumn{2}{c}{Shape} & \multirow{2}{*}{Avg}\\
        
        \cmidrule(lr){3-4} \cmidrule(lr){5-6} \cmidrule(lr){7-8} \cmidrule(lr){9-10} 
        
        & & few & many  & center & boundary & tiny & large  & horizontal & vertical  \\
        \midrule

        GT (Oracle) & 60.5/93.5	& 94.3/99.7 & 92.0/99.0 & 90.9/90.9 & 90.8/99.4 & 82.4/100.0 & 96.6/99.4 & 89.7/99.0 & 89.0/98.4 & 90.7/98.2 \\
        GT shuffled & 0.0/0.0 & 0.1/0.1 & 0.0/0.0 & 0.0/0.0 & 0.0/0.0 & 0.0/0.0 & 0.0/0.0 & 0.0/0.0 & 0.0/0.0 & 0.0/0.0 \\
        \midrule
        \ldm{} & 54.5/\textbf{91.8} & 14.0/48.7 & 4.7/20.7 & 5.5/28.0 & 5.9/15.1 & 0.0/0.0 & 37.8/68.2 & 2.0/12.8 & 9.3/38.5 & 9.9/29.0 \\
        \reco{} & 44.0/89.0 & 8.5/36.9 & 2.5/12.7 & 2.8/17.4 & 2.5/8.7 & 0.0/0.0 & 32.4/70.5 & 3.0/\textbf{19.3} & 8.7/37.8 & 7.6/25.4 \\
        
        \method{} (Ours) & \textbf{57.2}/90.8 & \textbf{50.4}/\textbf{80.5} & \textbf{52.4}/\textbf{87.7} & \textbf{49.6}/\textbf{83.8} & \textbf{50.1}/\textbf{82.1} & \textbf{2.4}/\textbf{7.9} & \textbf{63.1}/\textbf{92.6} & \textbf{4.7}/18.5 & \textbf{19.3}/\textbf{60.1} & \textbf{36.5}/\textbf{64.1} \\

        \bottomrule
    \end{tabular}

    \caption{
    Layout accuracy in AP/AP$_{50}$ (\%) on \clevr{} and \benchmark{}. Best (highest) values are \textbf{bolded}.}
    \label{tab:overall_results_map}
\end{table*}

\begin{table}[t]
\centering
    \resizebox{.8\linewidth}{!}{
    \begin{tabular}{l c c}
        \toprule
        \multirow{1}{*}{Method} & {\clevr{}} & {\benchmark{}} \\

        \midrule

        \ldm{} & 3.4/\textbf{13.0} & 31.1/57.9 \\
        \reco{} & \textbf{2.8}/13.6 & \textbf{30.4}/58.2 \\
        \method{} (Ours) & 12.7/36.3 & 31.4/\textbf{49.0} \\

        \bottomrule
    \end{tabular}
    }
    \caption{Image quality in FID/SceneFID on \clevr{} and \benchmark{}.
    Best (lowest) values are in \textbf{bolded}.
    }
    \label{tab:overall_results_fid}
\end{table}

During inference, as shown in \cref{fig:inference}, we iteratively update foreground objects and background,
starting from a blank image.
For each step, we update the image by composing context image $x^{ctx}$ and the generated image $x^{gen}$ using a mask $m$: $x^{new} = (1-m) * x^{gen} + m * x^{ctx}$.
For a layout with $N$ objects, our \method{} method
generates the final image with $N+1$ (foreground + background) iterations.
Overall, \method{} allows users to control the generation order of each region and interactively manipulate the image from an intermediate generation step.

\section{Experiments and Analysis}
\label{sec:experiments}

\subsection{Experimental Setup}
\label{sec:exp_setup}

In addition to our \method{}, we evaluate two recent and strong layout-guided image generation models, \ldm{}~\cite{Rombach_2022_CVPR}, and \reco{}~\cite{Yang2022ReCo}.
To focus on layout control evaluation, we match the implementation details of three models.

\paragraph{Dataset details.}
We train models on the 70K training images in \clevr{}~\cite{johnson2017clevr}.
As the original \clevr{} dataset does not provide the bounding box annotations, we use the bounding box annotations provided by~\cite{Krishna2018ReferringRelationships}.
The original images have 480x360 (WxH) sizes. For training, we resize the images into 768x512 and center crop to 512x512.

\paragraph{Model details.}
We initialize all model parameters with \stable{} v1 checkpoints. 
We train all models for 20K steps with batch size 128 (\ie, single batch at each of 16 V100 GPUs with 8 gradient accumulation steps),
and AdamW optimizer~\cite{Loshchilov2019AdamW} with constant learning rate 1e-4.
Following~\cite{Yang2022ReCo}, we update U-Net and CLIP text encoder parameters, while freezing the autoencoder. During inference, we use classifier-free guidance~\cite{Ho2022CFG} scale of 4.0 and 50 PLMS~\cite{Liu2022PLMS} steps.

\paragraph{Bounding box encoding.}
For \ldm{} and \reco{}, we quantize each of the bounding box coordinates ($x_1, y_1, x_2, y_2$) into 1000 quantized bins.
For \ldm{}, we learn 48 class embeddings for \clevr{} objects.
We describe the layout by concatenating the list of object class tokens and quantized bounding boxes (\eg, ``\texttt{<020> <230> <492> <478> <cls23> <121> $\cdots$}'') and encode it with CLIP text encoder.
Unlike \ldm{}, \reco{} takes the text description for each region instead of class embedding (\eg, ``\texttt{<020> <230> <492> <478> cyan metal sphere <121> $\cdots$}'') as input.

\paragraph{Evaluation metrics.}
For quantitative evaluation, we measure layout accuracy and image quality.
Layout accuracy is measured by AP (average precision) based on DETR-R101-DC5~\cite{Carion2020}, as mentioned in \cref{sec:layout_accuracy}. Higher AP indicates that the generated images follow the input layouts more closely.
FID~\cite{Heusel2017} and SceneFID~\cite{Sylvain2021} are adopted to measure image quality.
Lower FID (SceneFID) indicates that the generated images (boxes) have a more similar feature distribution to the ground-truth ones.

\begin{table*}[t]
\centering
    \resizebox{\linewidth}{!}{
    
      \begin{tabular}{l c c c c c c c c c}

          \toprule
    
           \multirow{3}{*}{{ Method}} & \clevr{} & \multicolumn{8}{c} {\benchmark{}} \\
          \cmidrule(lr){2-2} \cmidrule(lr){3-10}
    
           &\multirow{2}{*}{{val}} &  \multicolumn{2}{c}{Number}  & \multicolumn{2}{c}{Position}  & \multicolumn{2}{c}{Size} &  \multicolumn{2}{c}{Shape} \\
          \cmidrule(lr){3-4} \cmidrule(lr){5-6} \cmidrule(lr){7-8} \cmidrule(lr){9-10}
          & & few & many & center & boundary & tiny & large & horizontal & vertical \\
          \midrule
          GT &
          \adjustbox{valign=c}{\includegraphics[height=0.1\textwidth]{images/images_for_main_example_table/clevr/gt_512x512.png}} &
          \adjustbox{valign=c}{\includegraphics[height=0.1\textwidth]{images/images_for_main_example_table/number_few/gt_512x512.png}} &
          \adjustbox{valign=c}{\includegraphics[height=0.1\textwidth]{images/images_for_main_example_table/number_many/gt_512x512.png}} &
          \adjustbox{valign=c}{\includegraphics[height=0.1\textwidth]{images/images_for_main_example_table/position_center/gt_512x512.png}} &
          \adjustbox{valign=c}{\includegraphics[height=0.1\textwidth]{images/images_for_main_example_table/position_boundary/gt_512x512.png}} &
          \adjustbox{valign=c}{\includegraphics[height=0.1\textwidth]{images/images_for_main_example_table/size_tiny/gt_512x512.png}} &
          \adjustbox{valign=c}{\includegraphics[height=0.1\textwidth]{images/images_for_main_example_table/size_large/gt_512x512.png}} &
          \adjustbox{valign=c}{\includegraphics[height=0.1\textwidth]{images/images_for_main_example_table/shape_horizontal/gt_512x512.png}} &
          \adjustbox{valign=c}{\includegraphics[height=0.1\textwidth]{images/images_for_main_example_table/shape_vertical/gt_512x512.png}} \\
          \midrule
          \ldm{} &
          \adjustbox{valign=c}{\includegraphics[height=0.1\textwidth]{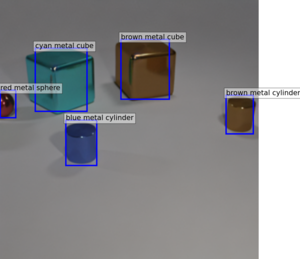}} &
          \adjustbox{valign=c}{\includegraphics[height=0.1\textwidth]{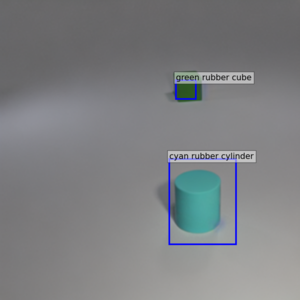}} &
          \adjustbox{valign=c}{\includegraphics[height=0.1\textwidth]{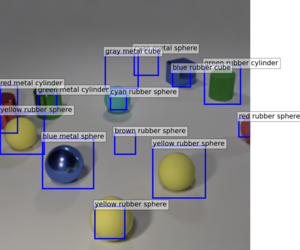}} &
          \adjustbox{valign=c}{\includegraphics[height=0.1\textwidth]{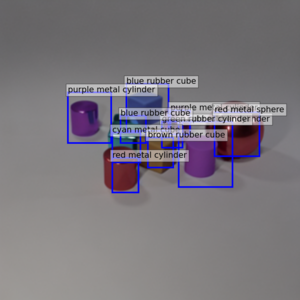}} &
          \adjustbox{valign=c}{\includegraphics[height=0.1\textwidth]{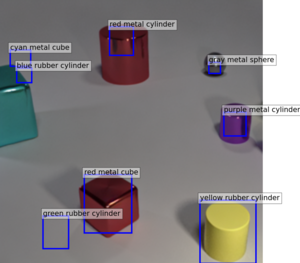}} &
          \adjustbox{valign=c}{\includegraphics[height=0.1\textwidth]{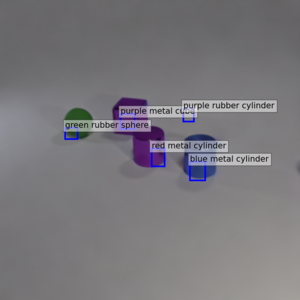}} &
          \adjustbox{valign=c}{\includegraphics[height=0.1\textwidth]{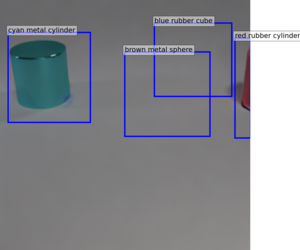}} &
          \adjustbox{valign=c}{\includegraphics[height=0.1\textwidth]{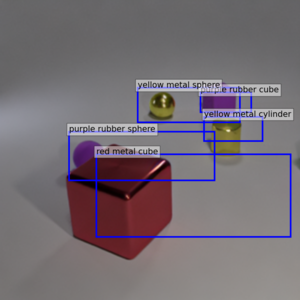}} &
          \adjustbox{valign=c}{\includegraphics[height=0.1\textwidth]{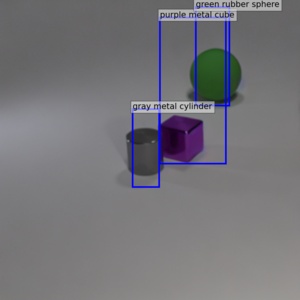}} \\
          \midrule
          \reco{} &
          \adjustbox{valign=c}{\includegraphics[height=0.1\textwidth]{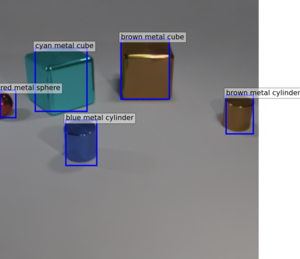}} &
          \adjustbox{valign=c}{\includegraphics[height=0.1\textwidth]{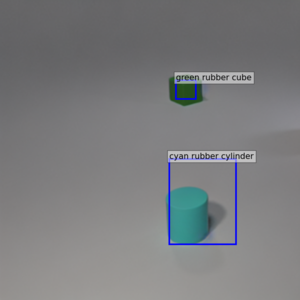}} &
          \adjustbox{valign=c}{\includegraphics[height=0.1\textwidth]{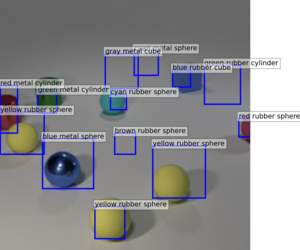}} &
          \adjustbox{valign=c}{\includegraphics[height=0.1\textwidth]{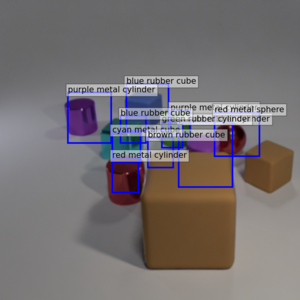}} &
          \adjustbox{valign=c}{\includegraphics[height=0.1\textwidth]{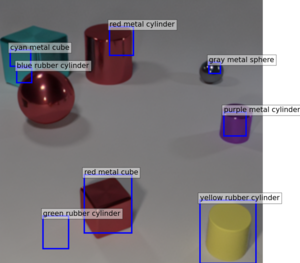}} &
          \adjustbox{valign=c}{\includegraphics[height=0.1\textwidth]{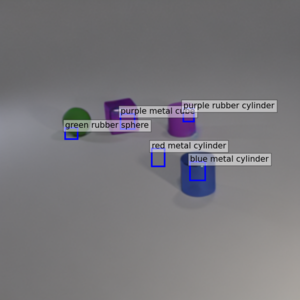}} &
          \adjustbox{valign=c}{\includegraphics[height=0.1\textwidth]{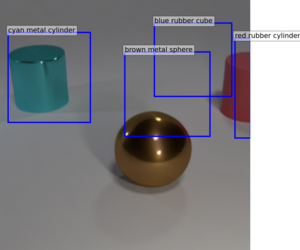}} &
          \adjustbox{valign=c}{\includegraphics[height=0.1\textwidth]{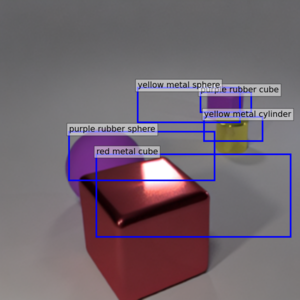}} &
          \adjustbox{valign=c}{\includegraphics[height=0.1\textwidth]{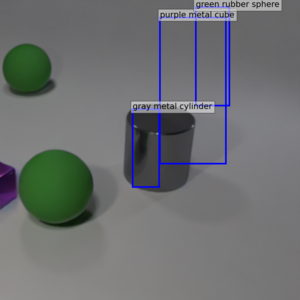}} \\        
          \midrule
          \specialcelll{\method{} \\ (Ours)} &
          \adjustbox{valign=c}{\includegraphics[height=0.1\textwidth]{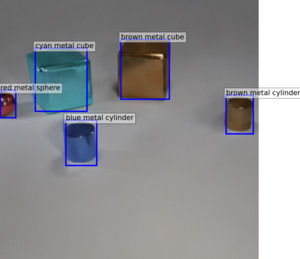}} &
          \adjustbox{valign=c}{\includegraphics[height=0.1\textwidth]{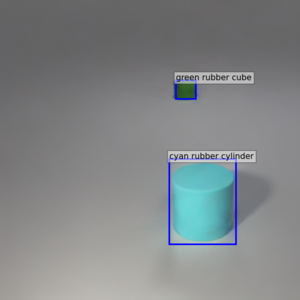}} &
          \adjustbox{valign=c}{\includegraphics[height=0.1\textwidth]{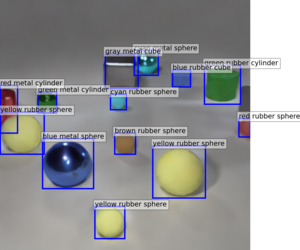}} &
          \adjustbox{valign=c}{\includegraphics[height=0.1\textwidth]{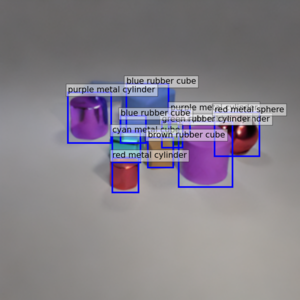}} &
          \adjustbox{valign=c}{\includegraphics[height=0.1\textwidth]{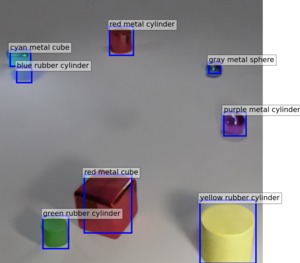}} &
          \adjustbox{valign=c}{\includegraphics[height=0.1\textwidth]{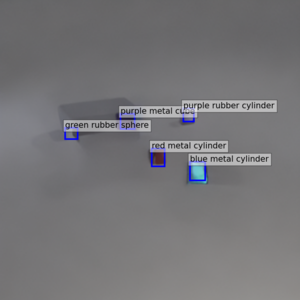}} &
          \adjustbox{valign=c}{\includegraphics[height=0.1\textwidth]{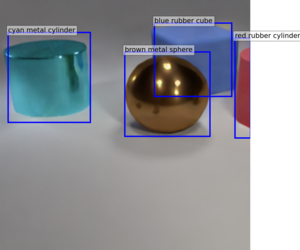}} &
          \adjustbox{valign=c}{\includegraphics[height=0.1\textwidth]{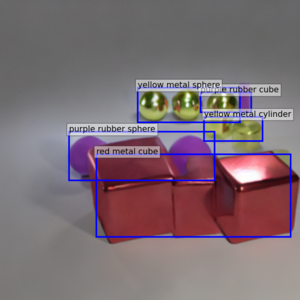}} &
          \adjustbox{valign=c}{\includegraphics[height=0.1\textwidth]{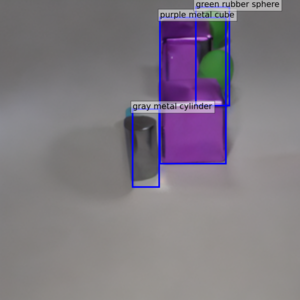}} \\
          \bottomrule
        \end{tabular}
    }
\caption{
Comparison of generated images on \clevr{} (ID) and \benchmark{} (OOD). GT boxes are shown in \textcolor{blue}{blue}. 
}
\label{tab:generation_samples}
\end{table*}
\begin{figure*}[t]
  \centering
  \includegraphics[width=.9\textwidth]{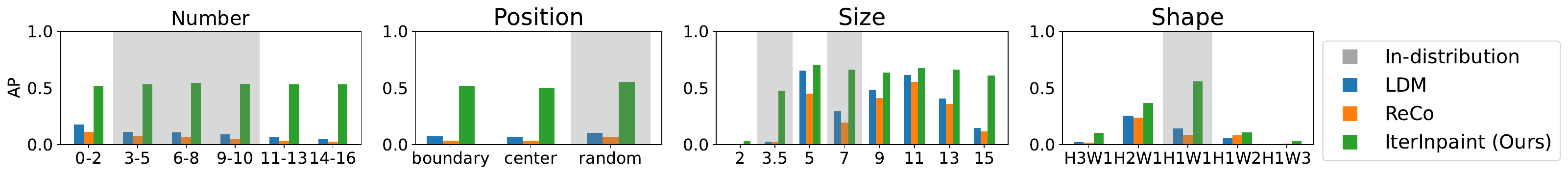}
  \caption{Detailed layout accuracy analysis with fine-grained splits of 4 \benchmark{} skills.
  In-distribution (same attributes to \clevr{}) splits are colored in \textcolor{gray}{gray}.
  For the Shape skill, the splits are named after their height/width ratio (\eg H2W1 split consists of the objects with a 2:1 ratio of height:width).
  }
  \label{fig:analysis}
\end{figure*}

\subsection{Evaluation on \benchmark{}}
\label{sec:quant_eval}

\paragraph{Quantitative evaluation.}
We first evaluate the layout accuracy on generated images in \cref{tab:overall_results_map}.
The first row  shows the layout accuracy based on the ground-truth (GT) images.
Our object detector evaluator can achieve high accuracy on both \clevr{} and \benchmark{} datasets, showing the high reliability of the detection-based layout accuracy evaluation results.
Especially on \benchmark{}, 
our detector achieves above 98\% AP$_{50}$.\footnote{The AP of GT \clevr{} images (60.5) is a bit lower than that of GT \benchmark{} (90.7), because \clevr{} bounding box annotations provided by Krishna \etal{} (2018) \cite{Krishna2018ReferringRelationships} have minor errors.
On our re-rendered \clevr{} images with precise bounding boxes, the object detector could achieve 99\% AP (see appendix for details).} 
The second row (GT shuffled) shows a setting where given a target layout, we randomly sample an image from the GT images to be the generated image.
The 0\% AP on both \clevr{} and \benchmark{} means that it is impossible to obtain high AP by only generating high-fidelity images but in the wrong layouts.

In the bottom half of \cref{tab:overall_results_map}, we see that 
while all 3 models achieve high layout accuracy with above 89\% AP$_{50}$ on \clevr{}, the layout accuracy drop by large margins on \benchmark{}, showing the ID-OOD layout gap.
Specifically, \ldm{} and \reco{} fail substantially on \benchmark{} across all skill splits, with an average performance drop of 57$\sim$70\% per skill on AP$_{50}$, compared to the high AP on in-domain \clevr{} validation split. Both models especially struggle with layout configurations with \emph{many}, \emph{tiny}, \emph{horizontally-shaped} objects and when objects are placed in \emph{center/boundary}.
This is not surprising, as we have shown in~\cref{fig:teaser}; we will also show qualitative evaluation later in~\cref{tab:generation_samples} that \ldm{} and \reco{} demonstrate poor generalizability to OOD layouts in \benchmark{}.

In contrast, \method{} can generalize better to OOD layouts in \benchmark{} while maintaining or even slightly improving the layout accuracy on ID layouts in \clevr{}.
Specifically, we observe an average performance gain of 49.5\% for Number, 61.3\% for Position, 15.0\% for Size, and 10.7\% for Shape in AP$_{50}$, compared to \ldm{} and \reco{}.
Even on the extremely challenging Size-tiny split, where \ldm{} and \reco{} fails to render any tiny objects onto the given positions, \method{} can at least break the zero performance. Another challenging case is the shape-horizontal split, where all three models struggle, we further conduct detailed analysis on the difficulty levels of each split and the pain points of each model in \cref{sec:fine_analysis}.

In \cref{tab:overall_results_fid}, we compare the quality of generated images by reporting the FID/SceneFID scores. 
On \clevr{},
the \ldm{} and \reco{} achieves better FID/SceneFID metrics than \method{},
indicating that the strong layout control performance of \method{} comes with a trade-off in these image quality metrics.
However, on \benchmark{}, the three models achieve similar FID scores, despite the significant layout errors of \ldm{} and \reco{},
which suggests that image quality measures alone are not sufficient for evaluating layout-guided image generation~\cite{Frolov2021} and further justify using layout accuracy to examine layout control closely.

\begin{table*}[h]
\centering
      \resizebox{\linewidth}{!}{
      \begin{tabular}{c c c c c c c c}
          \toprule
          \multicolumn{2}{c} {Skill 1: Number} & \multicolumn{2}{c} {Skill 2: Position} & \multicolumn{2}{c} {Skill 3: Size} & \multicolumn{2}{c} {Skill 4: Combination}\\
          \cmidrule(lr){1-2} \cmidrule(lr){3-4} \cmidrule(lr){5-6} \cmidrule(lr){7-8}
          few & many & center & boundary & tiny & large & common & uncommon \\
          \cmidrule(lr){1-2} \cmidrule(lr){3-4} \cmidrule(lr){5-6} \cmidrule(lr){7-8}
4 chairs &
10 cars &
5 buses &
5 suitcases &
3 cars &
3 broccolis &
person is holding tennis racket &
parking meter is next to clock \\

          \adjustbox{valign=c}{\includegraphics[height=0.1\textwidth]{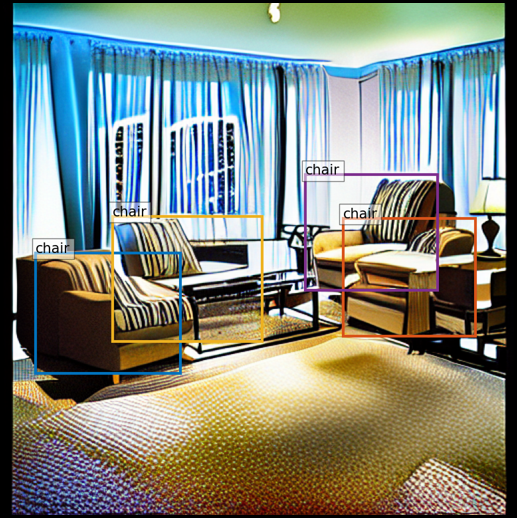}} &
          \adjustbox{valign=c}{\includegraphics[height=0.1\textwidth]{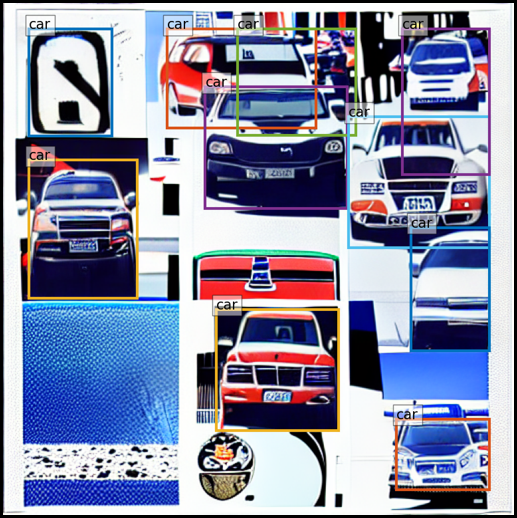}} &
          \adjustbox{valign=c}{\includegraphics[height=0.1\textwidth]{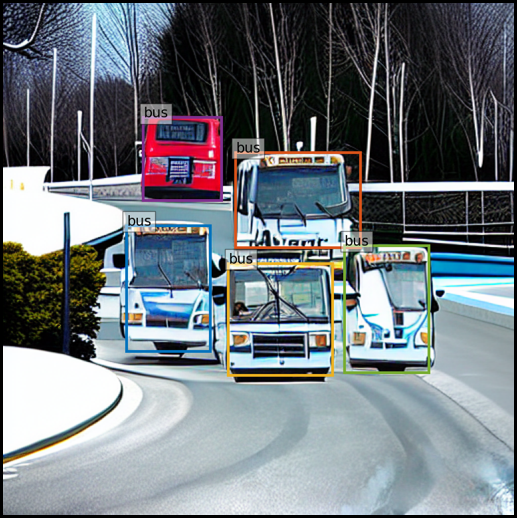}} &
          \adjustbox{valign=c}{\includegraphics[height=0.1\textwidth]{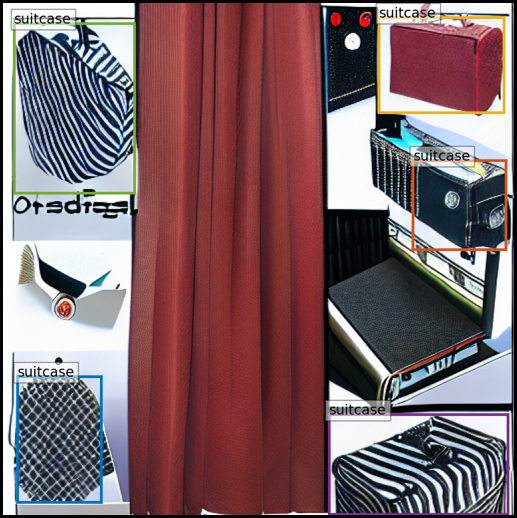}} &
          \adjustbox{valign=c}{\includegraphics[height=0.1\textwidth]{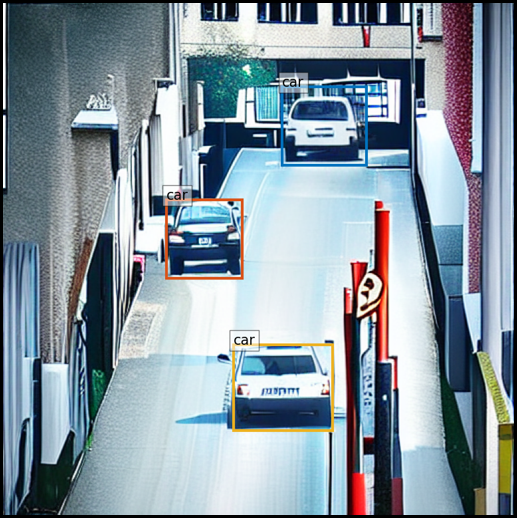}} &
          \adjustbox{valign=c}{\includegraphics[height=0.1\textwidth]{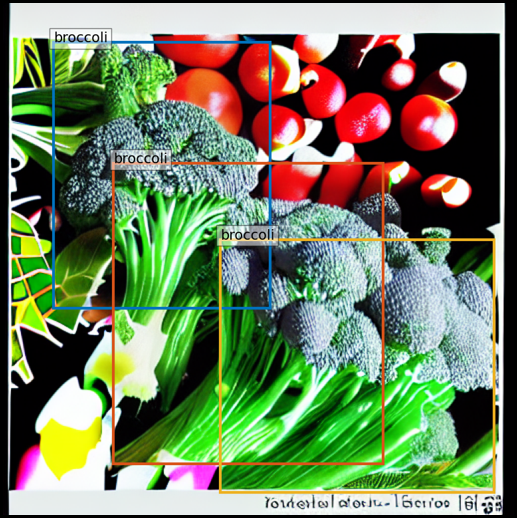}} &
          \adjustbox{valign=c}{\includegraphics[height=0.1\textwidth]{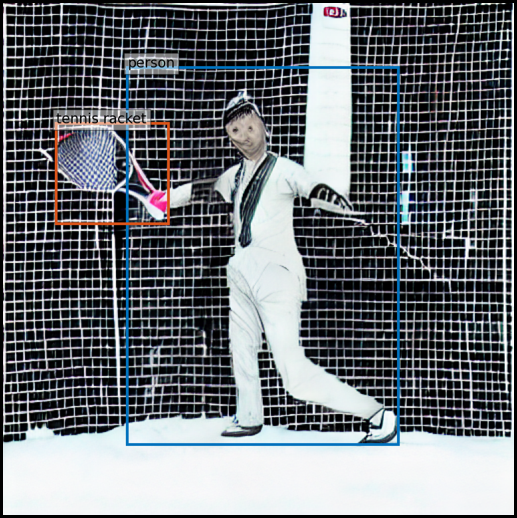}} &
          \adjustbox{valign=c}{\includegraphics[height=0.1\textwidth]{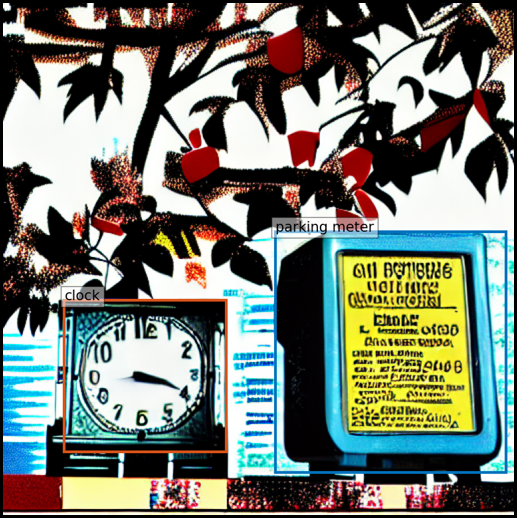}}
          \\
          \bottomrule
        \end{tabular}
        }
\caption{
Example images generated by \method{}
given four splits of caption and layouts from \benchmarkreal{}.
}
\label{tab:layoutbench_coco_images_iterinpaint}
\end{table*}

\paragraph{Qualitative evaluation.} 
\cref{tab:generation_samples} compares the GT images and the images generated by the three models.
On \clevr{}, all three models can follow the ID layout inputs to place the correct objects precisely.
On \benchmark{}, \ldm{} and \reco{} often
make mistakes, such as
generating objects that are much smaller (\eg, Number-few) / bigger (\eg, Size-tiny, Position-center) than the given bounding boxes and missing some objects (\eg, Number-many, Position-center, Position-boundary, Size-large).
However, \method{} can generate objects more accurately aligned to the given bounding boxes in general, consistent with the higher layout accuracy in \cref{tab:overall_results_map}. Especially for the extremely small bounding boxes in Size-tiny, only \method{}, among the three models, generates objects that fit.
Interestingly, on Shape-horizontal/Shape-vertical,
while all three struggle to generate long objects that are not seen in \clevr{},
\method{} tries to fill the given long bounding boxes by generating multiple objects. More qualitative examples per skill are included in Appendix.

\subsection{Fine-grained Skill Analysis}
\label{sec:fine_analysis}

We perform a more detailed analysis on each \benchmark{} skill to better understand the challenges presented in \benchmark{} and examine each method's weakness.
Specifically, we divide the 4 skills into more fine-grained splits to cover both in-distribution (ID; \clevr{} configurations) and out-of-distribution (OOD; \benchmark{} configurations) examples.
We sample 200 images for each split and report layout accuracy in \cref{fig:analysis}.

\paragraph{Overall.}
Comparing across 4 skills,
the majority of Size skill splits (except for size=2) are the least challenging, while the Position/Number skill is the most challenging. \method{} significantly outperforms \ldm{} and \reco{} on all splits.
Among the other two, \ldm{} has slightly higher scores than \reco{} overall.

\paragraph{Number.}
As the number of objects increases in the first plot of \cref{fig:analysis}, \ldm{} and \reco{} performance decreases, while the \method{} performance remains consistent.

\paragraph{Position.}
As shown in the second plot of \cref{fig:analysis}, there is a slight ID-OOD performance gap for all three models.
The models perform similarly on boundary and center splits, while slightly lower than the random ID split.

\paragraph{Size.}
As shown in the third plot of \cref{fig:analysis}, the models are better at generating large objects than small objects.
Notably, all models fail at size=2, the smallest object scale in our experiment.
As shown in \cref{tab:generation_samples}'s Size-tiny column, \ldm{} and \reco{} tend to generate bigger objects from small bounding boxes, whereas \method{} could correctly generate small objects in the right location but misses the details of right shapes or attributes.

\paragraph{Shape.}
As shown in the last plot of \cref{fig:analysis}, there is a strong ID-OOD gap for all three models.
The models generate vertically long (H2W1 and H3W1) better than horizontally long (H1W2 and H1W3) objects.
From our manual analysis, there were some trends for models to prioritize fitting the height to the width of the bounding boxes.
This results in trends of generating small square boxes for horizontally long boxes that are too small to cover the box and generating big square boxes that can sometimes cover some vertically long boxes (see appendix for more examples).

\begin{table}
\centering
    \resizebox{.9\linewidth}{!}{
      \begin{tabular}{l c c c c}
          \toprule
          Method & Number & Position & Size & Combination \\
          \midrule

        ControlNet	 & 9.2 & 15.3 & 10.8 & 6.4 \\
        GLIGEN	 & 30.7 & 36.4 & 33.3 & 36.3 \\
        \reco{}	 & 30.9 & 38.9 & 24.1 & 18.7 \\
        \method{} (Ours)	 & \textbf{31.4} & \textbf{39.1} & \textbf{33.5} & \textbf{44.1} \\
        
          \bottomrule
        \end{tabular}
        }
\caption{Zero-shot Layout Accuracy in AP (\%) on \benchmarkreal{}.
}
\label{tab:layoutbench_coco}
\end{table}

\subsection{Ablation of \method{}}
\label{sec:ablation}

We conduct ablation studies of \method{} design choices:
(1) Pasting (default) \vs Repaint based update,
(2) training task ratio for foreground \& background inpainting,
and (3) object generation order.
In summary, we found that (1) repaint-based update suffers from error propagation,
(2) the 3:7 fg/bg training task ratio performs best, but other task ratios perform similarly,
and (3) \method{} is robust in arbitrary generation orders, allowing flexible object layout manipulation without full image re-rendering.
Please see appendix for detailed analysis.

\subsection{\benchmarkreal{}: Zero-shot Evaluation of Layout-guided Image Generation Models}
\label{sec:layoutbench_coco}

Although our main focus is to provide a benchmark of layout-guided image generation models with full control,
including the same computation and training data with arbitrary objects (\eg{}, blue metal cube),
we also test the spatial control capabilities of existing pretained models 
with layouts of real objects (\eg{}, cars)
in zero-shot.
For this,
we create \benchmarkreal{},
a real object version of \benchmark{} with 4 splits (Number, Position, Size, Combination), whose objects are from MS COCO~\cite{Lin2014COCO}.
The new `combination' split consists of layouts with two objects in different spatial relations, and the remaining three splits are similar to those of \benchmark{}.

We compare four models,
covering both models with segmentation mask inputs (ControlNet~\cite{Zhang2023ControlNet}; we create segmentation masks by drawing bounding boxes with class-specific colors),
and bounding box inputs (GLIGEN~\cite{li2023gligen},
\reco{}~\cite{Yang2022ReCo}, and our \method{} trained on COCO).
For evaluation metric, we use layout accuracy in AP with a state-of-the-art object detector, YOLOv7~\cite{wang2023yolov7}.

In \cref{tab:layoutbench_coco_images_iterinpaint}, we show images generated by \method{} on four splits of \benchmarkreal{} (see appendix for more generation examples from other methods).
\cref{tab:layoutbench_coco} shows the layout accuracy of four evaluated models.
The bounding box-based models outperform the
segmentation mask-based model (ControlNet).
Our \method{} achieves higher layout accuracy than baselines in all four splits, especially with a large margin in the combination split.
The experiment results indicate the effectiveness of \method{} handling the challenging layouts.

\section{Conclusion}
\label{sec:conclusion}

We introduce \benchmark{}, a diagnostic benchmark that systemically evaluates four spatial control skills of layout-guided image generation models: number, position, size, and shape.
We show that recent layout-guided image generation methods do not generalize well on OOD layouts (\emph{e.g.}, many/large objects).
Next, we propose~\method{}, a new baseline that generates foreground and background regions step-by-step.
In our detailed analysis of spatial control skills,
\method{} has stronger generalizability than baselines on OOD layouts.
We hope that our
work facilitates future work on controllable image generation.

{
    \small
    \bibliographystyle{ieeenat_fullname}
    \bibliography{references}
}


\appendix

\section*{Appendix}

\renewcommand{\thesection}{\Alph{section}}

In this Appendix,
we provide
additional \method{} experiments, including image generation with user-defined layouts, interactive image manipulation, ablation studies, and training on COCO images
(\cref{appendix_sec:additional_experiments}),
\benchmarkreal{} details (\cref{appendix_sec:layoutbench_coco_details}),
examination of \clevr{} GT layout accuracy (\cref{appendix_sec:obj_det_acc}),
additional GAN baseline (\cref{appendix_sec:additional_GAN_baseline}),
and additional \benchmark{} image samples (\cref{appendix_sec:additional_image_samples}).

\section{Additional \method{} Experiments}
\label{appendix_sec:additional_experiments}

\begin{table}[h]
\begin{center}
    \resizebox{\linewidth}{!}{
      \begin{tabular}{l c c}
          \toprule

          Layout & \reco{} & \method{} \\
          \midrule

          \adjustbox{valign=c}{\includegraphics[height=0.14\textwidth]{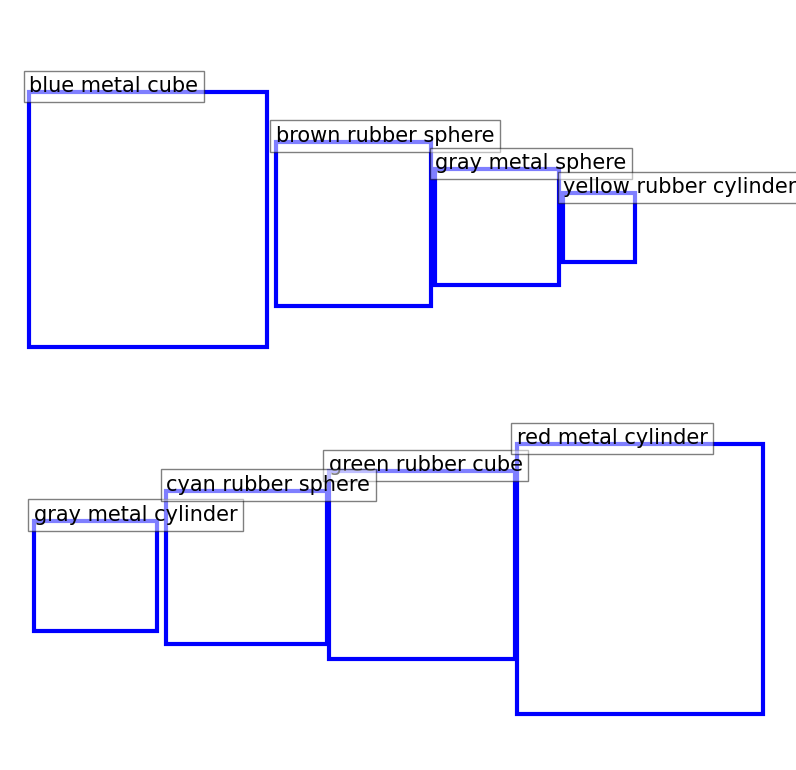}} &
          \adjustbox{valign=c}{\includegraphics[height=0.14\textwidth]{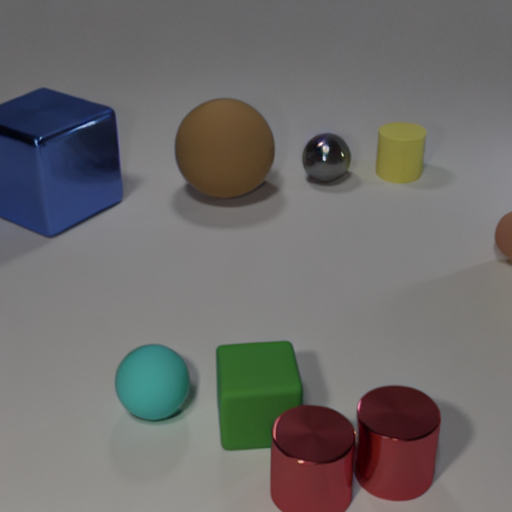}} &
          \adjustbox{valign=c}{\includegraphics[height=0.14\textwidth]{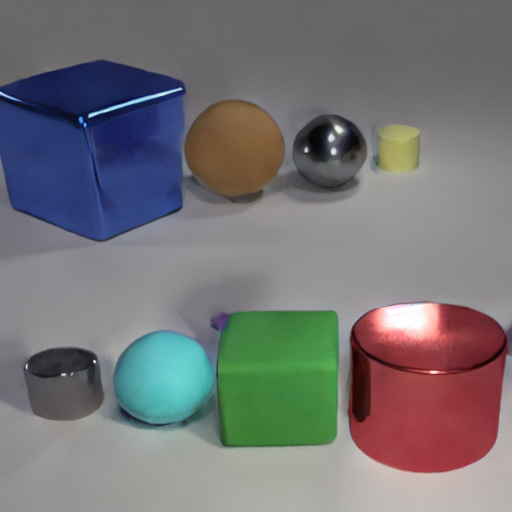}} \\

          \midrule

          \adjustbox{valign=c}{\includegraphics[height=0.14\textwidth]{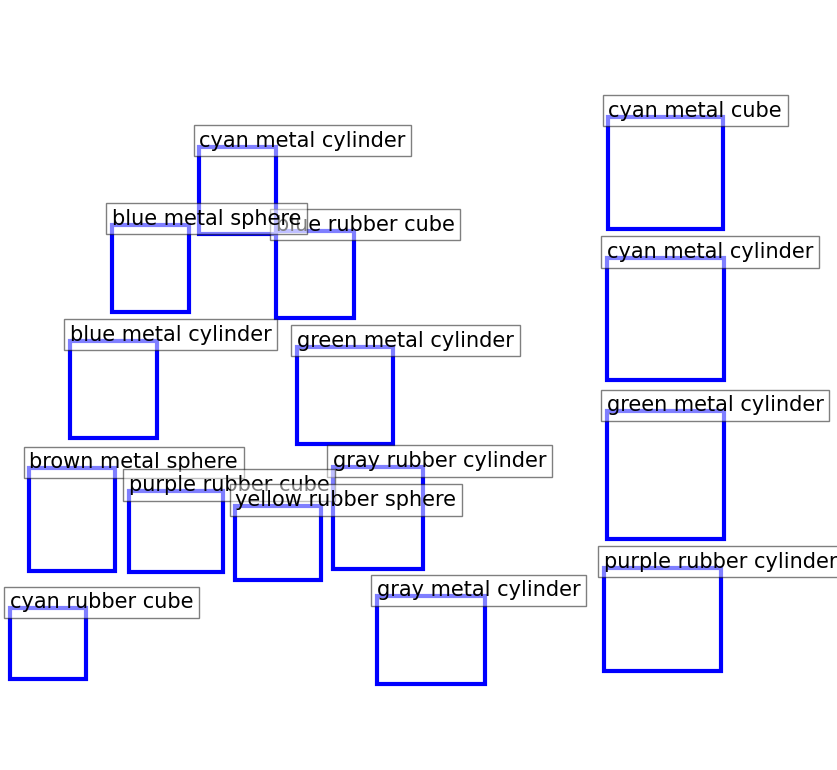}} &
          \adjustbox{valign=c}{\includegraphics[height=0.14\textwidth]{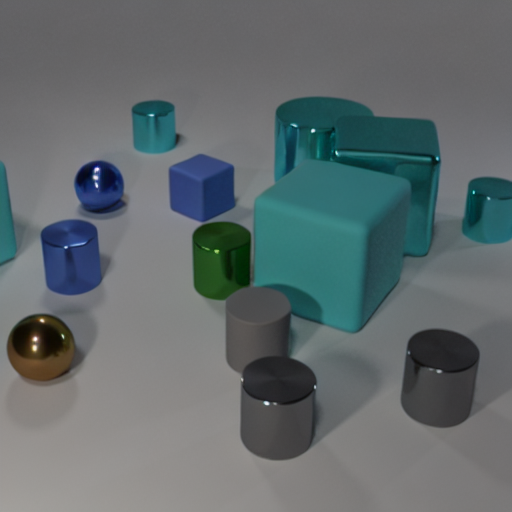}} &
          \adjustbox{valign=c}{\includegraphics[height=0.14\textwidth]{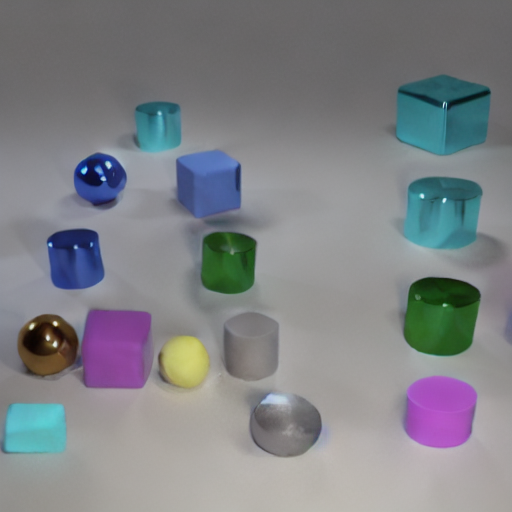}} \\

          \midrule

          \adjustbox{valign=c}{\includegraphics[height=0.14\textwidth]{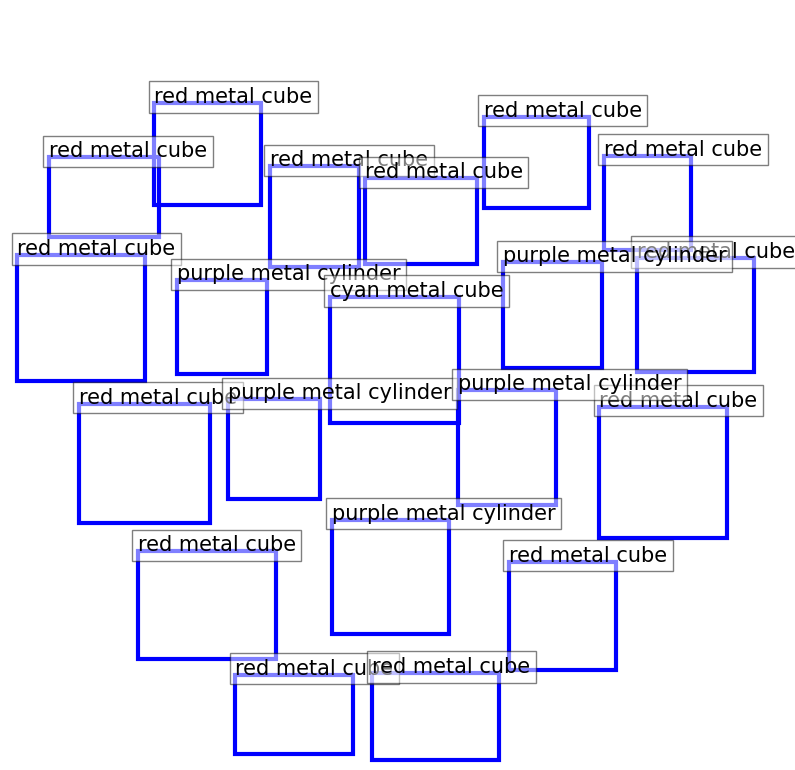}} &
          \adjustbox{valign=c}{\includegraphics[height=0.14\textwidth]{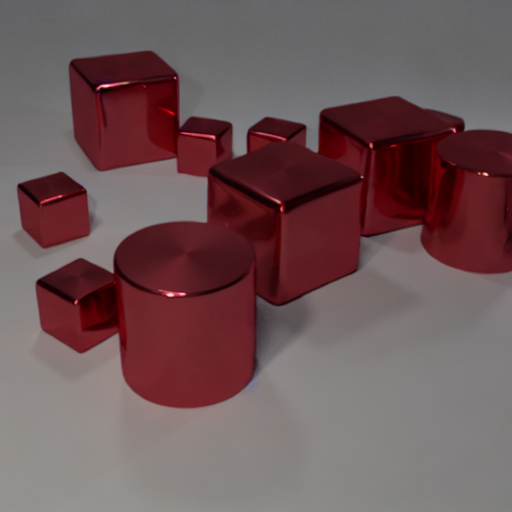}} &
          \adjustbox{valign=c}{\includegraphics[height=0.14\textwidth]{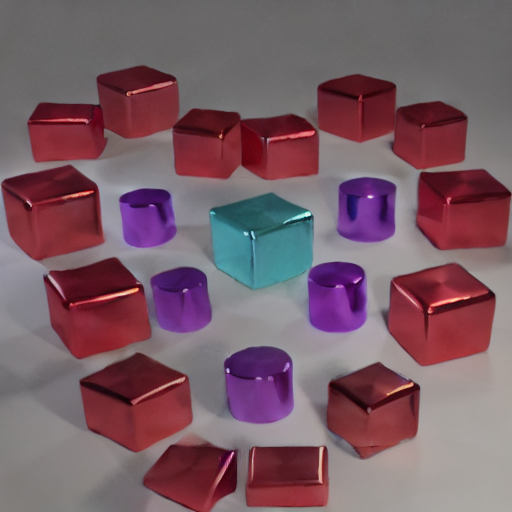}} \\
          
          \bottomrule
        \end{tabular}
    }
\end{center}
\caption{Image generation from some user-defined layouts with \method{} and \reco{} trained on \clevr{}.
On the leftmost column, we show three input layouts: (1) two rows of objects with different sizes, (2) `AI' written in the text, and (3) a heart shape.
}
\label{tab:arbitrary_layout_generation}
\end{table}

\subsection{Image Generation with User-defined Layouts}

\Cref{tab:arbitrary_layout_generation} shows image generation results from user-defined layouts with \clevr{} objects: (1) two rows of objects with different sizes, (2) `AI' written in the text, and (3) a heart shape.
While \reco{} often fails to ignore or misplace some objects, \method{} places objects significantly more accurately.
This shows the high robustness of \method{}, which can follow even more abstract and complex than the automatically generated layouts of \benchmark{}.

\subsection{Interactive Image Manipulation}

\begin{figure}[h]
\begin{center}
    \includegraphics[width=.75\linewidth]{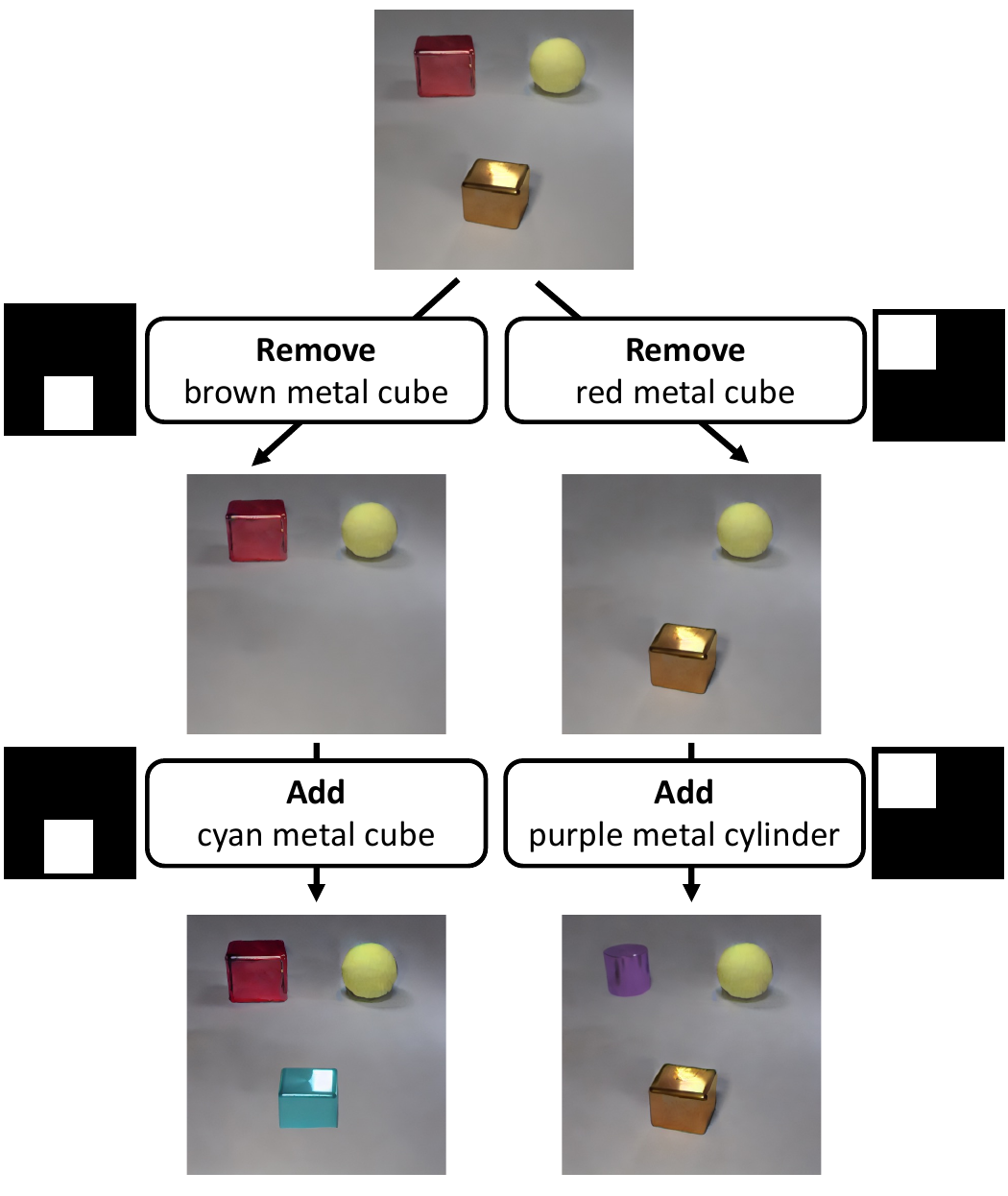}
\end{center}
    \caption{Interactive image manipulation with \method{}.
    Users can create a binary mask (white: region to update / black: region to preserve) to 
    add or remove objects at custom locations.
    }
    \label{fig:interactive_manupulation}
\end{figure}

\Cref{fig:interactive_manupulation} shows interactive image manipulation examples.
With \method{}, users can interactively remove or add objects from a given image at arbitrary locations.
When removing objects, we use a prompt `Add gray background'.

\begin{table*}[t]
\centering
      \resizebox{\linewidth}{!}{
            \begin{tabular}{l c c c c c c | c}
            \toprule
            Update & \multicolumn{6}{c |}{{ Intermediate Generation}}

            & {  GT image} \\
            \midrule

            & Step 1 & Step 2 & Step 3 & Step 4 & Step 5 & Step 6 &  \\
            Prompts & Add gray rubber sphere & Add green metal sphere & Add purple metal cylinder & Add purple rubber cube & Add cyan rubber cylinder & Add gray background &  \\
            \midrule
            
            \makecell[l]{Crop\&Paste \\ (default)}& \adjustbox{valign=c}{\includegraphics[height=0.1\textwidth]{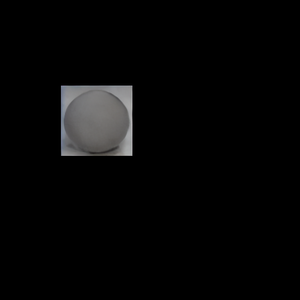}} &
            \adjustbox{valign=c}{\includegraphics[height=0.1\textwidth]{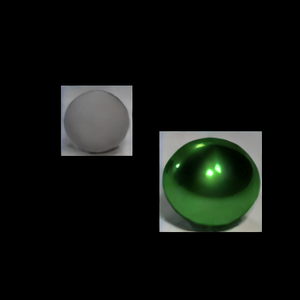}} &
            \adjustbox{valign=c}{\includegraphics[height=0.1\textwidth]{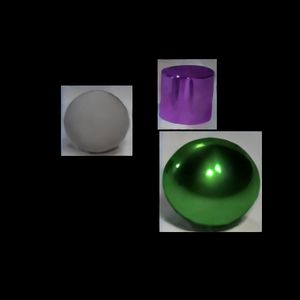}} &
            \adjustbox{valign=c}{\includegraphics[height=0.1\textwidth]{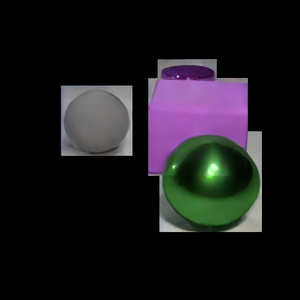}} &
            \adjustbox{valign=c}{\includegraphics[height=0.1\textwidth]{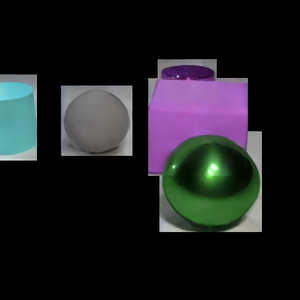}} &
            \adjustbox{valign=c}{\includegraphics[height=0.1\textwidth]{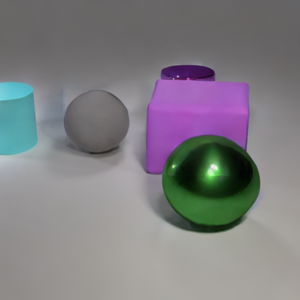}} &
            
            \adjustbox{valign=c}{\includegraphics[height=0.1\textwidth]{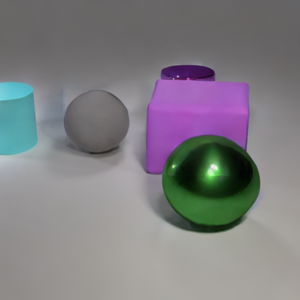}} \\

            \midrule

            Repaint & \adjustbox{valign=c}{\includegraphics[height=0.1\textwidth]{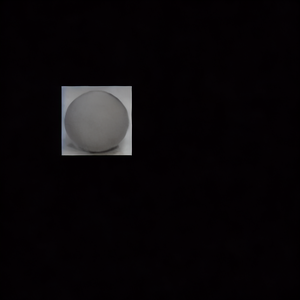}} &
            \adjustbox{valign=c}{\includegraphics[height=0.1\textwidth]{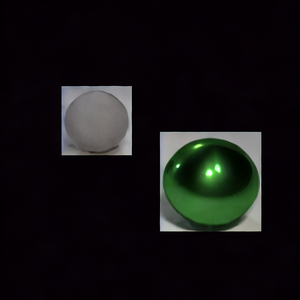}} &
            \adjustbox{valign=c}{\includegraphics[height=0.1\textwidth]{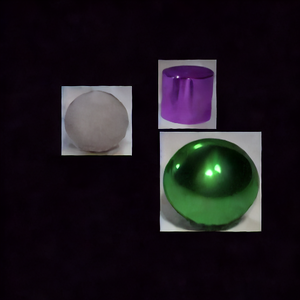}} &
            \adjustbox{valign=c}{\includegraphics[height=0.1\textwidth]{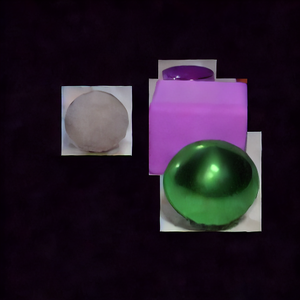}} &
            \adjustbox{valign=c}{\includegraphics[height=0.1\textwidth]{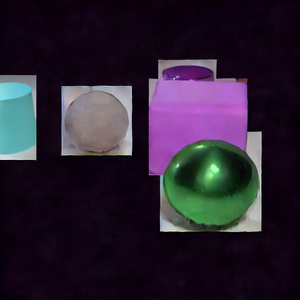}} &
            \adjustbox{valign=c}{\includegraphics[height=0.1\textwidth]{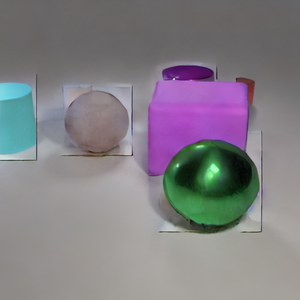}} &
            
            \adjustbox{valign=c}{\includegraphics[height=0.1\textwidth]{images/intermediate_generation/LayoutBench_val_size_110_001984.png}} \\

            \bottomrule
          
        \end{tabular}
      }
  \caption{
  Intermediate generation samples of \method{} with crop\&paste (top) and repaint (bottom) based image updates.
  }
  \label{tab:intermediate_generation_samples}
\end{table*}

\begin{table}[t]

\begin{center}
    \resizebox{\linewidth}{!}{
      \begin{tabular}{c c c c c}
          \toprule
          Image Update & FG/BG Training Ratio & Generation Order & Checkpoint & \benchmark{} AP \\
          \midrule
          
          Crop \& Paste & 3:7 & Random & SD Inpaint & 36.5 \\

          \midrule

          Repaint & & & & 24.4 \textcolor{red}{(-12.1)} \\

          \midrule

          & 1:9 & & & 36.0 \textcolor{red}{(-0.5)} \\
          & 2:8 & & & 35.2 \textcolor{red}{(-1.3)} \\
          & 4:6 & & & 35.8 \textcolor{red}{(-0.7)} \\
          & 5:5 & & & 35.0 \textcolor{red}{(-1.5)} \\
          & 6:4 & & & 34.4 \textcolor{red}{(-2.1)} \\
          & 7:3 & & & 34.2 \textcolor{red}{(-2.3)} \\
          & 8:2 & & & 34.2 \textcolor{red}{(-2.3)} \\
          & 9:1 & & & 32.9 \textcolor{red}{(-3.6)} \\
          & 10:0 (No BG) & & & 27.5 \textcolor{red}{(-9.0)} \\

          \midrule
          
          & & Top $\rightarrow$ Bottom & & 37.2 \textcolor{green}{(+0.7)} \\
          & & Bottom $\rightarrow$ Top & & 36.1 \textcolor{red}{(-0.4)} \\
          
          \midrule
          
          & & & SD v1.4 & 31.4 \textcolor{red}{(-5.1)} \\
          
          \bottomrule
        \end{tabular}
    }
\end{center}
\caption{Ablation studies on \method{}
}
\label{tab:ablation_extended}
\end{table}

\subsection{\method{} Ablation Study}
\label{appendix_sec:ablation_study}

\paragraph{Pasting \vs Repaint.}
Instead of crop\&paste, we experiment the repainting the entire image during the inference.
As shown in \cref{tab:ablation_extended}, repainting results in -12.1\% AP.
\cref{tab:intermediate_generation_samples}
shows that the repaint-based update encodes/decodes the whole image at each step and suffers from error propagation (\ie{}, early objects get distorted with step progress).

\paragraph{Training task ratio for foreground \& background inpainting.}
Sec.~4.3 in the main paper
shows the \benchmark{} layout accuracy with different ratios for foreground/background inpainting tasks.
We found that the 3:7 ratio performs best, but other task ratios perform similarly,
except for the case of 10:0, where not using background inpainting tasks results in a significant -9.0\% drop in AP.

\paragraph{Object generation order.}
Because of the camera angle of \clevr{} simulator, if there is an occlusion between objects, it always takes the form of an object in the bottom (front), occluding the object above it in 2D coordinates. 
We compare 3 generation orders for rendering objects given the spatial coordinates: top$\rightarrow$bottom, bottom$\rightarrow$top, and random.
Unlike other iterative generation models that are sensitive in generation orders~\cite{Cho2020,Ramesh2021},
\cref{tab:ablation_extended} shows that \method{} achieves similar AP with all 3 generation orders.
The robustness in arbitrary generation order would be useful when users want to manipulate object layouts without re-rendering images from scratch.

\paragraph{SD Checkpoint.}
In our experiments, the SD `inpainting' checkpoint~\footnote{\url{https://huggingface.co/runwayml/stable-diffusion-inpainting}} shows slightly better layout accuracy than v1.4 checkpoint\footnote{\url{https://huggingface.co/CompVis/stable-diffusion-v1-4}}, so we use the inpainting checkpoint by default for \method{}.

\subsection{Training on COCO images}

Although our main focus is constructing a diagnostic \benchmark{} benchmark with full scene control and evaluating the layout-guided image generation models,
we also test whether our proposed \method{} baseline model could also perform well on real images.
We train \method{} 
on MS COCO~\cite{Lin2014COCO} train2014 split,
for around 100 epochs with batch size 512 (after applying gradient accumulations).
In \Cref{tab:coco_image_samples}, we show some image generation samples from \reco{} (COCO checkpoint)
and \method{} from COCO layouts, where both models could locate objects in the correct positions.
In \Cref{tab:coco_image_samples_arbitrary}, we show some arbitrary custom layouts with COCO objects.
While both models are correct in object locations, \reco{} sometimes fails to place wrong objects that are frequent in a given layout, while \method{} shows a more precise object recognition performance.
Note that the original \reco{} model was trained with much bigger expensive training resources (e.g., \reco{} uses batch size 2048 vs. our \method{} uses batch size 512), and we leave bigger scale training and hyperparameter tuning to future works.

\begin{table*}[t]
\begin{center}
    \resizebox{.95\linewidth}{!}{
      \begin{tabular}{l c c c}
          \toprule

          Layout & GT image & \reco{} & \method{} \\
          \midrule
          
          \adjustbox{valign=c}{\includegraphics[height=0.14\textwidth]{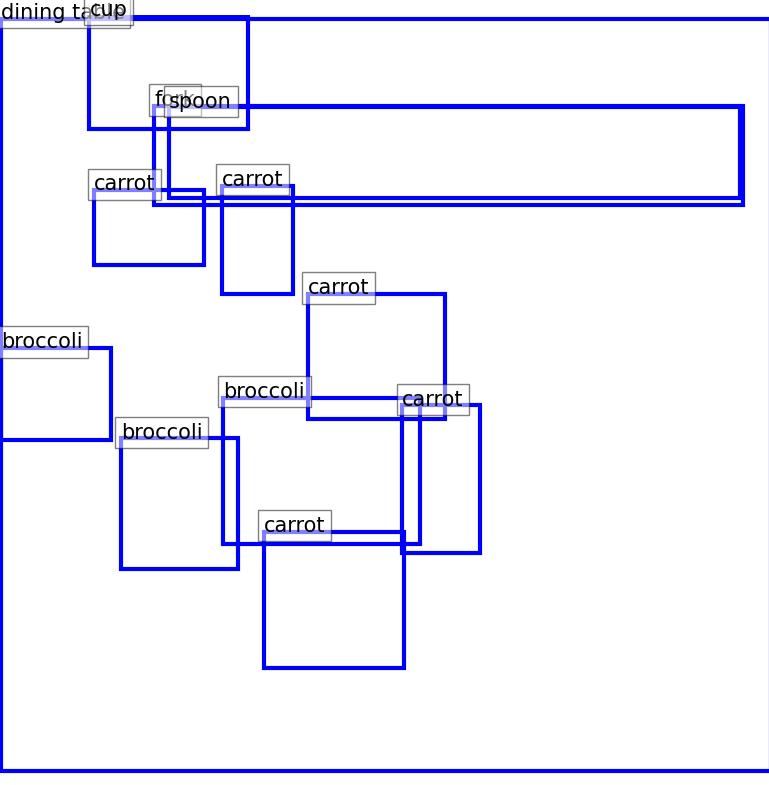}} &
          \adjustbox{valign=c}{\includegraphics[height=0.14\textwidth]{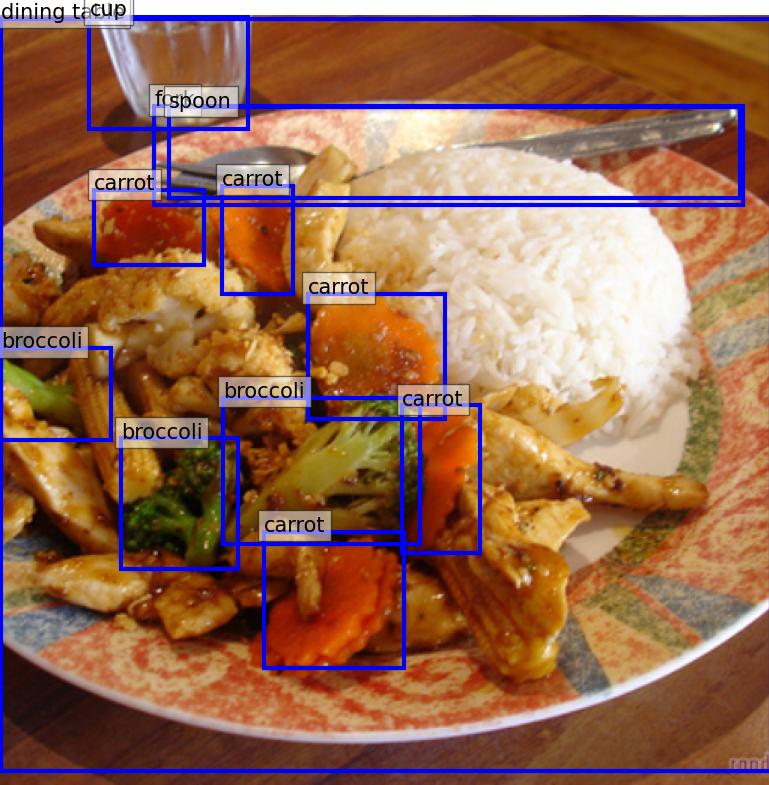}} &
          \adjustbox{valign=c}{\includegraphics[height=0.14\textwidth]{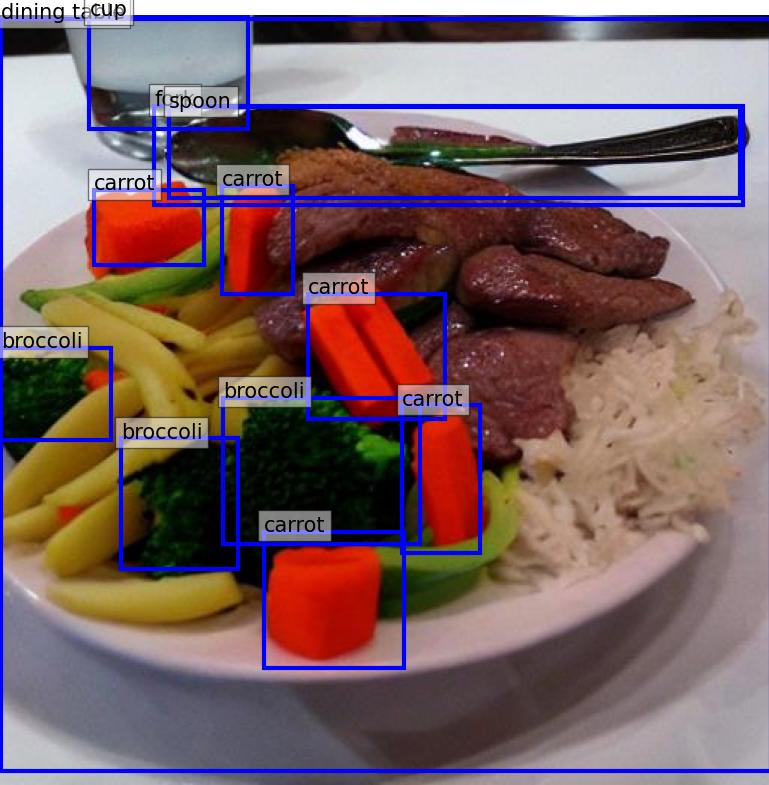}} &
          \adjustbox{valign=c}{\includegraphics[height=0.14\textwidth]{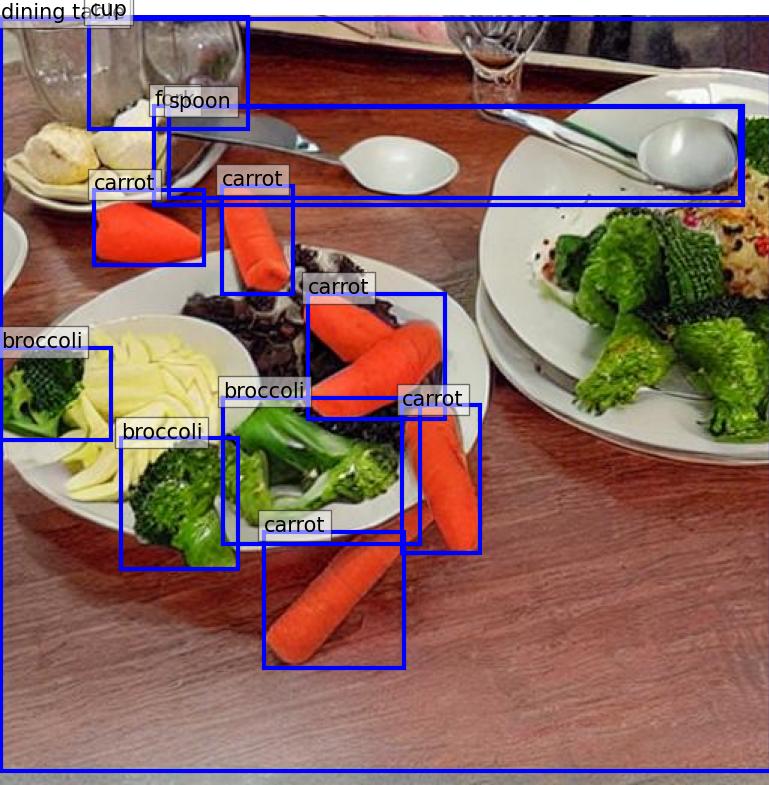}}
          \\

          \midrule

          \adjustbox{valign=c}{\includegraphics[height=0.14\textwidth]{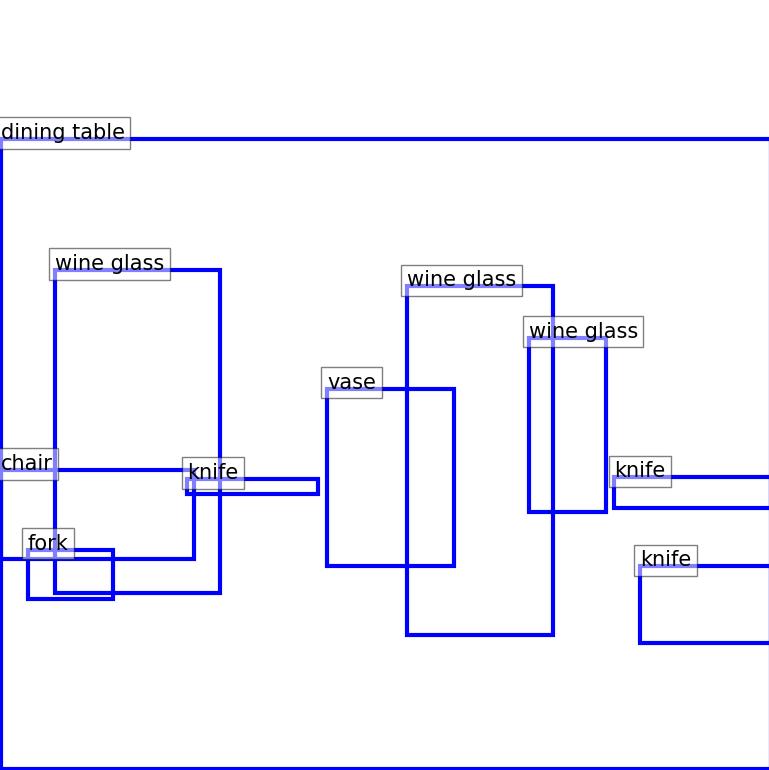}} &
          \adjustbox{valign=c}{\includegraphics[height=0.14\textwidth]{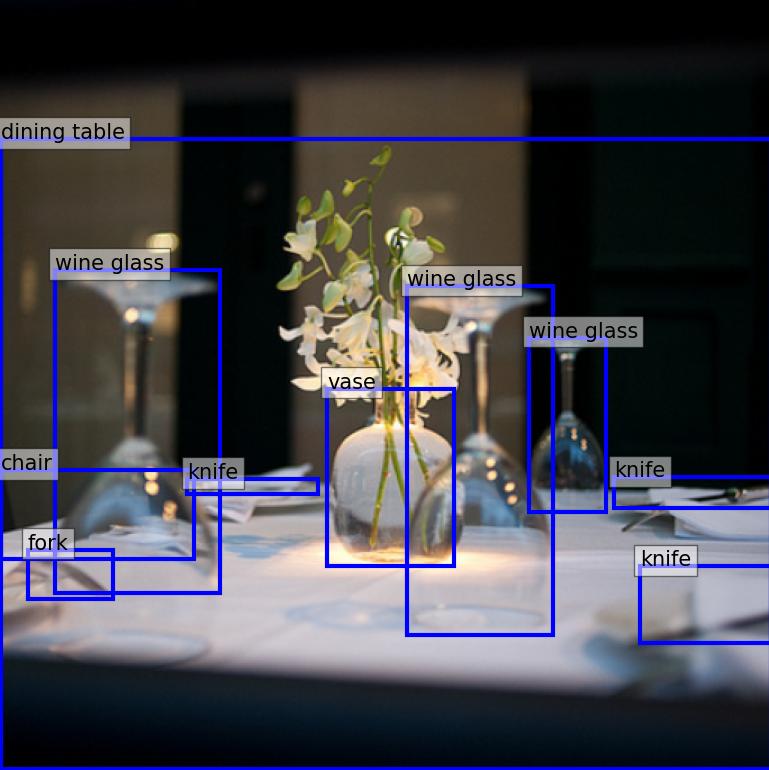}} &
          \adjustbox{valign=c}{\includegraphics[height=0.14\textwidth]{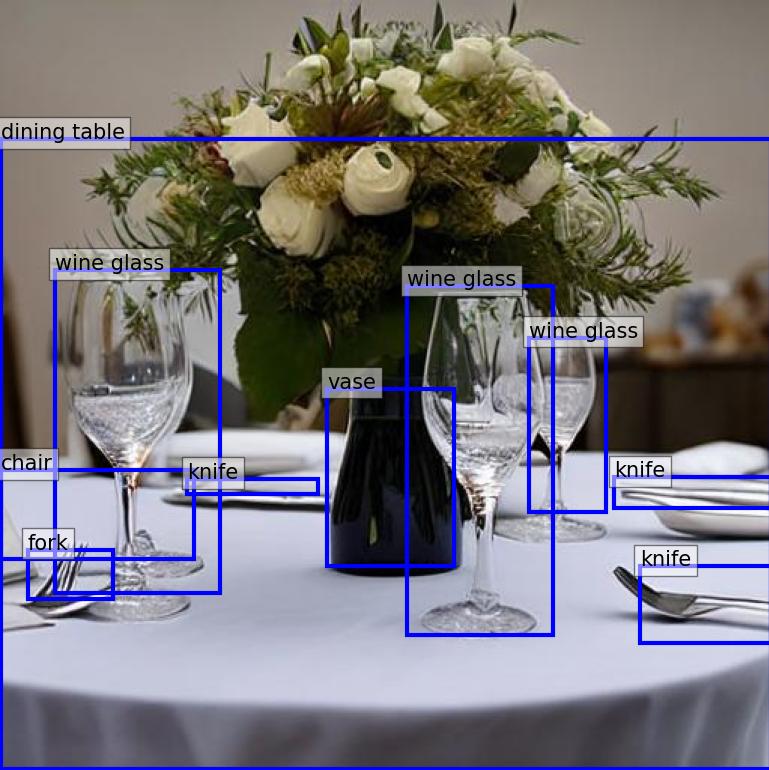}} &
          \adjustbox{valign=c}{\includegraphics[height=0.14\textwidth]{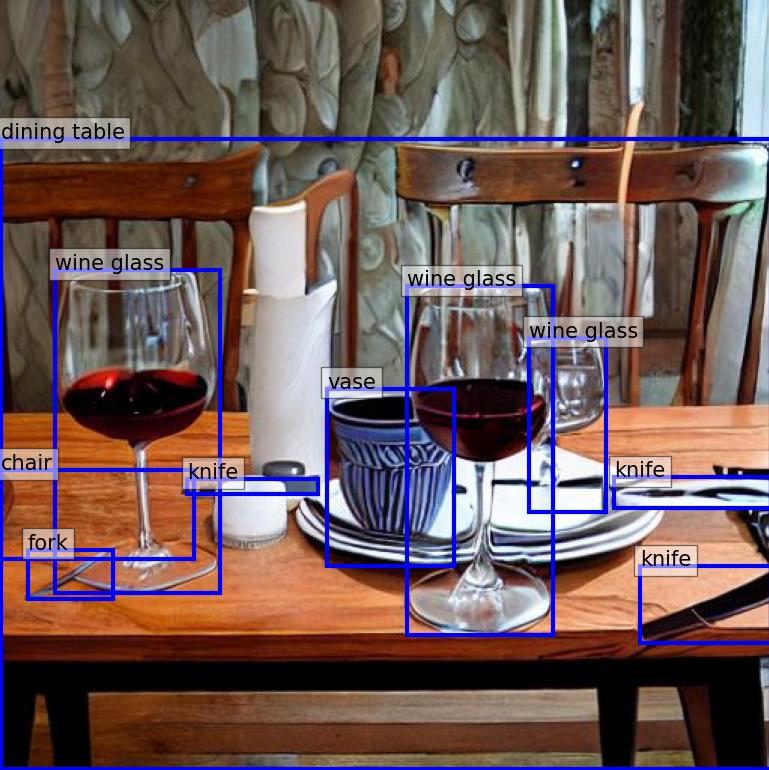}} \\

          \bottomrule
        \end{tabular}
    }
\end{center}
\caption{Image generation samples with COCO layouts (in-distribution) from \method{} and \reco{}.
}
\label{tab:coco_image_samples}
\end{table*}

\begin{table*}[t]
\begin{center}
    \resizebox{.8\linewidth}{!}{
      \begin{tabular}{l c c}
          \toprule

          Layout & \reco{} & \method{} \\
          \midrule

          \adjustbox{valign=c}{\includegraphics[height=0.14\textwidth]{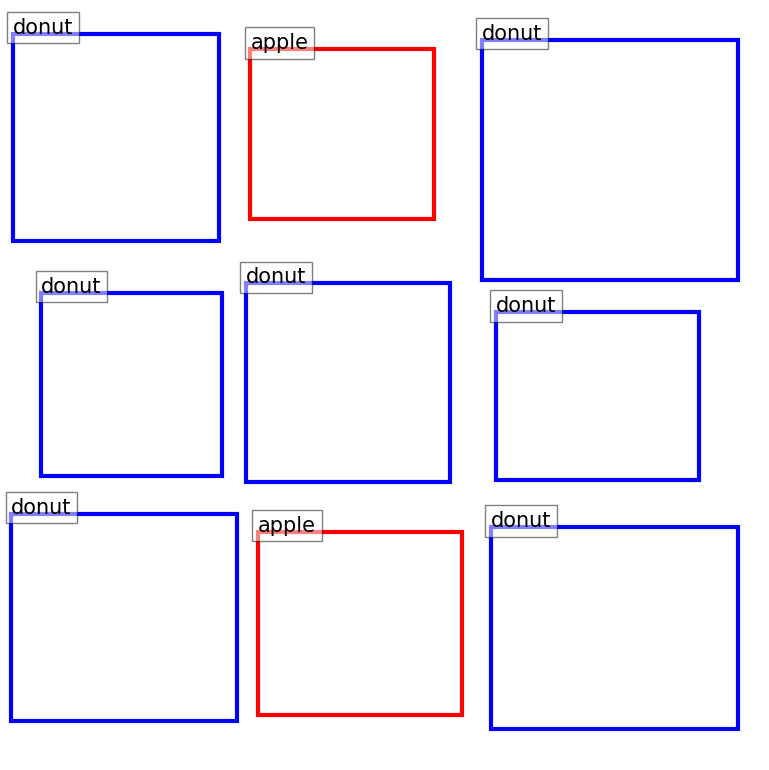}} &
          \adjustbox{valign=c}{\includegraphics[height=0.14\textwidth]{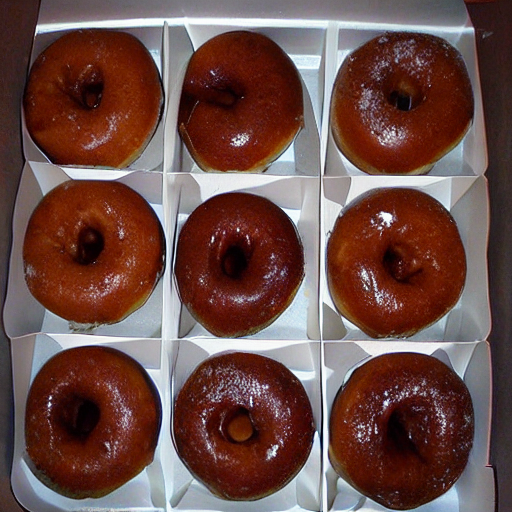}} &
          \adjustbox{valign=c}{\includegraphics[height=0.14\textwidth]{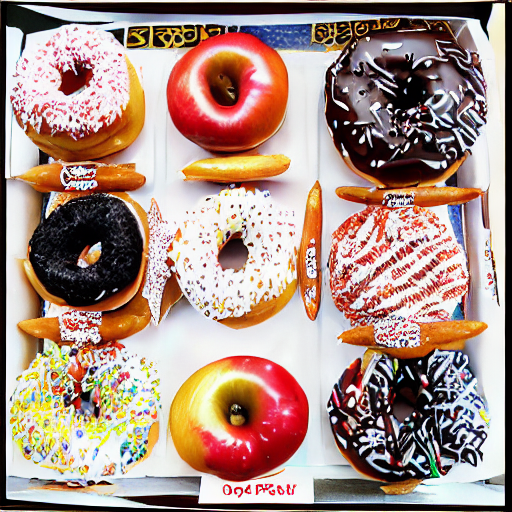}} \\

          \adjustbox{valign=c}{\includegraphics[height=0.14\textwidth]{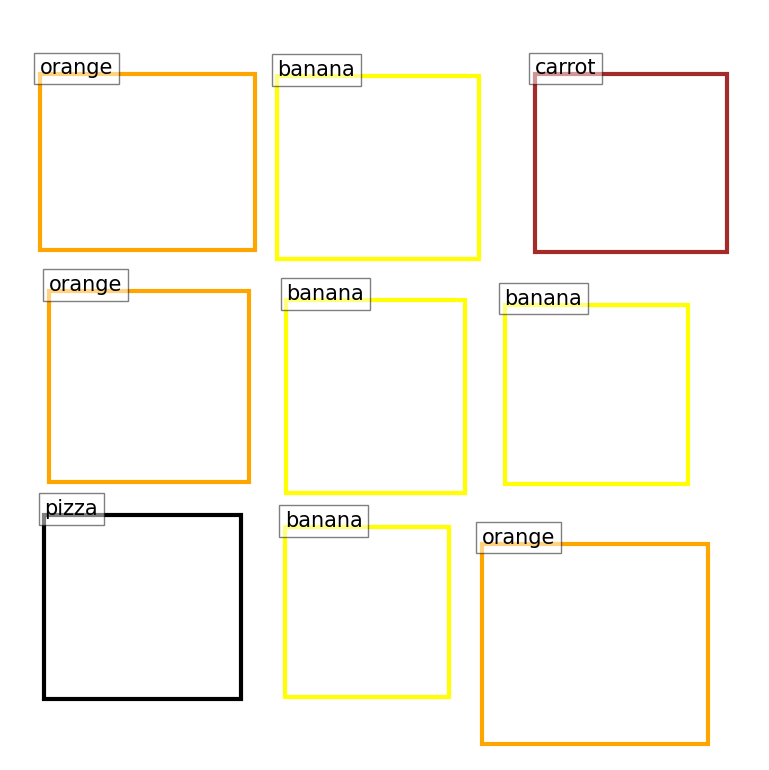}} &
          \adjustbox{valign=c}{\includegraphics[height=0.14\textwidth]{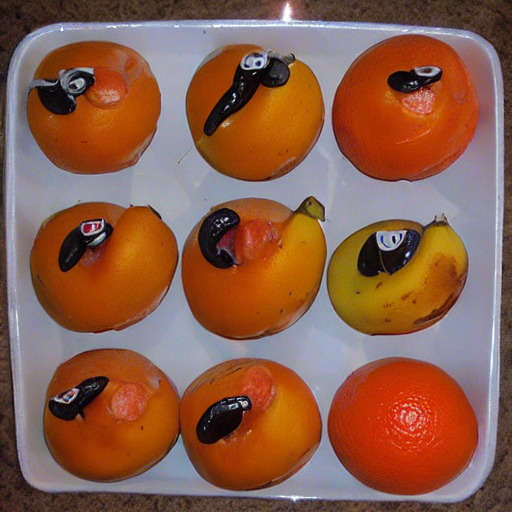}} &
          \adjustbox{valign=c}{\includegraphics[height=0.14\textwidth]{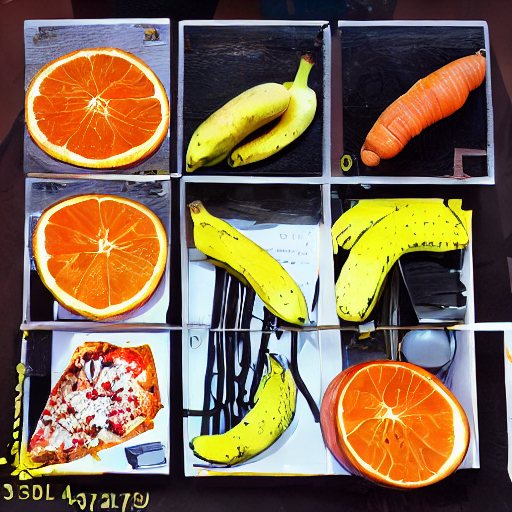}} \\

          \bottomrule
        \end{tabular}
    }
\end{center}
\caption{Image generation samples with custom layouts with COCO objects (out-of-distribution) from \method{} and \reco{}.
}
\label{tab:coco_image_samples_arbitrary}
\end{table*}

\section{\benchmarkreal{} Details}
\label{appendix_sec:layoutbench_coco_details}

\subsection{Dataset}

The \benchmarkreal{} (Sec. 5.5 in the main paper) benchmarks layout-guided image generation models in zero-shot fashion.
Following the design of \benchmark{}, \benchmarkreal{} measures four skills: number, position, size, and combination.
The idea of number/position/size skills is essentially the same as those splits of \benchmark{},
while we replace the shape skill of \benchmark{} with a new `combination' skill.
The combination skill measures whether a model can generate uncommon combinations of two objects, which is a specifically interesting evaluation scenario for zero-shot models; for \benchmark{} evaluation, the models finetuned on \clevr{} have seen many combinations of different objects, so the combination of different objects is a less interesting problem.

For each skill, we first define 2D bounding box layouts without specifying objects. Then, we compose COCO~\cite{Lin2014COCO} objects for each layout. This process ensures that the layouts have balanced distributions among objects.
In total, \benchmarkreal{} provides 2,120 layouts.
\begin{table*}[h]
\centering
      \resizebox{.8\linewidth}{!}{
      \begin{tabular}{l c}
          \toprule
          Skill 1: Number & 
          \adjustbox{valign=c}
          {\includegraphics[height=0.2\textwidth]{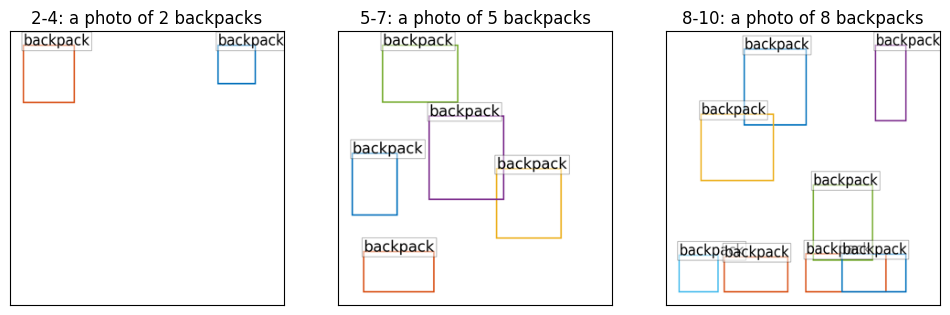}}
          \\
          
          \midrule

          Skill 2: Position & 
          \adjustbox{valign=c}
          {\includegraphics[height=0.2\textwidth]{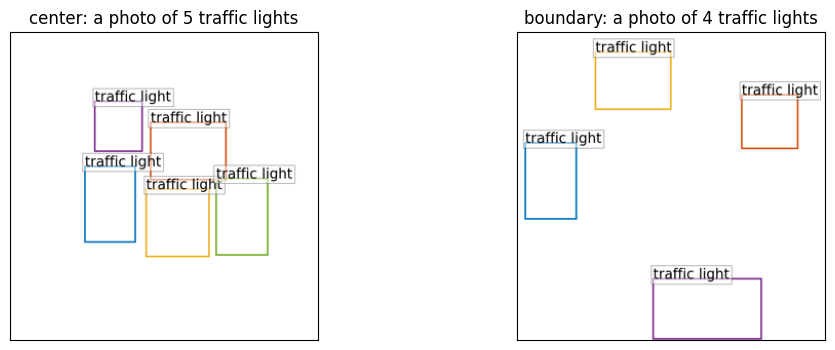}}
          \\

          \midrule

          Skill 3: Size & 
          \adjustbox{valign=c}
          {\includegraphics[height=0.2\textwidth]{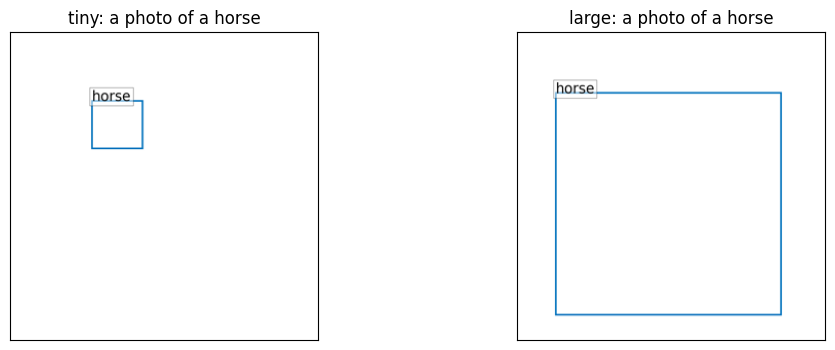}}
          \\
          \midrule

          Skill 4: Combination & 
          \adjustbox{valign=c}
          {\includegraphics[height=0.2\textwidth]{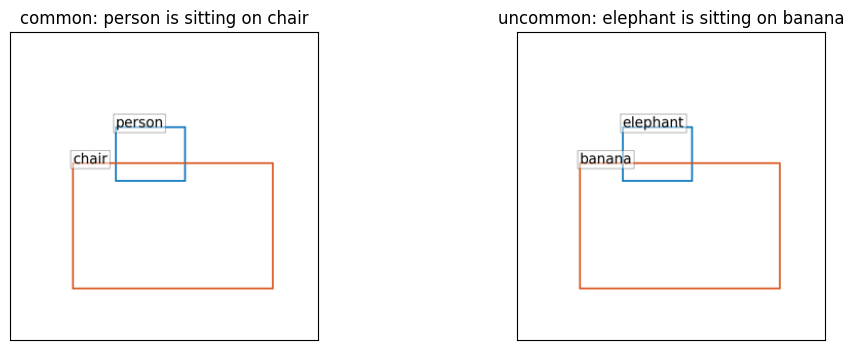}}
          \\

          \bottomrule
        \end{tabular}
        }
\caption{
Example layouts and captions of the four skills of \benchmarkreal{}.
}
\label{tab:layoutbench_coco_layouts}
\end{table*}
In \Cref{tab:layoutbench_coco_layouts}, we show example layouts for each skill.
In the following, we describe how we create the layouts for each skill in detail.

\noindent
\textbf{Skill 1: Number.} 
We define two layouts for 2$\sim$10 objects and use 40 COCO objects, resulting in 720 total layouts ($=2 \times 9 \times 40$).
We name the layouts with 2$\sim$4 and 8$\sim$10 objects as \textit{few} and \textit{many} splits.
The layouts are paired with captions with a template ``\texttt{a photo of [N] [objects]}''.

\noindent
\textbf{Skill 2: Position.} 
For each of \textit{boundary} and \textit{center} splits, we define four layouts with 40 COCO objects, resulting in 320 total layouts ($=2 \times 4 \times 40$).
The layouts are paired with captions with a template ``\texttt{a photo of [N] [objects]}''.

\noindent
\textbf{Skill 3: Size.}
For each of \textit{tiny} and \textit{large} splits, we define nine layouts with 40 COCO objects, resulting in 720 total layouts ($=9 \times 2 \times 40$).
The layouts are paired with captions with a template ``\texttt{a photo of [N] [objects]}''.

\noindent
\textbf{Skill 4: Combination.}
This skill measures whether a model can generate two objects that commonly or uncommonly appear in the real world.
For each of the three spatial relations (holding, next to, sitting on), 
we define three layouts without specifying objects.
For each of the three relations, we manually define 20 object pairs of COCO objects for \textit{common} and \textit{uncommon} splits.
For example, (1) `person sitting on chair' is more common than (2) `elephant sitting on banana' in real life.
This results in 360 total layouts ($=2 \times  3 \times 3 \times 20$).
The layouts are paired with captions with a template ``\texttt{[objA] [relation] [objB]}''.

\subsection{Qualitative Examples}

In \Cref{tab:layoutbench_coco_images_all_models_1} and \Cref{tab:layoutbench_coco_images_all_models_2}, we show sample images generated by different layout-guided image generation models from \benchmarkreal{} layouts.

\begin{table*}[h]
\centering
      \resizebox{\linewidth}{!}{
      \begin{tabular}{l c c c c c c c c}
          \toprule
          & 
          \multicolumn{2}{c} {Skill 1: Number} & \multicolumn{2}{c} {Skill 2: Position} & \multicolumn{2}{c} {Skill 3: Size} & \multicolumn{2}{c} {Skill 4: Combination}\\
          \cmidrule(lr){2-3} \cmidrule(lr){4-5} \cmidrule(lr){6-7} \cmidrule(lr){8-9}
          & few & many & center & boundary & tiny & large & common & uncommon \\
\midrule
Captions &
4 chairs &
10 cars &
5 buses &
5 suitcases &
3 cars &
3 broccolis &
{\tiny person is holding tennis racket} &
{\tiny parking meter is next to clock} \\

\midrule

\controlnet{} &
          \adjustbox{valign=c}{\includegraphics[height=0.1\textwidth]{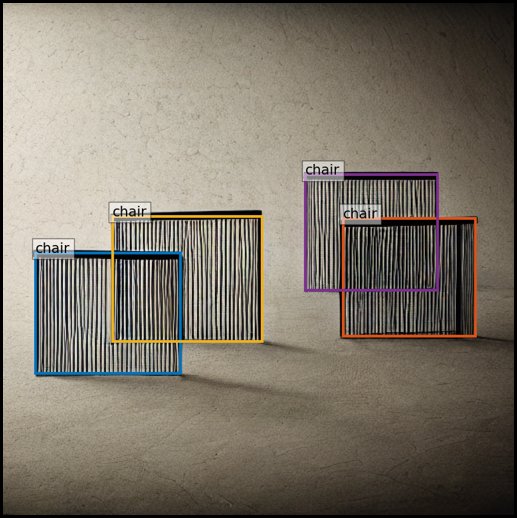}} &
          \adjustbox{valign=c}{\includegraphics[height=0.1\textwidth]{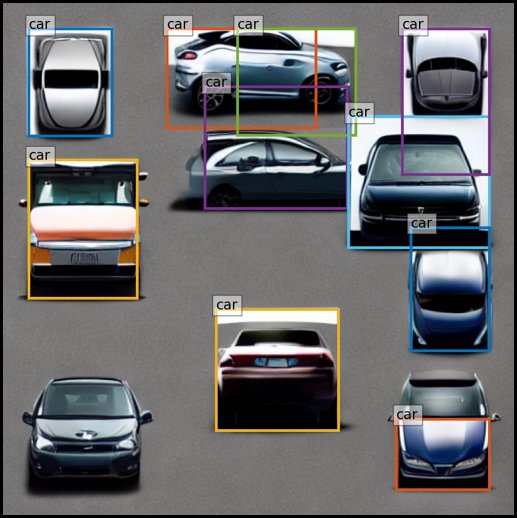}} &
          \adjustbox{valign=c}{\includegraphics[height=0.1\textwidth]{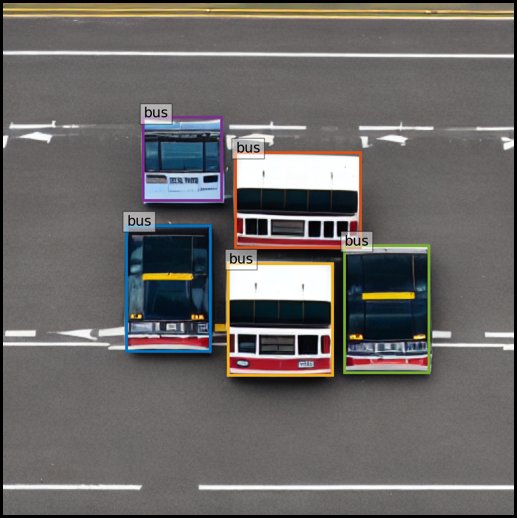}} &
          \adjustbox{valign=c}{\includegraphics[height=0.1\textwidth]{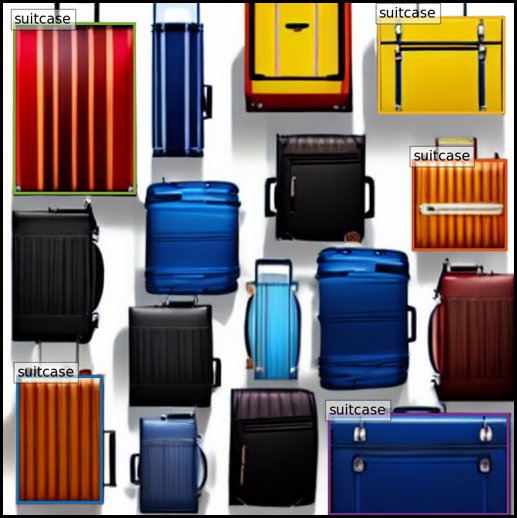}} &
          \adjustbox{valign=c}{\includegraphics[height=0.1\textwidth]{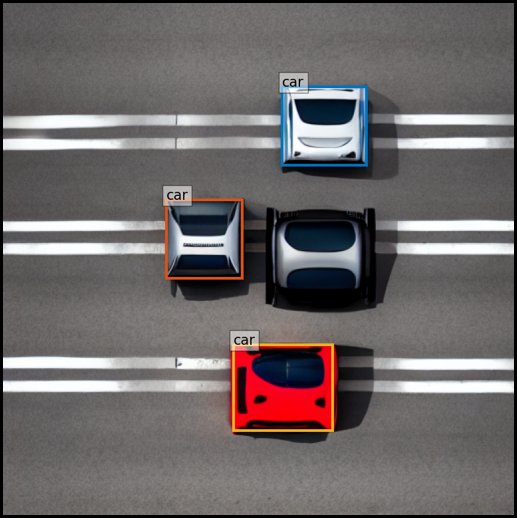}} &
          \adjustbox{valign=c}{\includegraphics[height=0.1\textwidth]{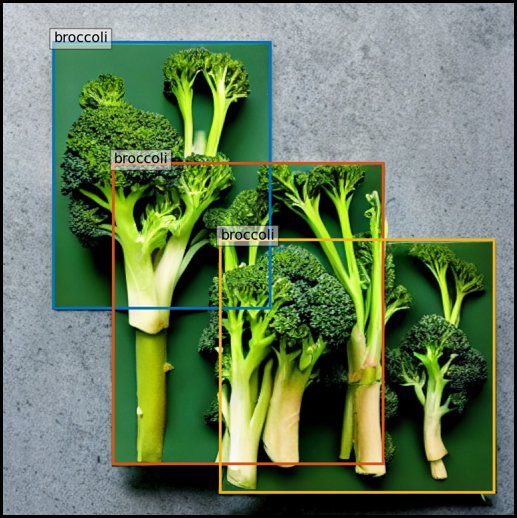}} &
          \adjustbox{valign=c}{\includegraphics[height=0.1\textwidth]{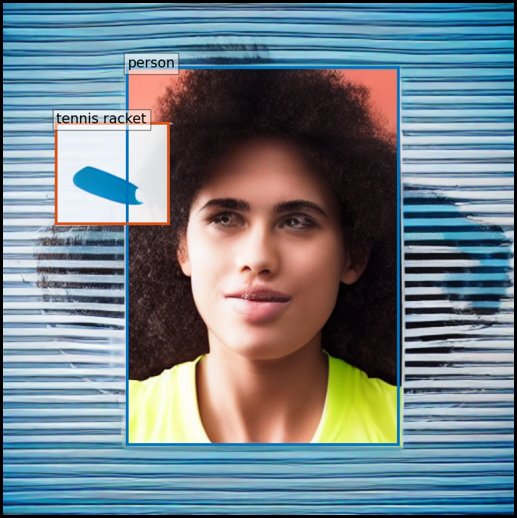}} &
          \adjustbox{valign=c}{\includegraphics[height=0.1\textwidth]{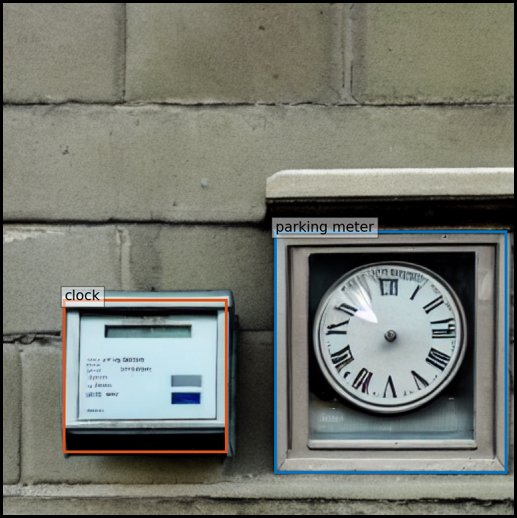}}
          \\

\midrule

\gligen{} &
          \adjustbox{valign=c}{\includegraphics[height=0.1\textwidth]{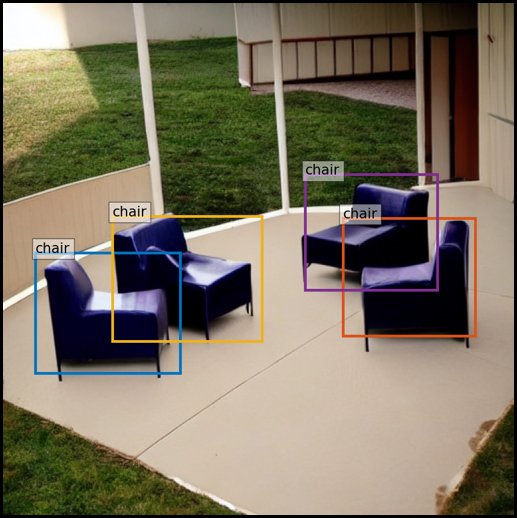}} &
          \adjustbox{valign=c}{\includegraphics[height=0.1\textwidth]{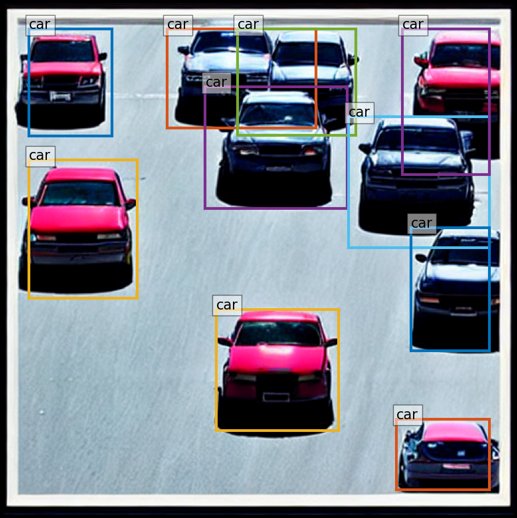}} &
          \adjustbox{valign=c}{\includegraphics[height=0.1\textwidth]{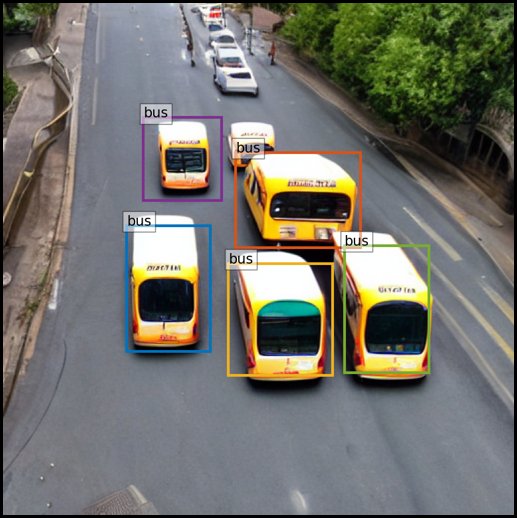}} &
          \adjustbox{valign=c}{\includegraphics[height=0.1\textwidth]{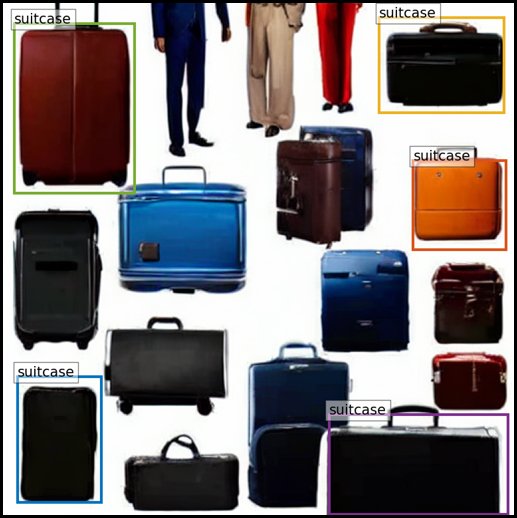}} &
          \adjustbox{valign=c}{\includegraphics[height=0.1\textwidth]{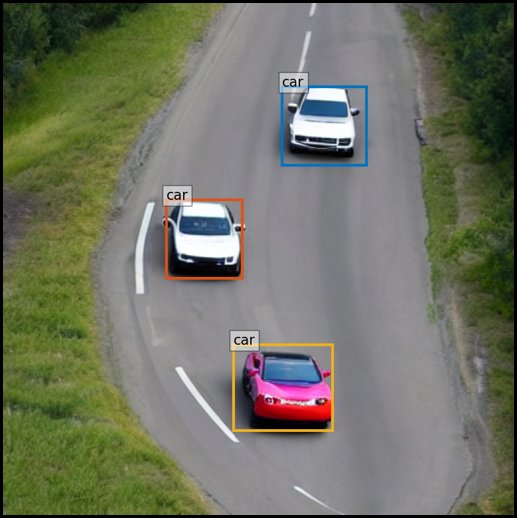}} &
          \adjustbox{valign=c}{\includegraphics[height=0.1\textwidth]{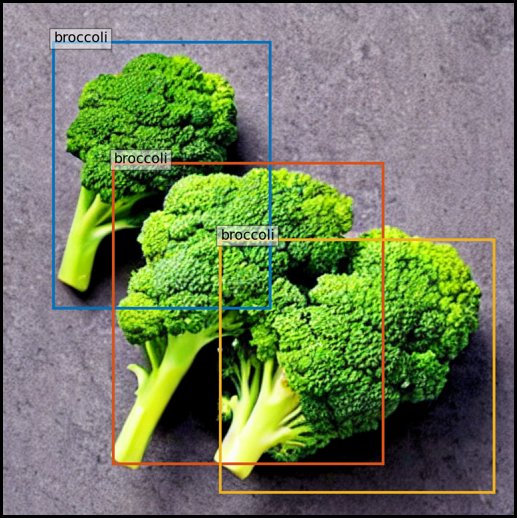}} &
          \adjustbox{valign=c}{\includegraphics[height=0.1\textwidth]{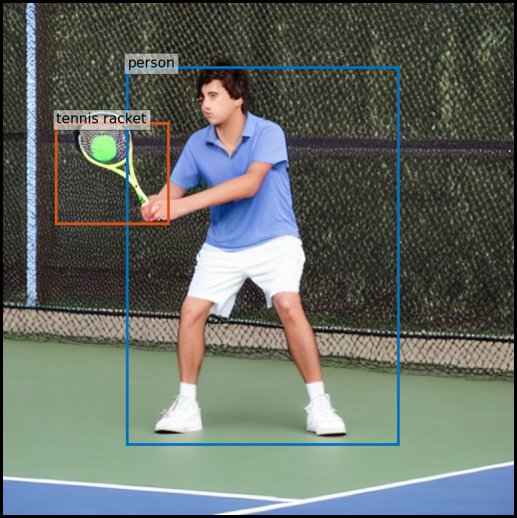}} &
          \adjustbox{valign=c}{\includegraphics[height=0.1\textwidth]{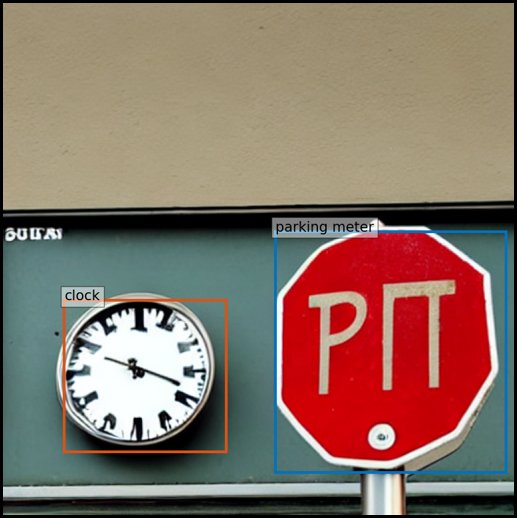}}
          \\

\midrule

\reco{} &
          \adjustbox{valign=c}{\includegraphics[height=0.1\textwidth]{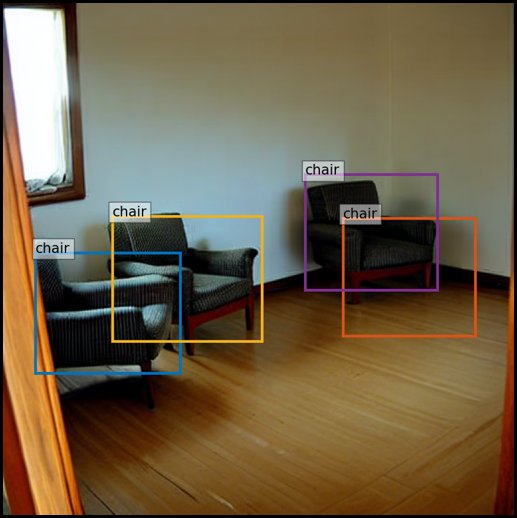}} &
          \adjustbox{valign=c}{\includegraphics[height=0.1\textwidth]{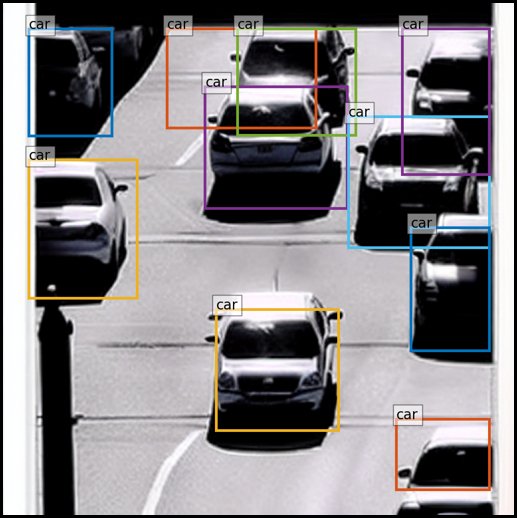}} &
          \adjustbox{valign=c}{\includegraphics[height=0.1\textwidth]{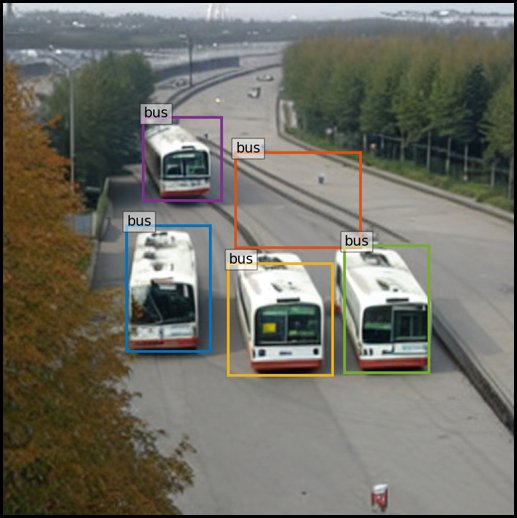}} &
          \adjustbox{valign=c}{\includegraphics[height=0.1\textwidth]{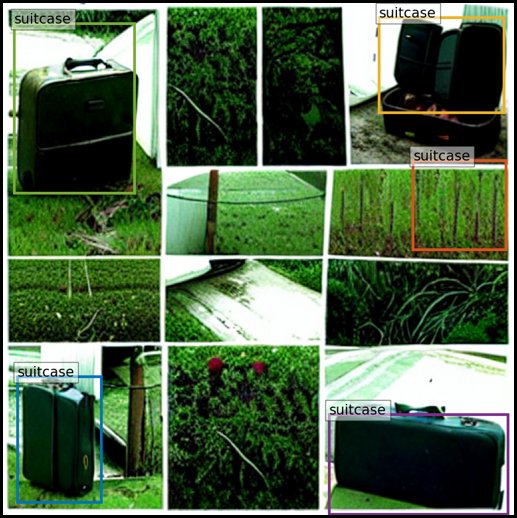}} &
          \adjustbox{valign=c}{\includegraphics[height=0.1\textwidth]{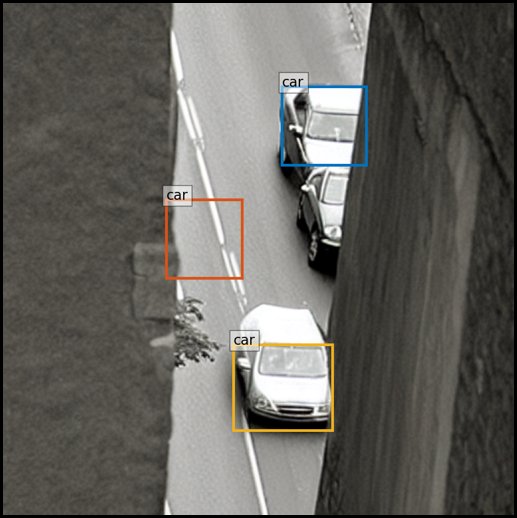}} &
          \adjustbox{valign=c}{\includegraphics[height=0.1\textwidth]{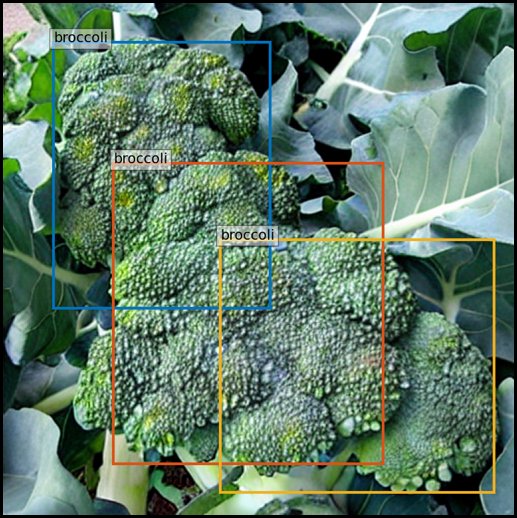}} &
          \adjustbox{valign=c}{\includegraphics[height=0.1\textwidth]{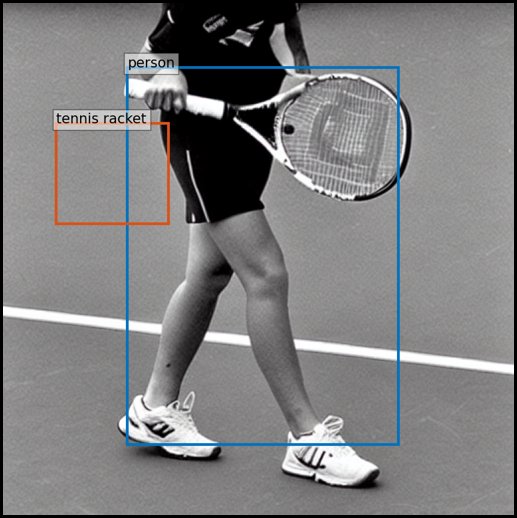}} &
          \adjustbox{valign=c}{\includegraphics[height=0.1\textwidth]{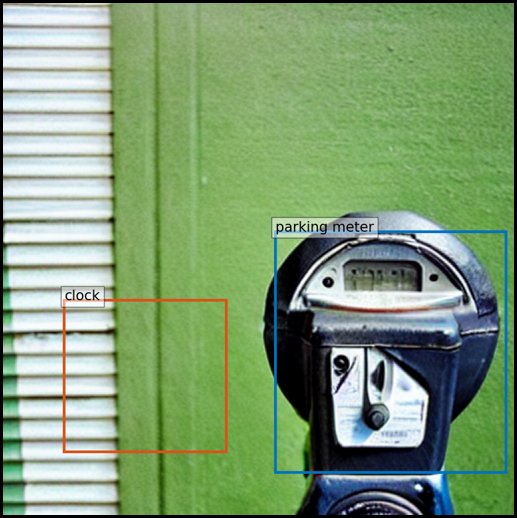}}
          \\

\midrule

\method{} &
          \adjustbox{valign=c}{\includegraphics[height=0.1\textwidth]{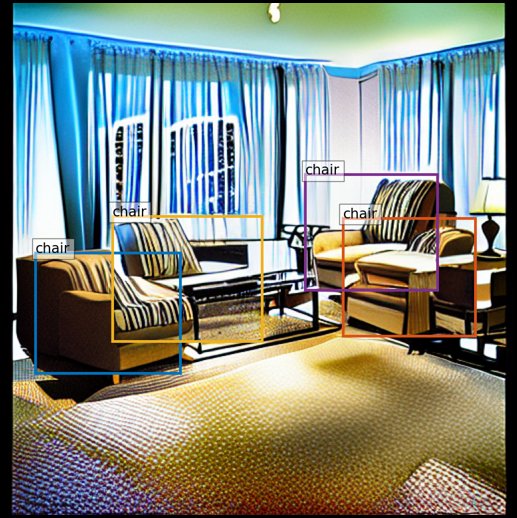}} &
          \adjustbox{valign=c}{\includegraphics[height=0.1\textwidth]{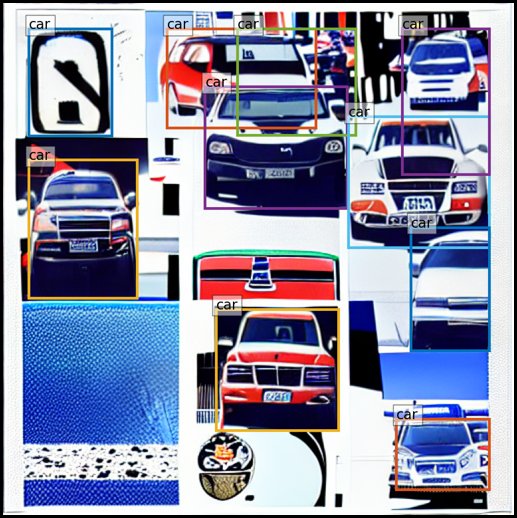}} &
          \adjustbox{valign=c}{\includegraphics[height=0.1\textwidth]{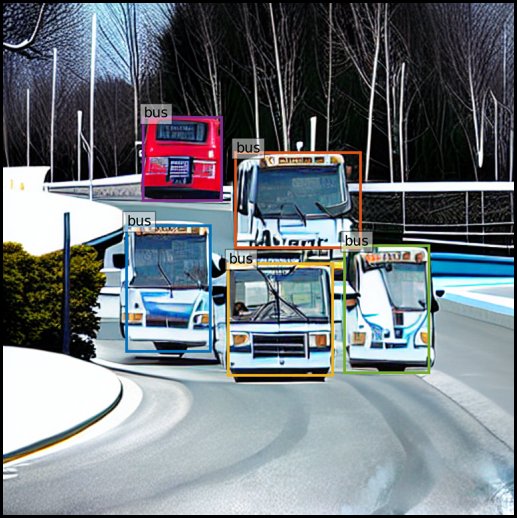}} &
          \adjustbox{valign=c}{\includegraphics[height=0.1\textwidth]{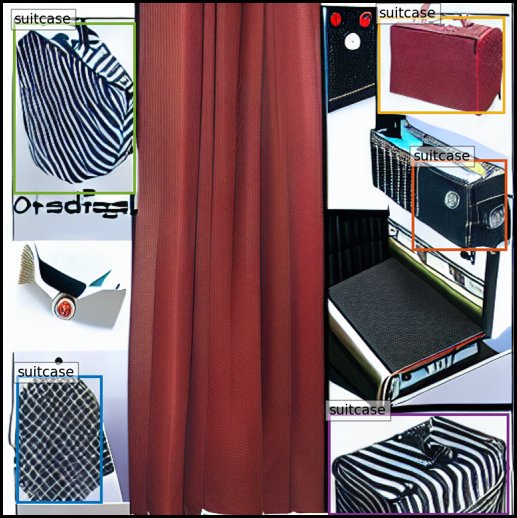}} &
          \adjustbox{valign=c}{\includegraphics[height=0.1\textwidth]{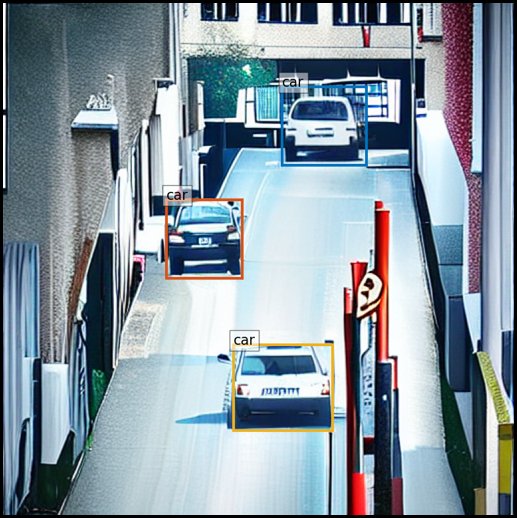}} &
          \adjustbox{valign=c}{\includegraphics[height=0.1\textwidth]{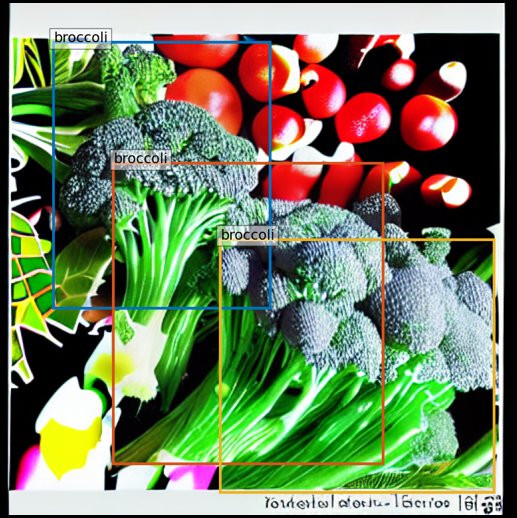}} &
          \adjustbox{valign=c}{\includegraphics[height=0.1\textwidth]{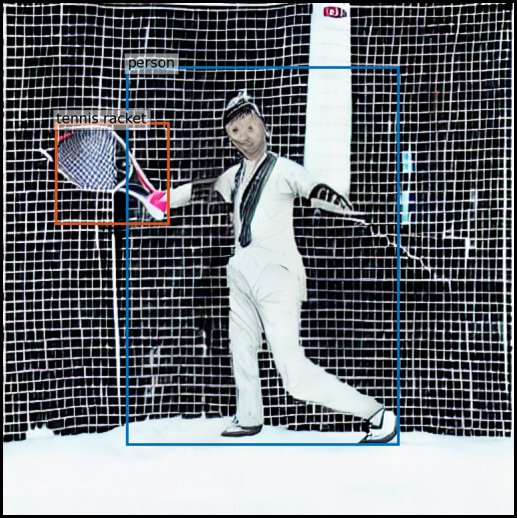}} &
          \adjustbox{valign=c}{\includegraphics[height=0.1\textwidth]{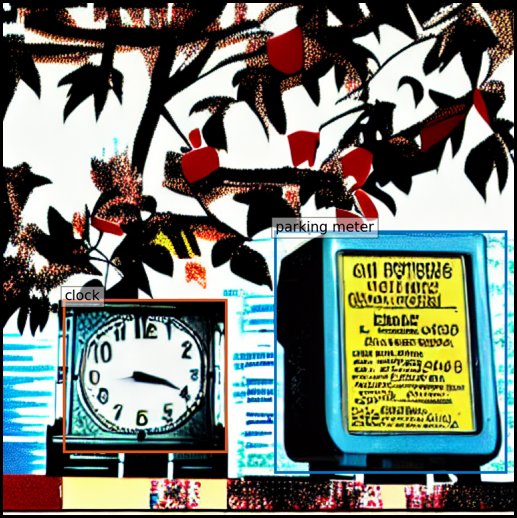}}
          \\
          
          \bottomrule
        \end{tabular}
        }
\caption{
Example images generated by four different methods
given four splits of caption and layouts from \benchmarkreal{}.
For number/position/size skills, the captions have the prefix `a photo of' before the `\texttt{[N] [objects]}' text.
}
\label{tab:layoutbench_coco_images_all_models_1}
\end{table*}
\begin{table*}[h]
\centering
      \resizebox{\linewidth}{!}{
      \begin{tabular}{l c c c c c c c c}
          \toprule
          & 
          \multicolumn{2}{c} {Skill 1: Number} & \multicolumn{2}{c} {Skill 2: Position} & \multicolumn{2}{c} {Skill 3: Size} & \multicolumn{2}{c} {Skill 4: Combination}\\
          \cmidrule(lr){2-3} \cmidrule(lr){4-5} \cmidrule(lr){6-7} \cmidrule(lr){8-9}
          & few & many & center & boundary & tiny & large & common & uncommon \\

\midrule
Captions &
4 suitcases &
9 broccolis &
5 motorcycles &
4 benches &
3 umbrellas &
a surfboard &
{\tiny handbag is holding cell phone} &
{\tiny fire hydrant is next to bed} \\

\midrule

\controlnet{} &
          \adjustbox{valign=c}{\includegraphics[height=0.1\textwidth]{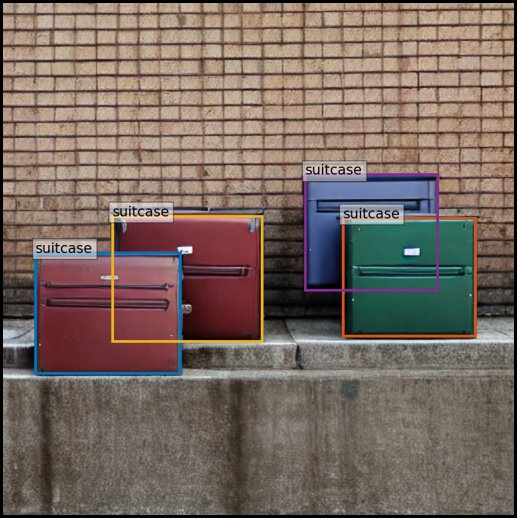}} &
          \adjustbox{valign=c}{\includegraphics[height=0.1\textwidth]{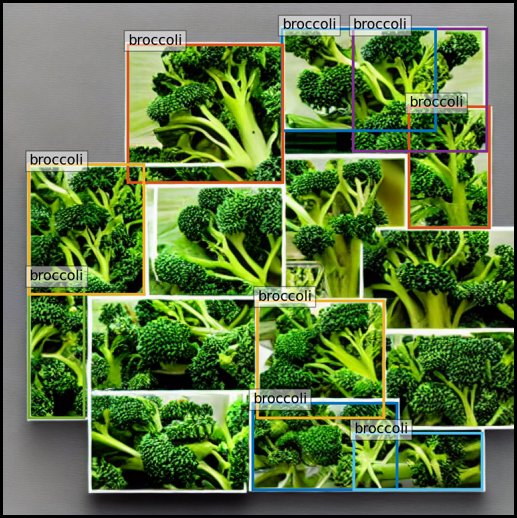}} &
          \adjustbox{valign=c}{\includegraphics[height=0.1\textwidth]{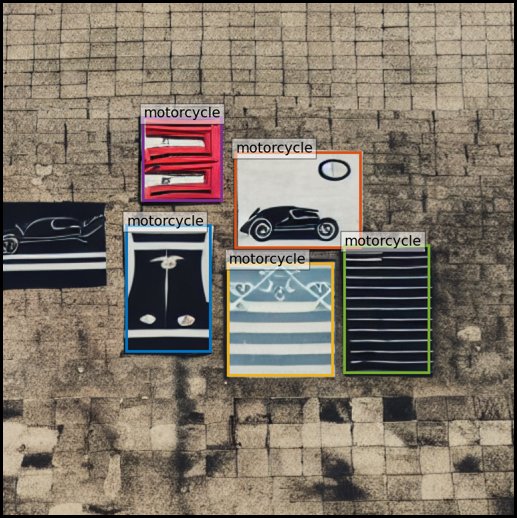}} &
          \adjustbox{valign=c}{\includegraphics[height=0.1\textwidth]{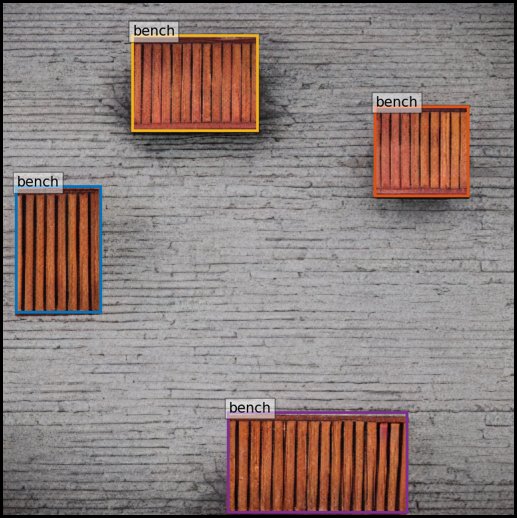}} &
          \adjustbox{valign=c}{\includegraphics[height=0.1\textwidth]{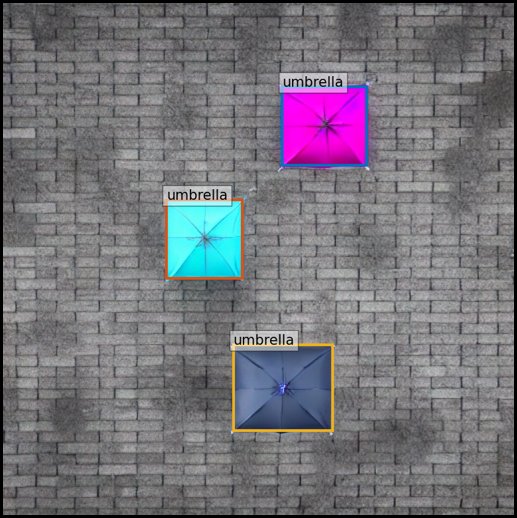}} &
          \adjustbox{valign=c}{\includegraphics[height=0.1\textwidth]{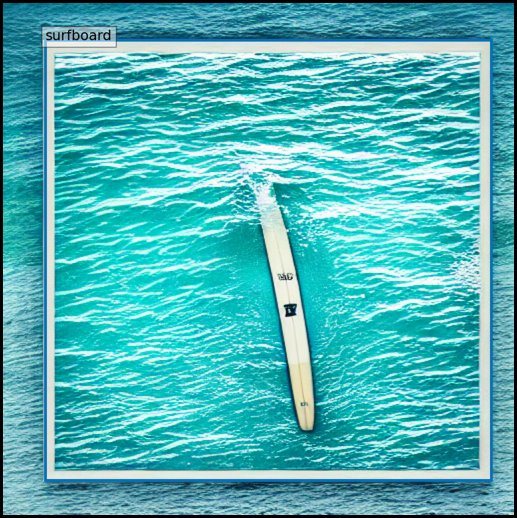}} &
          \adjustbox{valign=c}{\includegraphics[height=0.1\textwidth]{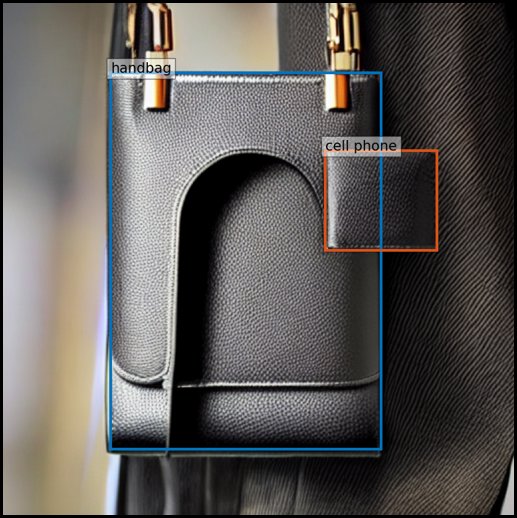}} &
          \adjustbox{valign=c}{\includegraphics[height=0.1\textwidth]{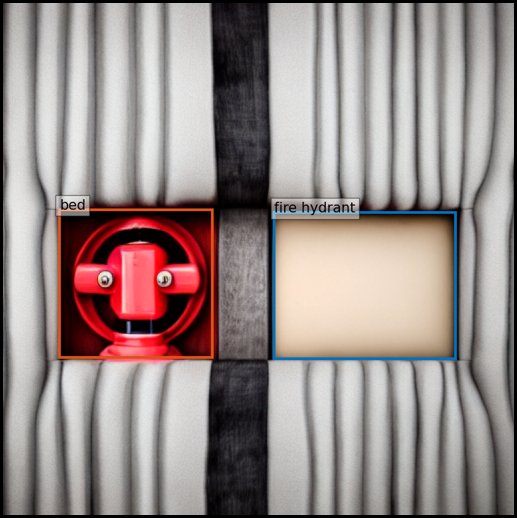}}
          \\

\midrule

\gligen{} &
          \adjustbox{valign=c}{\includegraphics[height=0.1\textwidth]{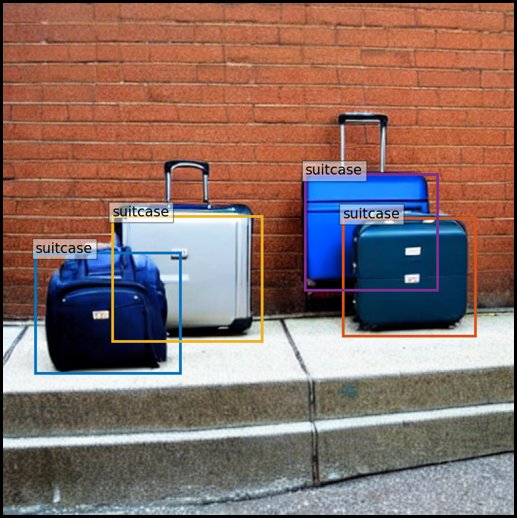}} &
          \adjustbox{valign=c}{\includegraphics[height=0.1\textwidth]{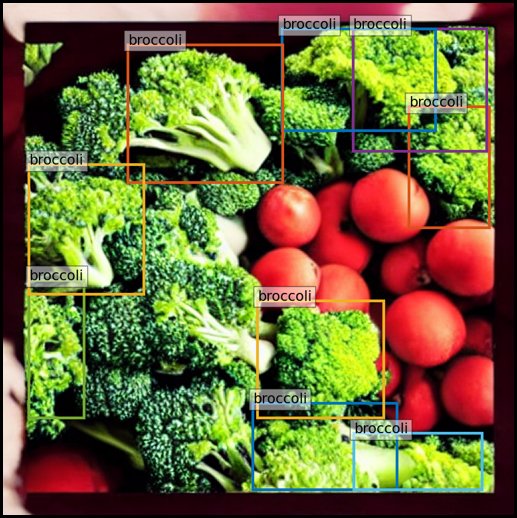}} &
          \adjustbox{valign=c}{\includegraphics[height=0.1\textwidth]{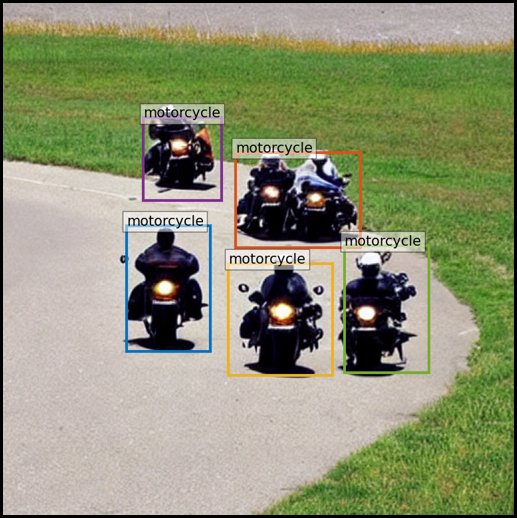}} &
          \adjustbox{valign=c}{\includegraphics[height=0.1\textwidth]{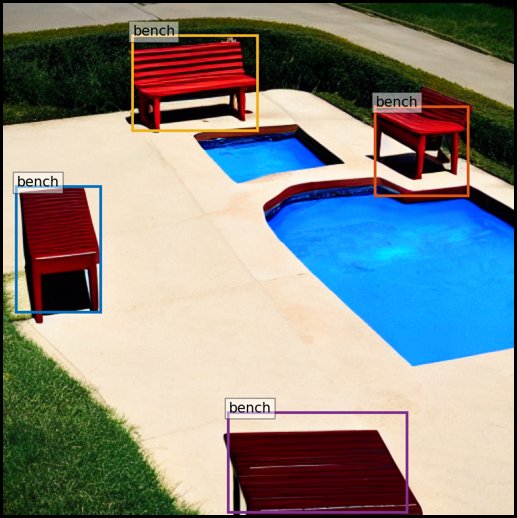}} &
          \adjustbox{valign=c}{\includegraphics[height=0.1\textwidth]{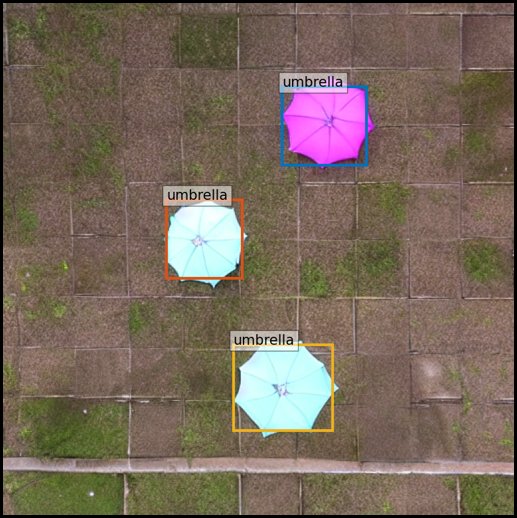}} &
          \adjustbox{valign=c}{\includegraphics[height=0.1\textwidth]{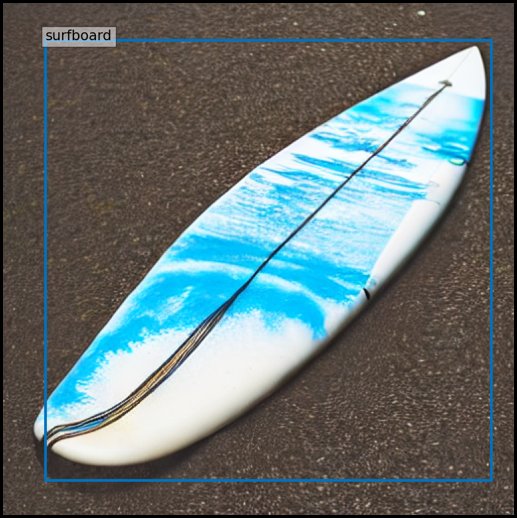}} &
          \adjustbox{valign=c}{\includegraphics[height=0.1\textwidth]{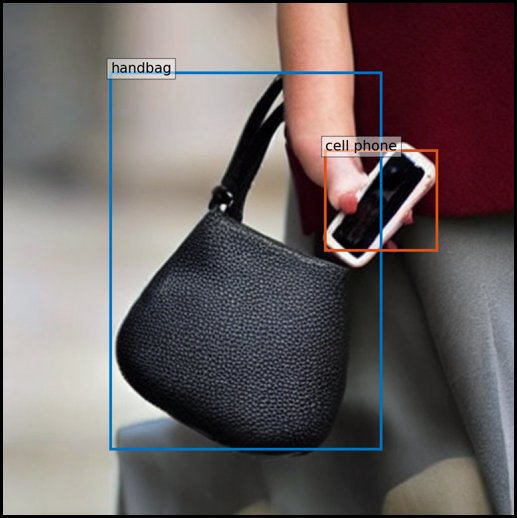}} &
          \adjustbox{valign=c}{\includegraphics[height=0.1\textwidth]{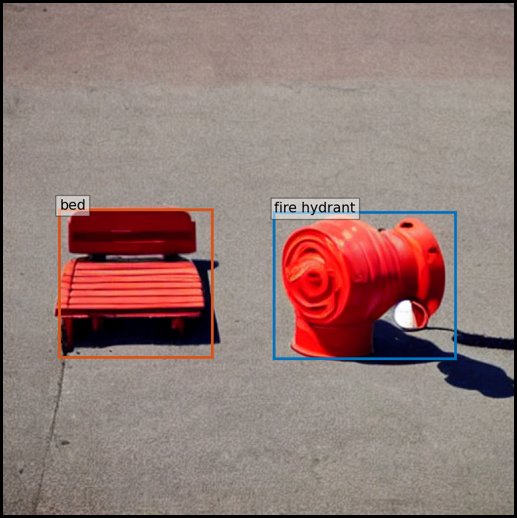}}
          \\

\midrule

\reco{} &
          \adjustbox{valign=c}{\includegraphics[height=0.1\textwidth]{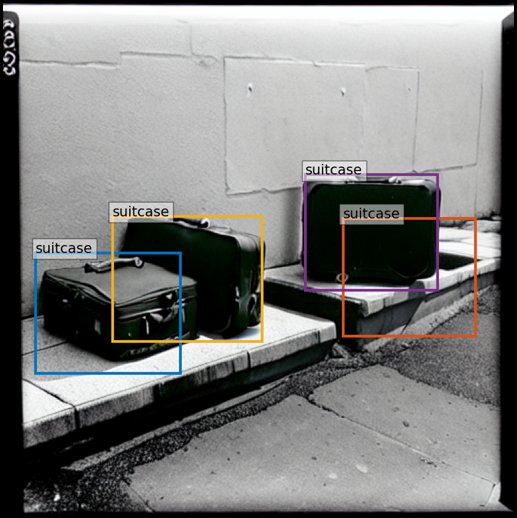}} &
          \adjustbox{valign=c}{\includegraphics[height=0.1\textwidth]{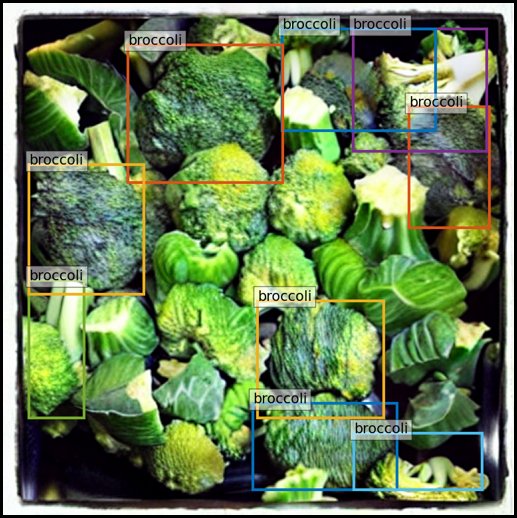}} &
          \adjustbox{valign=c}{\includegraphics[height=0.1\textwidth]{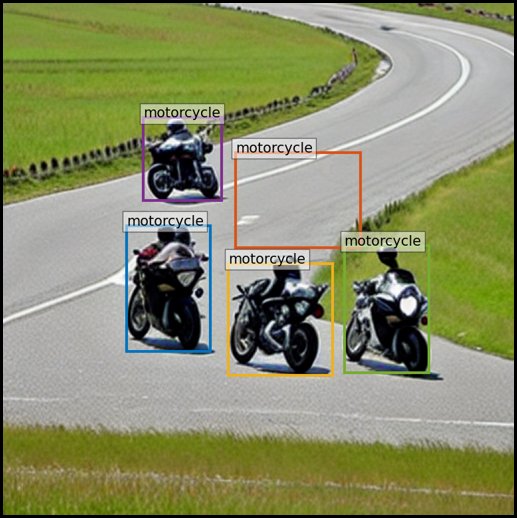}} &
          \adjustbox{valign=c}{\includegraphics[height=0.1\textwidth]{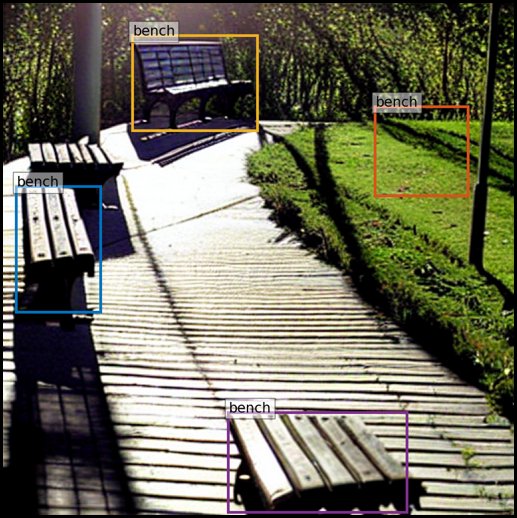}} &
          \adjustbox{valign=c}{\includegraphics[height=0.1\textwidth]{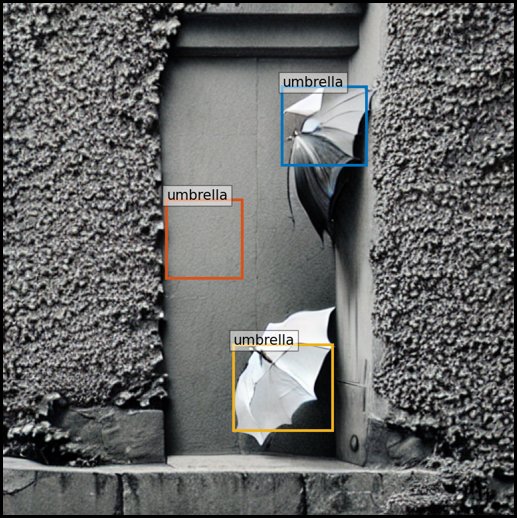}} &
          \adjustbox{valign=c}{\includegraphics[height=0.1\textwidth]{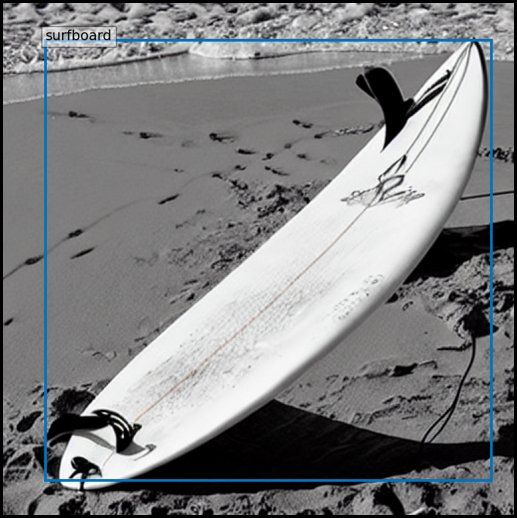}} &
          \adjustbox{valign=c}{\includegraphics[height=0.1\textwidth]{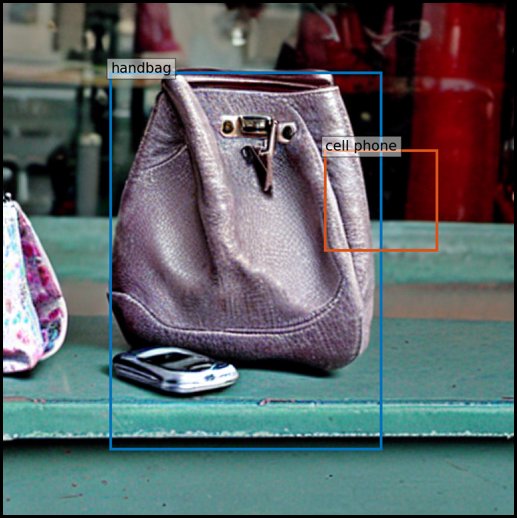}} &
          \adjustbox{valign=c}{\includegraphics[height=0.1\textwidth]{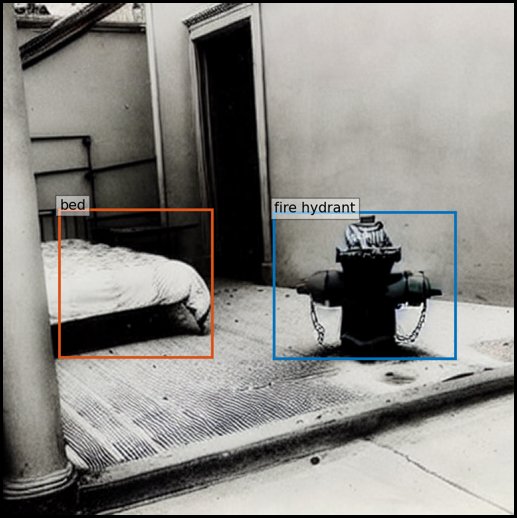}}
\\
\midrule

\method{} &
          \adjustbox{valign=c}{\includegraphics[height=0.1\textwidth]{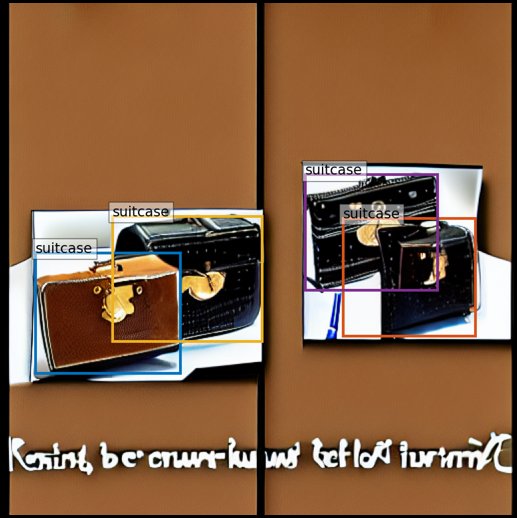}} &
          \adjustbox{valign=c}{\includegraphics[height=0.1\textwidth]{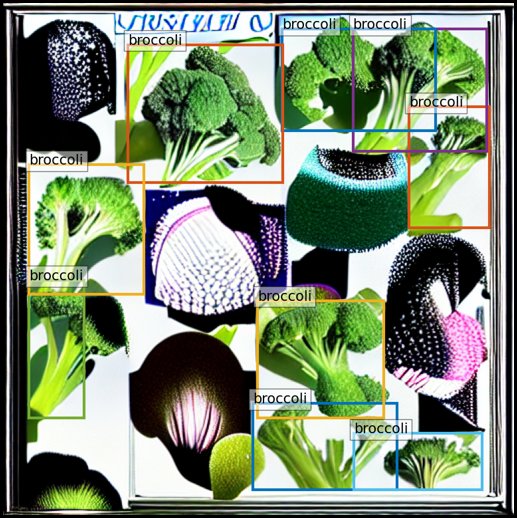}} &
          \adjustbox{valign=c}{\includegraphics[height=0.1\textwidth]{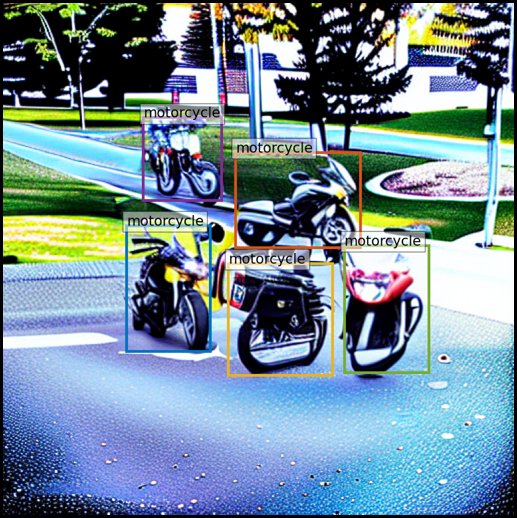}} &
          \adjustbox{valign=c}{\includegraphics[height=0.1\textwidth]{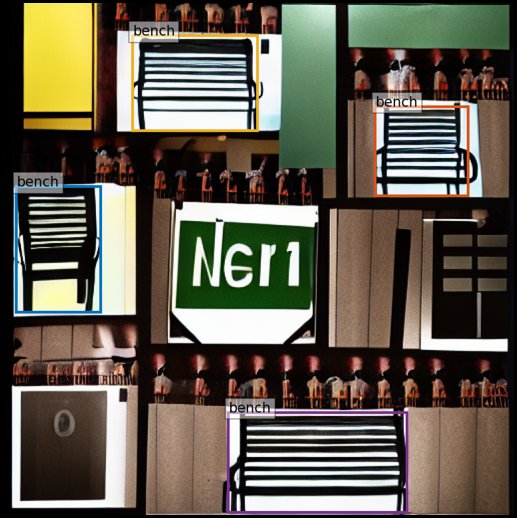}} &
          \adjustbox{valign=c}{\includegraphics[height=0.1\textwidth]{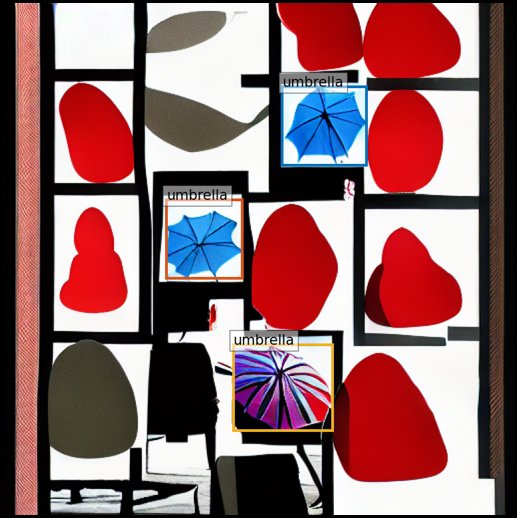}} &
          \adjustbox{valign=c}{\includegraphics[height=0.1\textwidth]{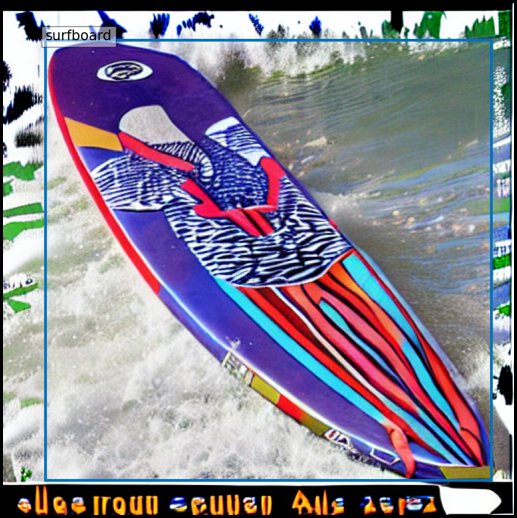}} &
          \adjustbox{valign=c}{\includegraphics[height=0.1\textwidth]{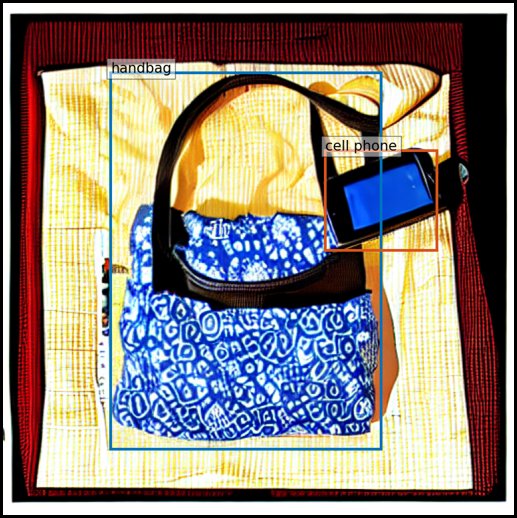}} &
          \adjustbox{valign=c}{\includegraphics[height=0.1\textwidth]{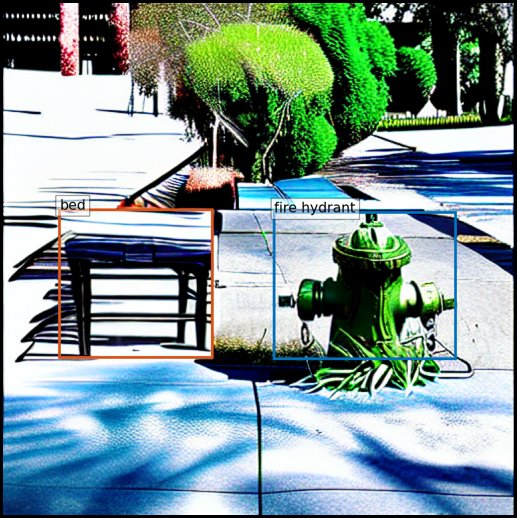}}

          \\

          \bottomrule
        \end{tabular}
        }
\caption{
Additional example images generated by four different methods
given four splits of caption and layouts from \benchmarkreal{}.
For number/position/size skills, the captions have the prefix `a photo of' before the `\texttt{[N] [objects]}' text.
}
\label{tab:layoutbench_coco_images_all_models_2}
\end{table*}

\begin{table*}[h]
\begin{center}
    \resizebox{\linewidth}{!}{
      \begin{tblr}{l l l l}
          \toprule

          \adjustbox{valign=c}{\includegraphics[height=0.14\textwidth]{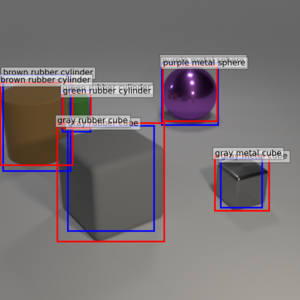}} &
          \adjustbox{valign=c}{\includegraphics[height=0.14\textwidth]{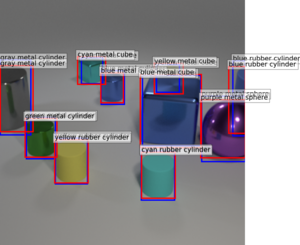}} &
          \adjustbox{valign=c}{\includegraphics[height=0.14\textwidth]{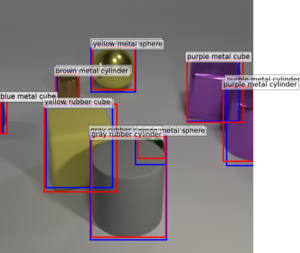}} &
          \adjustbox{valign=c}{\includegraphics[height=0.14\textwidth]{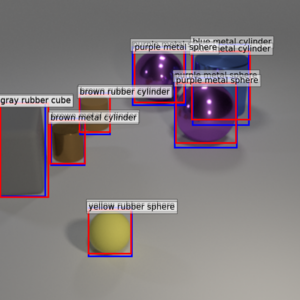}} \\
          \midrule
          \adjustbox{valign=c}{\includegraphics[height=0.14\textwidth]{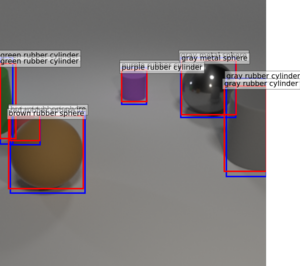}} &
          \adjustbox{valign=c}{\includegraphics[height=0.14\textwidth]{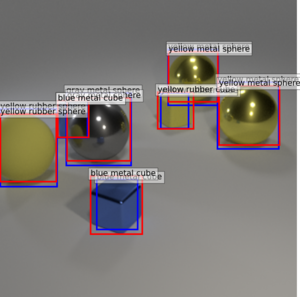}} &
          \adjustbox{valign=c}{\includegraphics[height=0.14\textwidth]{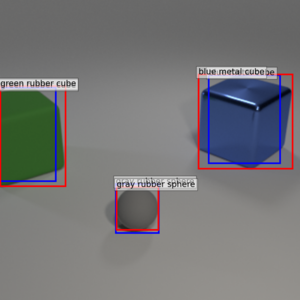}} &
          \adjustbox{valign=c}{\includegraphics[height=0.14\textwidth]{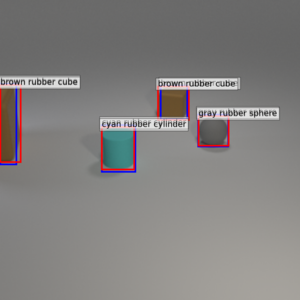}} \\
          
          \bottomrule
        \end{tblr}
    }
\end{center}
\caption{
\clevr{} images with (1) bounding box annotations provided by Krishna \etal{} (2018) \cite{Krishna2018ReferringRelationships} (colored in \textcolor{blue}{blue}) and (2) object detection results based on our \benchmark{}-trained DETR (colored in \textcolor{red}{red}).
}
\label{tab:clevr_obj_det}
\end{table*}

\section{\clevr{} GT Layout Accuracy}
\label{appendix_sec:obj_det_acc}

In Table 2, the AP on \clevr{} GT images (60.7) is not as high as that of \benchmark{} GT images (90.7).
This is because of the noise of the bounding box annotations provided by \cite{Krishna2018ReferringRelationships}.
\clevr{}~\cite{johnson2017clevr} images are collected by (1) sampling scene parameters such as object attributes, positions, lighting, and camera positions and (2) rendering images.
However, the public \clevr{} scene files do not contain the original bounding box coordinates and randomly jittered camera positions,
which makes it impossible to precisely reconstruct the original bounding box coordinates of \clevr{} images.\footnote{\url{https://github.com/facebookresearch/clevr-dataset-gen/blob/9742828c3667e81d5c381dbe1a0bcae4c1a7e89a/image\_generation/render\_images.py\#L270-L273}}
Krishna \etal{} \cite{Krishna2018ReferringRelationships} provides bounding box annotations for \clevr{} by approximating camera positions, but the annotations still have some errors.
In \Cref{tab:clevr_obj_det}, we compare (1) bounding box annotations provided by \cite{Krishna2018ReferringRelationships} (colored in \textcolor{blue}{blue}) and (2) object detection results based on our \benchmark{}-trained DETR (colored in \textcolor{red}{red}), where our DETR outputs boxes that bound the objects more tightly than the box annotations from \cite{Krishna2018ReferringRelationships}.
On the test set of these re-rendered \clevr{} images with precise bounding boxes, our DETR object detector could achieve 99\% AP.

\section{Additional GAN Baseline}
\label{appendix_sec:additional_GAN_baseline}

In addition to the denoising diffusion models (\ldm{} and \reco{}), we also experiment with Hintz \etal~\cite{hinz2019generating}, a GAN-based layout-guided image generation approach as our baseline on \clevr{} and \benchmark{} layout experiments.
As shown in \Cref{tab:gan_generation_samples}, while Hintz \etal{} tend to place objects in correct locations,
the objects are much blurrier than the other diffusion models \ldm{}, \reco{}, and \method{}.
The low image quality makes the objects very hard to recognize (\ie, we could not tell whether a generated object is a cylinder, a cube, or a sphere), making the model achieve much worse image quality (\eg, 180.9 FID on \clevr{}) and layout accuracy (\eg, 1.8 \% AP on \benchmark{}) metrics.

\begin{table*}[t]
\begin{center}
\tablestyle{4pt}{0.95}
    \resizebox{\linewidth}{!}{
      \begin{tabular}{l c c c c c c c c c}

          \toprule
           \multirow{3}{*}{{ Method}} & \clevr{} & \multicolumn{8}{c} {\benchmark{}} \\
          \cmidrule(lr){2-2} \cmidrule(lr){3-10}
           &\multirow{2}{*}{{val}} &  \multicolumn{2}{c}{Number}  & \multicolumn{2}{c}{Position}  & \multicolumn{2}{c}{Size} &  \multicolumn{2}{c}{Shape} \\
          \cmidrule(lr){3-4} \cmidrule(lr){5-6} \cmidrule(lr){7-8} \cmidrule(lr){9-10}
          & & few & many & center & boundary & tiny & large & horizontal & vertical \\
          \midrule
          GT &
          \adjustbox{valign=c}{\includegraphics[height=0.1\textwidth]{images/images_for_main_example_table/clevr/gt_512x512.png}} &
          \adjustbox{valign=c}{\includegraphics[height=0.1\textwidth]{images/images_for_main_example_table/number_few/gt_512x512.png}} &
          \adjustbox{valign=c}{\includegraphics[height=0.1\textwidth]{images/images_for_main_example_table/number_many/gt_512x512.png}} &
          \adjustbox{valign=c}{\includegraphics[height=0.1\textwidth]{images/images_for_main_example_table/position_center/gt_512x512.png}} &
          \adjustbox{valign=c}{\includegraphics[height=0.1\textwidth]{images/images_for_main_example_table/position_boundary/gt_512x512.png}} &
          \adjustbox{valign=c}{\includegraphics[height=0.1\textwidth]{images/images_for_main_example_table/size_tiny/gt_512x512.png}} &
          \adjustbox{valign=c}{\includegraphics[height=0.1\textwidth]{images/images_for_main_example_table/size_large/gt_512x512.png}} &
          \adjustbox{valign=c}{\includegraphics[height=0.1\textwidth]{images/images_for_main_example_table/shape_horizontal/gt_512x512.png}} &
          \adjustbox{valign=c}{\includegraphics[height=0.1\textwidth]{images/images_for_main_example_table/shape_vertical/gt_512x512.png}} \\
          \midrule
          
          \specialcelll{Hintz \etal~\cite{hinz2019generating}} &
          \adjustbox{valign=c}{\includegraphics[height=0.1\textwidth]{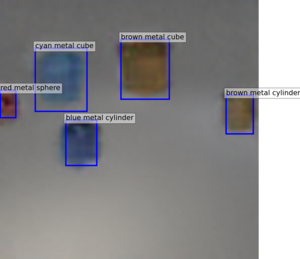}} &
          \adjustbox{valign=c}{\includegraphics[height=0.1\textwidth]{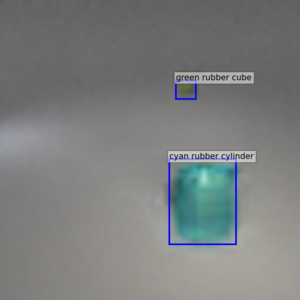}} &
          \adjustbox{valign=c}{\includegraphics[height=0.1\textwidth]{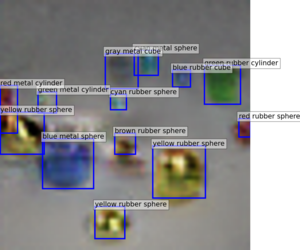}} &
          \adjustbox{valign=c}{\includegraphics[height=0.1\textwidth]{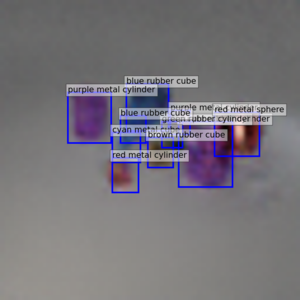}} &
          \adjustbox{valign=c}{\includegraphics[height=0.1\textwidth]{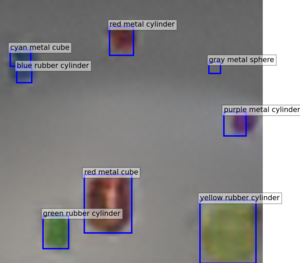}} &
          \adjustbox{valign=c}{\includegraphics[height=0.1\textwidth]{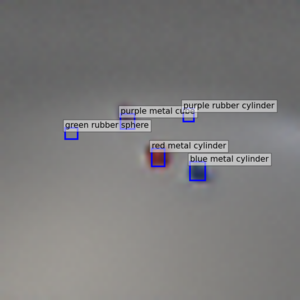}} &
          \adjustbox{valign=c}{\includegraphics[height=0.1\textwidth]{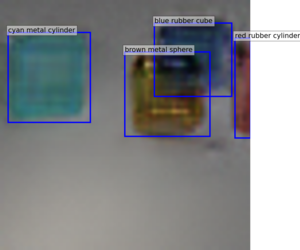}} &
          \adjustbox{valign=c}{\includegraphics[height=0.1\textwidth]{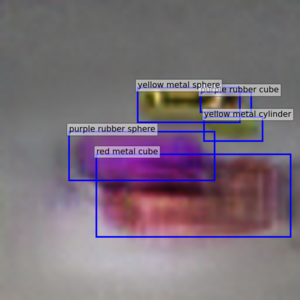}} &
          \adjustbox{valign=c}{\includegraphics[height=0.1\textwidth]{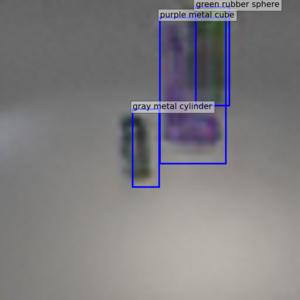}} \\
          \bottomrule
        \end{tabular}
    }
\end{center}
\caption{
Images generated by Hintz \etal~\cite{hinz2019generating} on \clevr{} (ID) and \benchmark{} (OOD) layouts. GT boxes are shown in \textcolor{blue}{blue}. 
}
\label{tab:gan_generation_samples}
\end{table*}

\section{Additional Image Generation Samples on \benchmark{}}
\label{appendix_sec:additional_image_samples}

In the following
\Cref{tab:additional_layoutbench_number_few},
\Cref{tab:additional_layoutbench_number_many},
\Cref{tab:additional_layoutbench_position_boundary},
\Cref{tab:additional_layoutbench_position_center},
\Cref{tab:additional_layoutbench_size_tiny},
\Cref{tab:additional_layoutbench_size_large},
\Cref{tab:additional_layoutbench_shape_horizontal},
and
\Cref{tab:additional_layoutbench_shape_vertical},
we show additional image samples of \benchmark{} and model generation results.

\begin{table*}[t]
    \begin{center}
        \resizebox{.75\linewidth}{!}{
          \begin{tblr}{l l l l l}
              \toprule
                
              GT &
              \adjustbox{valign=c}{\includegraphics[height=0.14\textwidth]{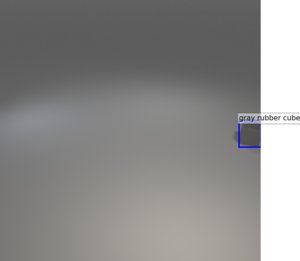}} &
              \adjustbox{valign=c}{\includegraphics[height=0.14\textwidth]{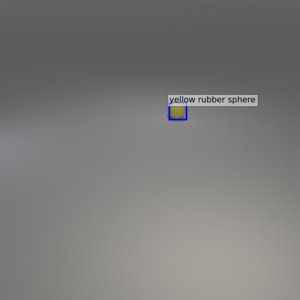}} &
              \adjustbox{valign=c}{\includegraphics[height=0.14\textwidth]{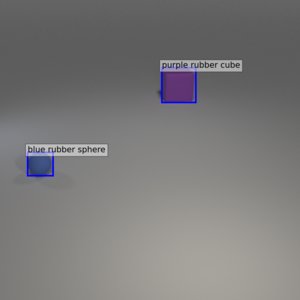}} &
              \adjustbox{valign=c}{\includegraphics[height=0.14\textwidth]{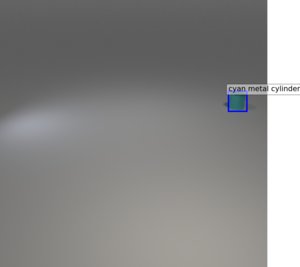}} \\
              
              \ldm{} &
              \adjustbox{valign=c}{\includegraphics[height=0.14\textwidth]{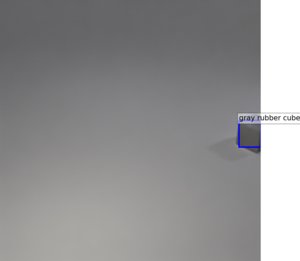}} &
              \adjustbox{valign=c}{\includegraphics[height=0.14\textwidth]{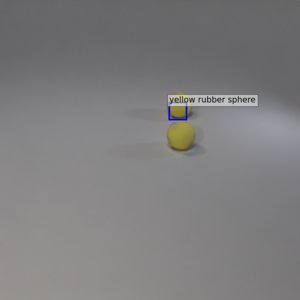}} &
              \adjustbox{valign=c}{\includegraphics[height=0.14\textwidth]{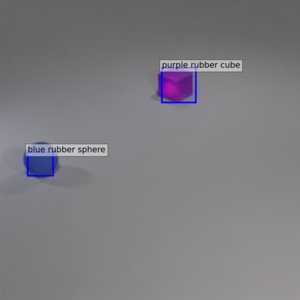}} &
              \adjustbox{valign=c}{\includegraphics[height=0.14\textwidth]{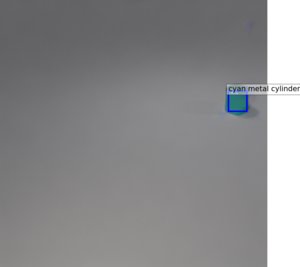}} \\
  
              \reco{} &
              \adjustbox{valign=c}{\includegraphics[height=0.14\textwidth]{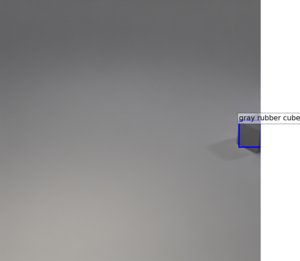}} &
              \adjustbox{valign=c}{\includegraphics[height=0.14\textwidth]{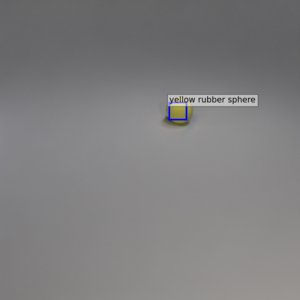}} &
              \adjustbox{valign=c}{\includegraphics[height=0.14\textwidth]{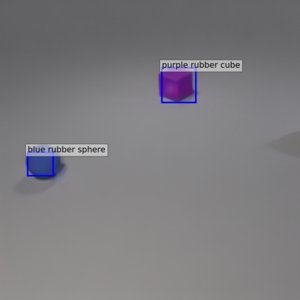}} &
              \adjustbox{valign=c}{\includegraphics[height=0.14\textwidth]{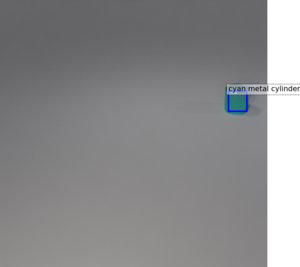}} \\
  
              {\method{} \\ (Ours)} & 
              \adjustbox{valign=c}{\includegraphics[height=0.14\textwidth]{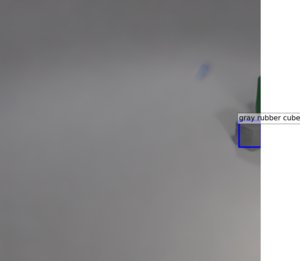}} &
              \adjustbox{valign=c}{\includegraphics[height=0.14\textwidth]{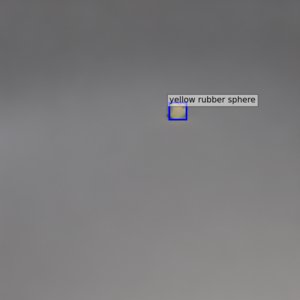}} &
              \adjustbox{valign=c}{\includegraphics[height=0.14\textwidth]{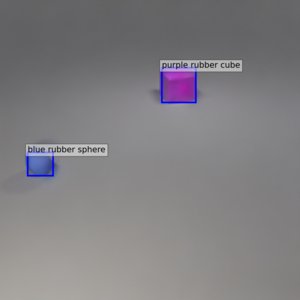}} &
              \adjustbox{valign=c}{\includegraphics[height=0.14\textwidth]{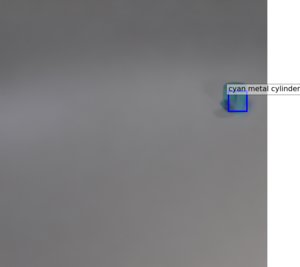}} \\
  
              \midrule
  
              GT &
              \adjustbox{valign=c}{\includegraphics[height=0.14\textwidth]{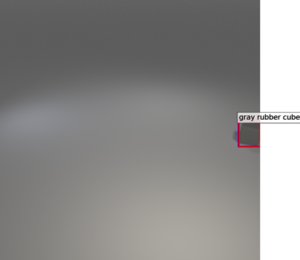}} &
              \adjustbox{valign=c}{\includegraphics[height=0.14\textwidth]{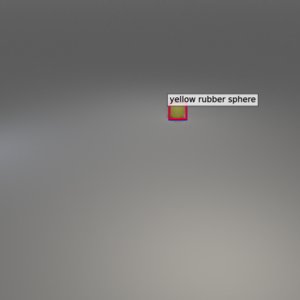}} &
              \adjustbox{valign=c}{\includegraphics[height=0.14\textwidth]{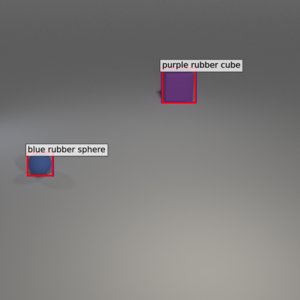}} &
              \adjustbox{valign=c}{\includegraphics[height=0.14\textwidth]{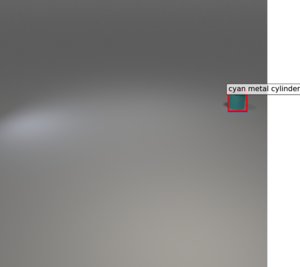}} \\
  
              \ldm{} &
              \adjustbox{valign=c}{\includegraphics[height=0.14\textwidth]{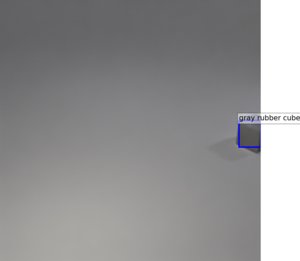}} &
              \adjustbox{valign=c}{\includegraphics[height=0.14\textwidth]{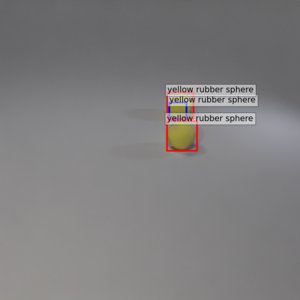}} &
              \adjustbox{valign=c}{\includegraphics[height=0.14\textwidth]{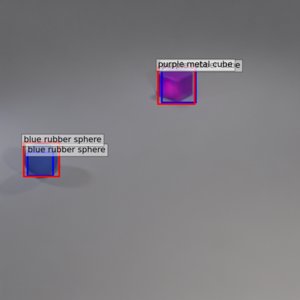}} &
              \adjustbox{valign=c}{\includegraphics[height=0.14\textwidth]{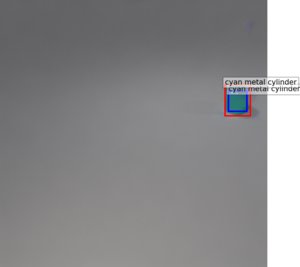}} \\
  
              \reco{} &
              \adjustbox{valign=c}{\includegraphics[height=0.14\textwidth]{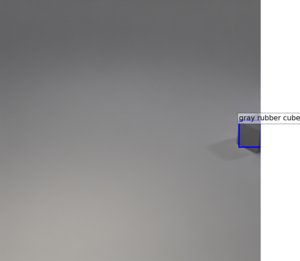}} &
              \adjustbox{valign=c}{\includegraphics[height=0.14\textwidth]{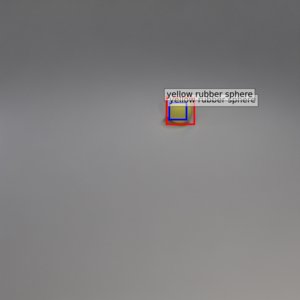}} &
              \adjustbox{valign=c}{\includegraphics[height=0.14\textwidth]{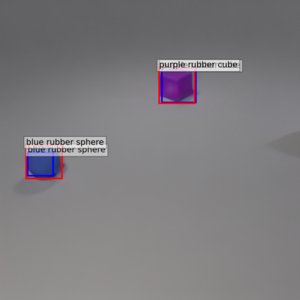}} &
              \adjustbox{valign=c}{\includegraphics[height=0.14\textwidth]{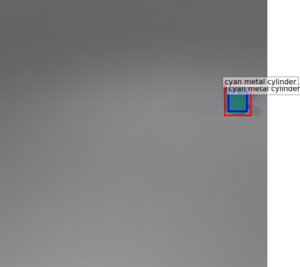}} \\
  
              {\method{} \\ (Ours)} &
              \adjustbox{valign=c}{\includegraphics[height=0.14\textwidth]{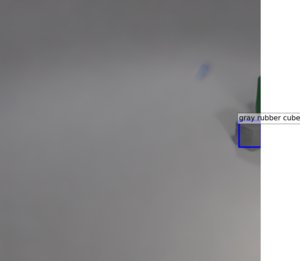}} &
              \adjustbox{valign=c}{\includegraphics[height=0.14\textwidth]{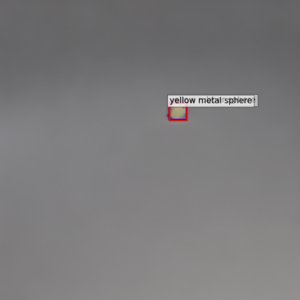}} &
              \adjustbox{valign=c}{\includegraphics[height=0.14\textwidth]{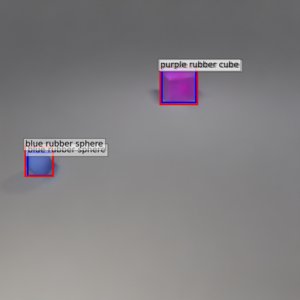}} &
              \adjustbox{valign=c}{\includegraphics[height=0.14\textwidth]{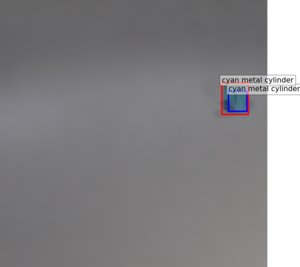}} \\
  
              \bottomrule
            \end{tblr}
        }
    \end{center}
    \caption{
    Additional \benchmark{} Number-few task samples.
    Top 4 rows: Images with GT boxes (in \textcolor{blue}{blue}).
    Bottom 4 rows: Images with GT boxes (in \textcolor{blue}{blue}) and object detection results (in \textcolor{red}{red}).
    }
    \label{tab:additional_layoutbench_number_few}
    \end{table*}
\begin{table*}[t]
    \begin{center}
        \resizebox{.75\linewidth}{!}{
          \begin{tblr}{l l l l l}
              \toprule
                
              GT &
              \adjustbox{valign=c}{\includegraphics[height=0.14\textwidth]{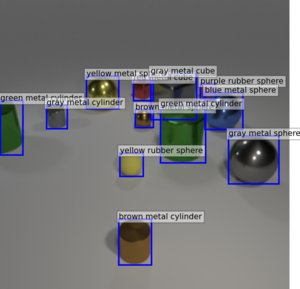}} &
              \adjustbox{valign=c}{\includegraphics[height=0.14\textwidth]{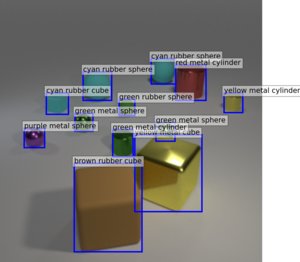}} &
              \adjustbox{valign=c}{\includegraphics[height=0.14\textwidth]{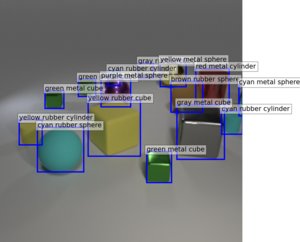}} &
              \adjustbox{valign=c}{\includegraphics[height=0.14\textwidth]{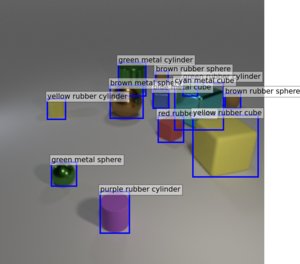}} \\
              
              \ldm{} &
              \adjustbox{valign=c}{\includegraphics[height=0.14\textwidth]{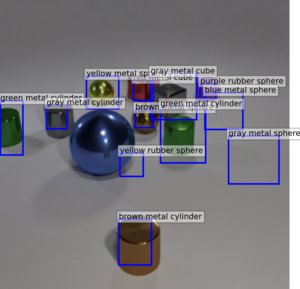}} &
              \adjustbox{valign=c}{\includegraphics[height=0.14\textwidth]{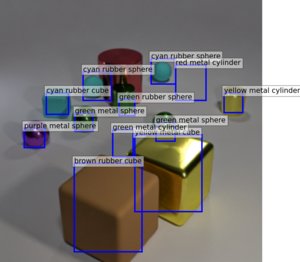}} &
              \adjustbox{valign=c}{\includegraphics[height=0.14\textwidth]{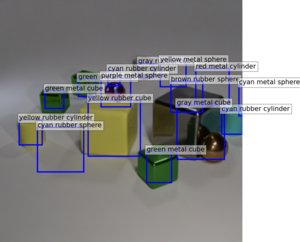}} &
              \adjustbox{valign=c}{\includegraphics[height=0.14\textwidth]{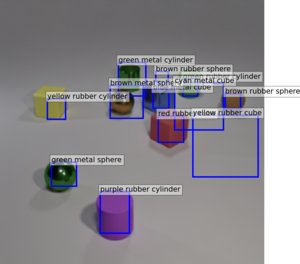}} \\
  
              \reco{} &
              \adjustbox{valign=c}{\includegraphics[height=0.14\textwidth]{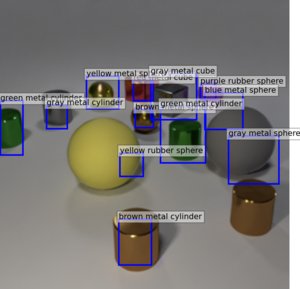}} &
              \adjustbox{valign=c}{\includegraphics[height=0.14\textwidth]{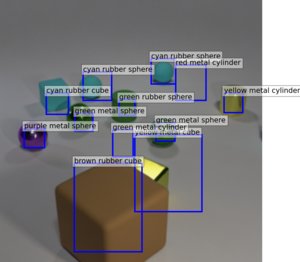}} &
              \adjustbox{valign=c}{\includegraphics[height=0.14\textwidth]{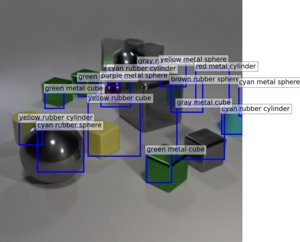}} &
              \adjustbox{valign=c}{\includegraphics[height=0.14\textwidth]{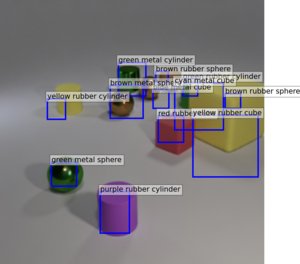}} \\
  
              {\method{} \\ (Ours)} & 
              \adjustbox{valign=c}{\includegraphics[height=0.14\textwidth]{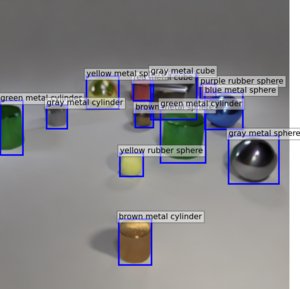}} &
              \adjustbox{valign=c}{\includegraphics[height=0.14\textwidth]{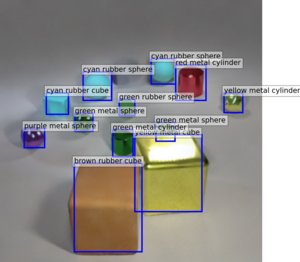}} &
              \adjustbox{valign=c}{\includegraphics[height=0.14\textwidth]{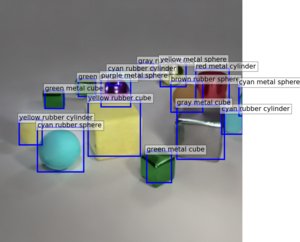}} &
              \adjustbox{valign=c}{\includegraphics[height=0.14\textwidth]{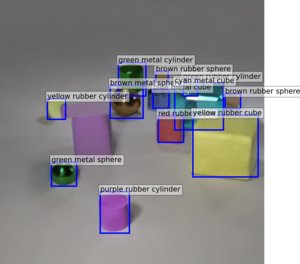}} \\
  
              \midrule
  
              GT &
              \adjustbox{valign=c}{\includegraphics[height=0.14\textwidth]{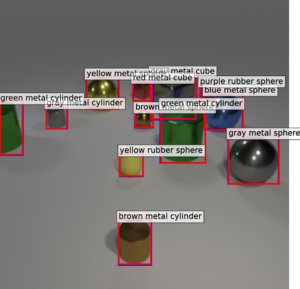}} &
              \adjustbox{valign=c}{\includegraphics[height=0.14\textwidth]{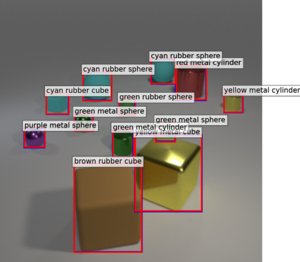}} &
              \adjustbox{valign=c}{\includegraphics[height=0.14\textwidth]{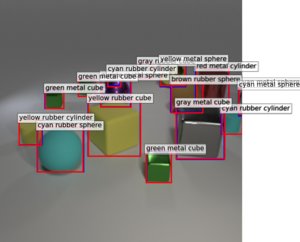}} &
              \adjustbox{valign=c}{\includegraphics[height=0.14\textwidth]{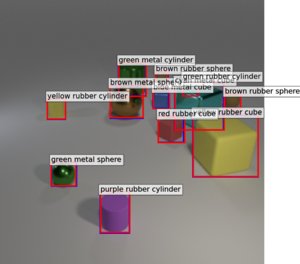}} \\
  
              \ldm{} &
              \adjustbox{valign=c}{\includegraphics[height=0.14\textwidth]{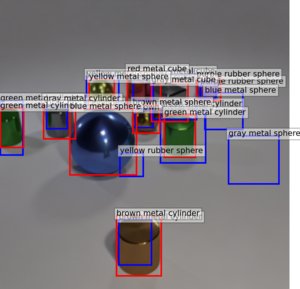}} &
              \adjustbox{valign=c}{\includegraphics[height=0.14\textwidth]{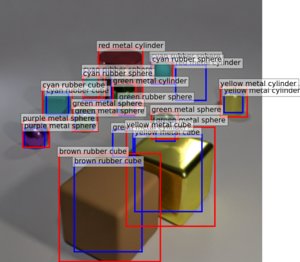}} &
              \adjustbox{valign=c}{\includegraphics[height=0.14\textwidth]{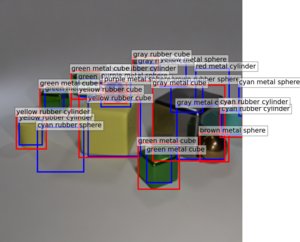}} &
              \adjustbox{valign=c}{\includegraphics[height=0.14\textwidth]{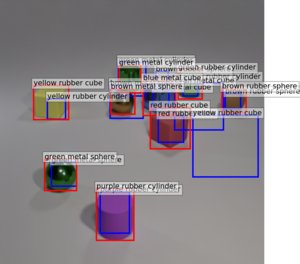}} \\
  
              \reco{} &
              \adjustbox{valign=c}{\includegraphics[height=0.14\textwidth]{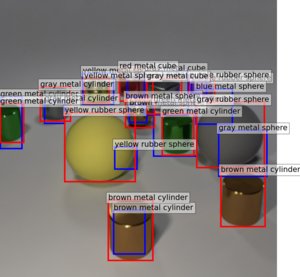}} &
              \adjustbox{valign=c}{\includegraphics[height=0.14\textwidth]{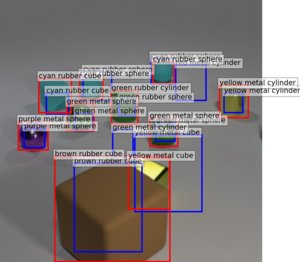}} &
              \adjustbox{valign=c}{\includegraphics[height=0.14\textwidth]{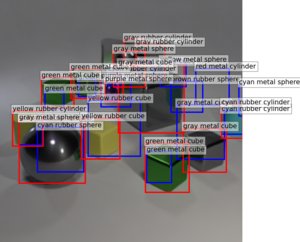}} &
              \adjustbox{valign=c}{\includegraphics[height=0.14\textwidth]{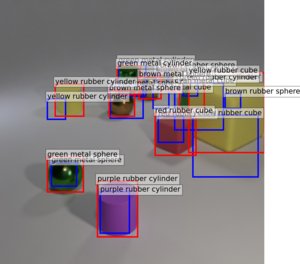}} \\
  
              {\method{} \\ (Ours)} &
              \adjustbox{valign=c}{\includegraphics[height=0.14\textwidth]{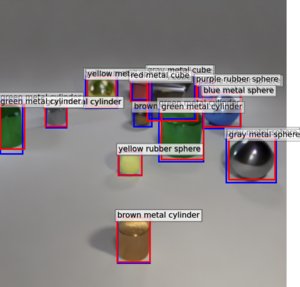}} &
              \adjustbox{valign=c}{\includegraphics[height=0.14\textwidth]{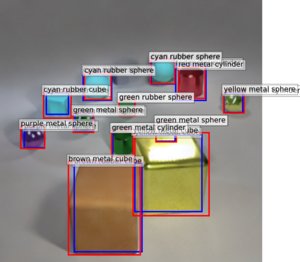}} &
              \adjustbox{valign=c}{\includegraphics[height=0.14\textwidth]{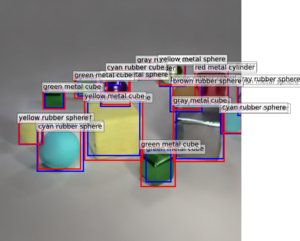}} &
              \adjustbox{valign=c}{\includegraphics[height=0.14\textwidth]{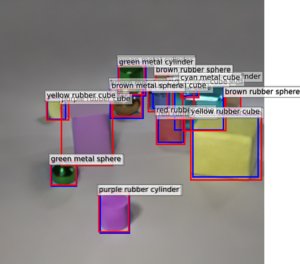}} \\
  
              \bottomrule
            \end{tblr}
        }
    \end{center}
    \caption{
    Additional \benchmark{} Number-many task samples.
    Top 4 rows: Images with GT boxes (in \textcolor{blue}{blue}).
    Bottom 4 rows: Images with GT boxes (in \textcolor{blue}{blue}) and object detection results (in \textcolor{red}{red}).
    }
    \label{tab:additional_layoutbench_number_many}
    \end{table*}
\begin{table*}[t]
    \begin{center}
        \resizebox{.75\linewidth}{!}{
          \begin{tblr}{l l l l l}
              \toprule
                
              GT &
              \adjustbox{valign=c}{\includegraphics[height=0.14\textwidth]{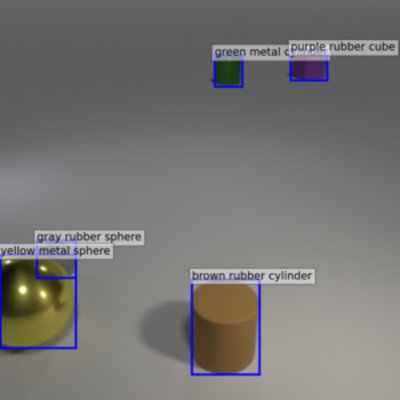}} &
              \adjustbox{valign=c}{\includegraphics[height=0.14\textwidth]{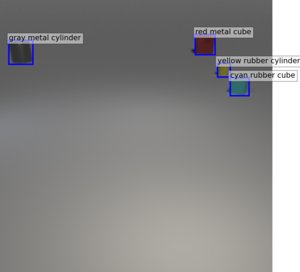}} &
              \adjustbox{valign=c}{\includegraphics[height=0.14\textwidth]{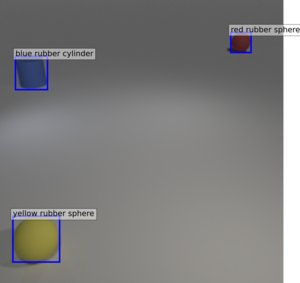}} &
              \adjustbox{valign=c}{\includegraphics[height=0.14\textwidth]{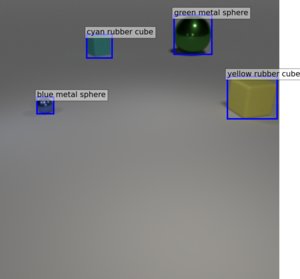}} \\
              
              \ldm{} &
              \adjustbox{valign=c}{\includegraphics[height=0.14\textwidth]{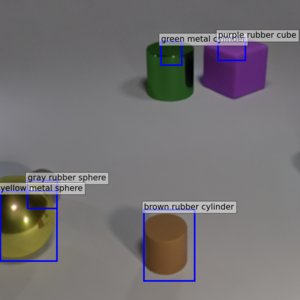}} &
              \adjustbox{valign=c}{\includegraphics[height=0.14\textwidth]{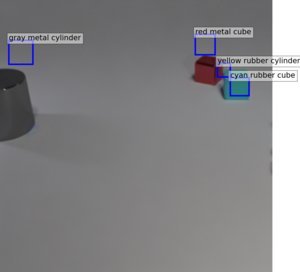}} &
              \adjustbox{valign=c}{\includegraphics[height=0.14\textwidth]{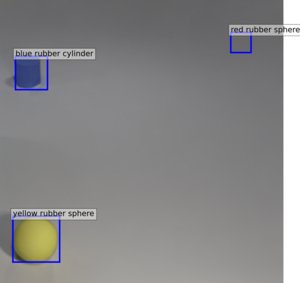}} &
              \adjustbox{valign=c}{\includegraphics[height=0.14\textwidth]{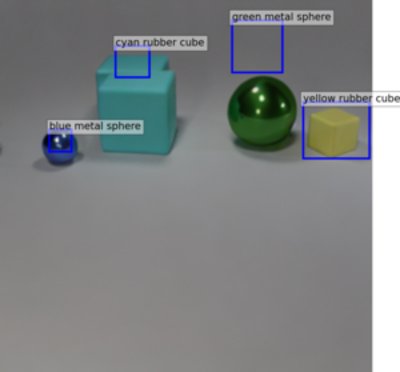}} \\
  
              \reco{} &
              \adjustbox{valign=c}{\includegraphics[height=0.14\textwidth]{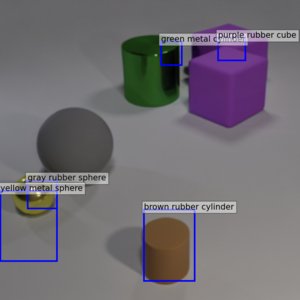}} &
              \adjustbox{valign=c}{\includegraphics[height=0.14\textwidth]{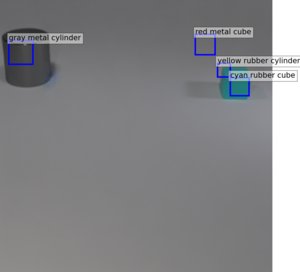}} &
              \adjustbox{valign=c}{\includegraphics[height=0.14\textwidth]{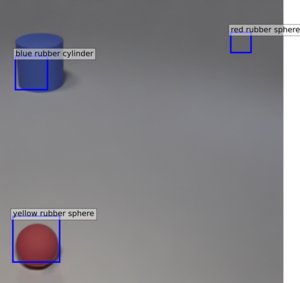}} &
              \adjustbox{valign=c}{\includegraphics[height=0.14\textwidth]{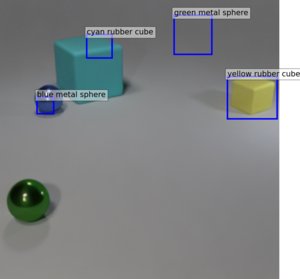}} \\
  
              {\method{} \\ (Ours)} & 
              \adjustbox{valign=c}{\includegraphics[height=0.14\textwidth]{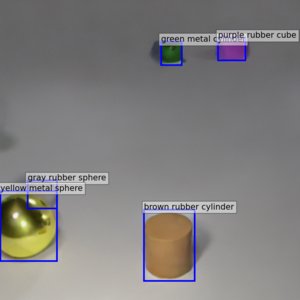}} &
              \adjustbox{valign=c}{\includegraphics[height=0.14\textwidth]{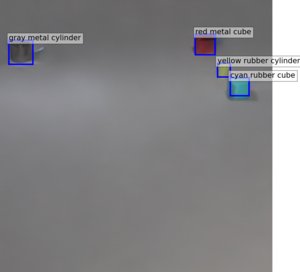}} &
              \adjustbox{valign=c}{\includegraphics[height=0.14\textwidth]{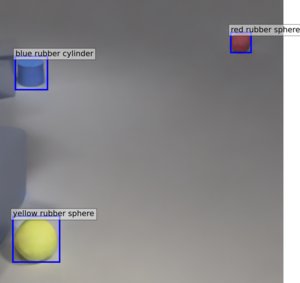}} &
              \adjustbox{valign=c}{\includegraphics[height=0.14\textwidth]{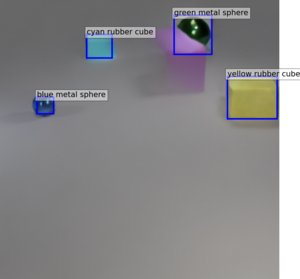}} \\
  
              \midrule
  
              GT &
              \adjustbox{valign=c}{\includegraphics[height=0.14\textwidth]{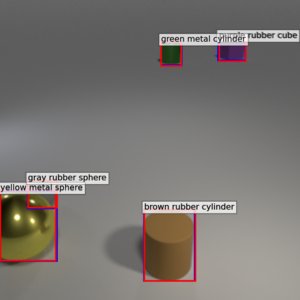}} &
              \adjustbox{valign=c}{\includegraphics[height=0.14\textwidth]{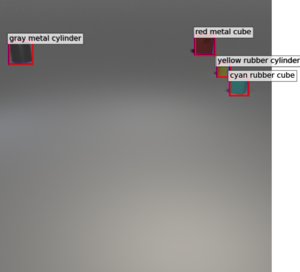}} &
              \adjustbox{valign=c}{\includegraphics[height=0.14\textwidth]{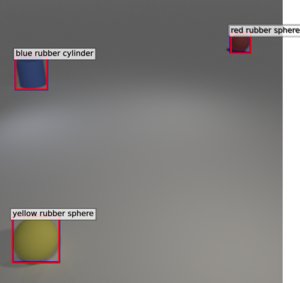}} &
              \adjustbox{valign=c}{\includegraphics[height=0.14\textwidth]{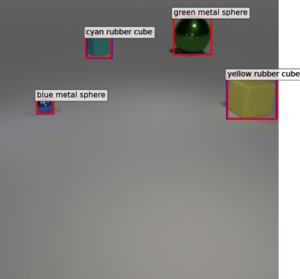}} \\
  
              \ldm{} &
              \adjustbox{valign=c}{\includegraphics[height=0.14\textwidth]{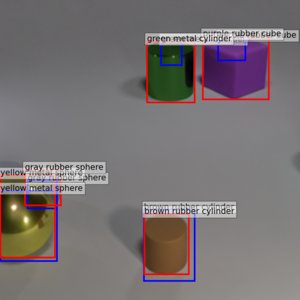}} &
              \adjustbox{valign=c}{\includegraphics[height=0.14\textwidth]{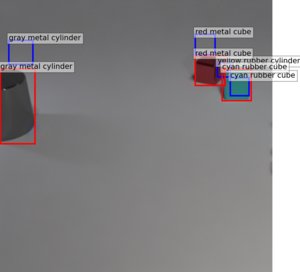}} &
              \adjustbox{valign=c}{\includegraphics[height=0.14\textwidth]{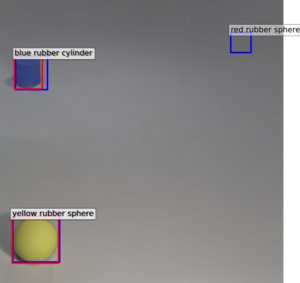}} &
              \adjustbox{valign=c}{\includegraphics[height=0.14\textwidth]{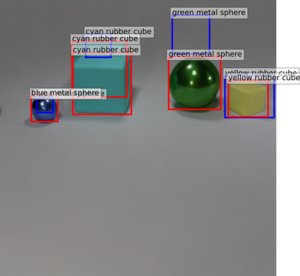}} \\
  
              \reco{} &
              \adjustbox{valign=c}{\includegraphics[height=0.14\textwidth]{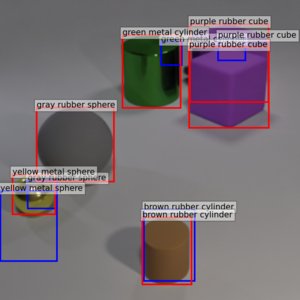}} &
              \adjustbox{valign=c}{\includegraphics[height=0.14\textwidth]{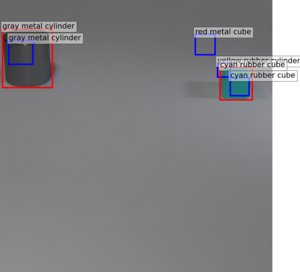}} &
              \adjustbox{valign=c}{\includegraphics[height=0.14\textwidth]{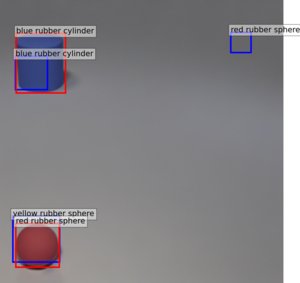}} &
              \adjustbox{valign=c}{\includegraphics[height=0.14\textwidth]{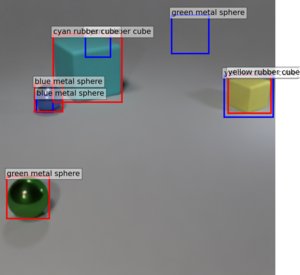}} \\
  
              {\method{} \\ (Ours)} &
              \adjustbox{valign=c}{\includegraphics[height=0.14\textwidth]{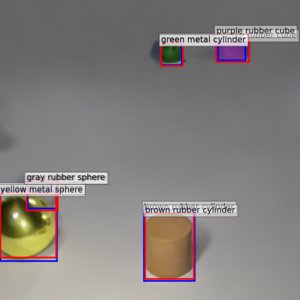}} &
              \adjustbox{valign=c}{\includegraphics[height=0.14\textwidth]{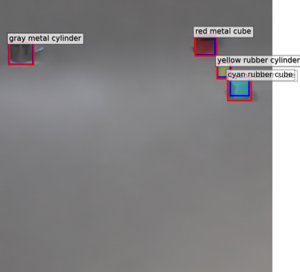}} &
              \adjustbox{valign=c}{\includegraphics[height=0.14\textwidth]{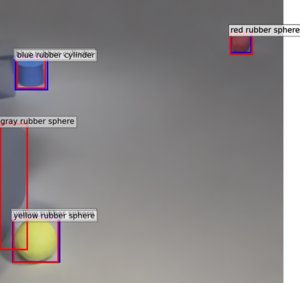}} &
              \adjustbox{valign=c}{\includegraphics[height=0.14\textwidth]{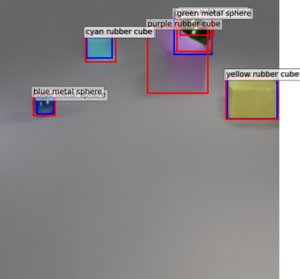}} \\
  
              \bottomrule
            \end{tblr}
        }
    \end{center}
    \caption{
    Additional \benchmark{} Position-boundary task samples.
    Top 4 rows: Images with GT boxes (in \textcolor{blue}{blue}).
    Bottom 4 rows: Images with GT boxes (in \textcolor{blue}{blue}) and object detection results (in \textcolor{red}{red}).
    }
    \label{tab:additional_layoutbench_position_boundary}
    \end{table*}
\begin{table*}[t]
    \begin{center}
        \resizebox{.75\linewidth}{!}{
          \begin{tblr}{l l l l l}
              \toprule
                
              GT &
              \adjustbox{valign=c}{\includegraphics[height=0.14\textwidth]{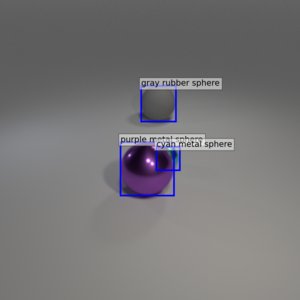}} &
              \adjustbox{valign=c}{\includegraphics[height=0.14\textwidth]{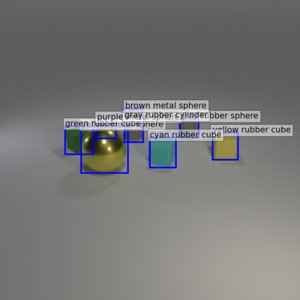}} &
              \adjustbox{valign=c}{\includegraphics[height=0.14\textwidth]{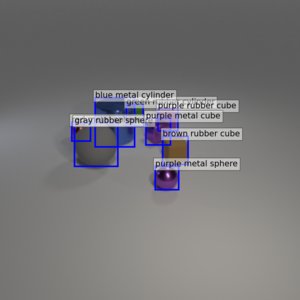}} &
              \adjustbox{valign=c}{\includegraphics[height=0.14\textwidth]{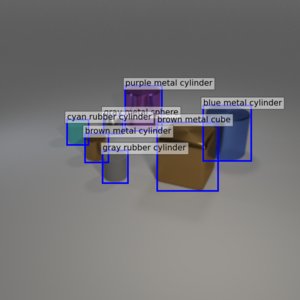}} \\
              
              \ldm{} &
              \adjustbox{valign=c}{\includegraphics[height=0.14\textwidth]{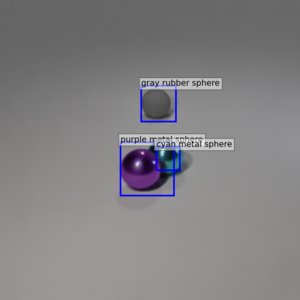}} &
              \adjustbox{valign=c}{\includegraphics[height=0.14\textwidth]{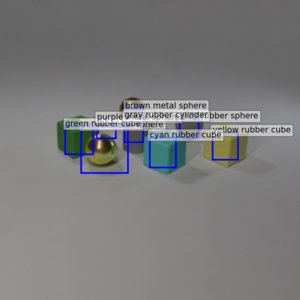}} &
              \adjustbox{valign=c}{\includegraphics[height=0.14\textwidth]{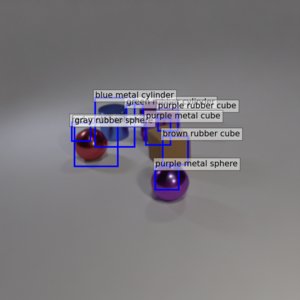}} &
              \adjustbox{valign=c}{\includegraphics[height=0.14\textwidth]{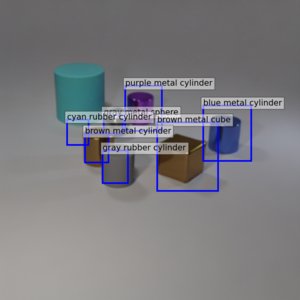}} \\
  
              \reco{} &
              \adjustbox{valign=c}{\includegraphics[height=0.14\textwidth]{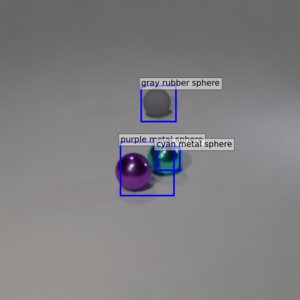}} &
              \adjustbox{valign=c}{\includegraphics[height=0.14\textwidth]{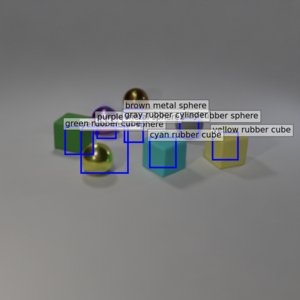}} &
              \adjustbox{valign=c}{\includegraphics[height=0.14\textwidth]{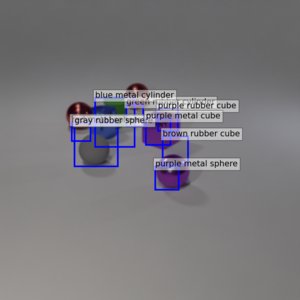}} &
              \adjustbox{valign=c}{\includegraphics[height=0.14\textwidth]{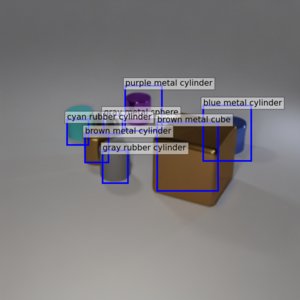}} \\
  
              {\method{} \\ (Ours)} & 
              \adjustbox{valign=c}{\includegraphics[height=0.14\textwidth]{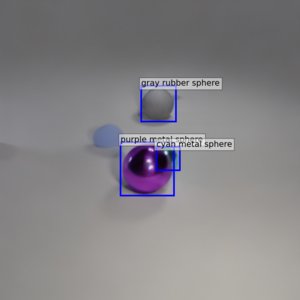}} &
              \adjustbox{valign=c}{\includegraphics[height=0.14\textwidth]{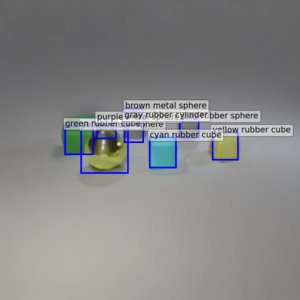}} &
              \adjustbox{valign=c}{\includegraphics[height=0.14\textwidth]{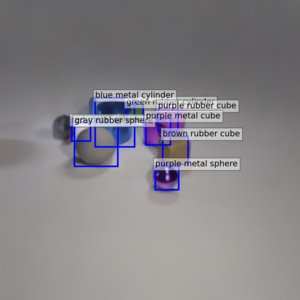}} &
              \adjustbox{valign=c}{\includegraphics[height=0.14\textwidth]{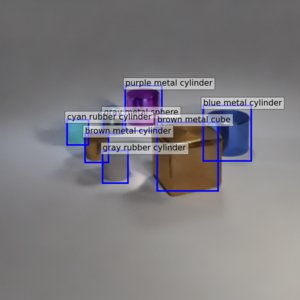}} \\
  
              \midrule
  
              GT &
              \adjustbox{valign=c}{\includegraphics[height=0.14\textwidth]{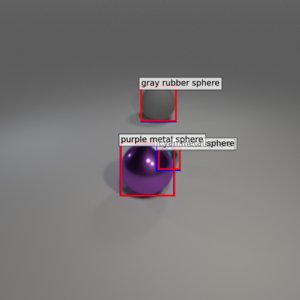}} &
              \adjustbox{valign=c}{\includegraphics[height=0.14\textwidth]{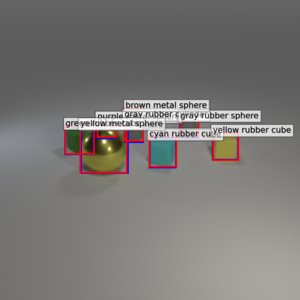}} &
              \adjustbox{valign=c}{\includegraphics[height=0.14\textwidth]{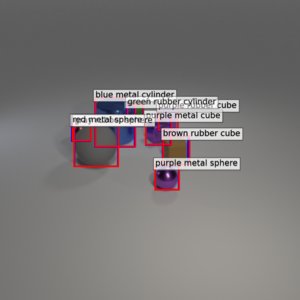}} &
              \adjustbox{valign=c}{\includegraphics[height=0.14\textwidth]{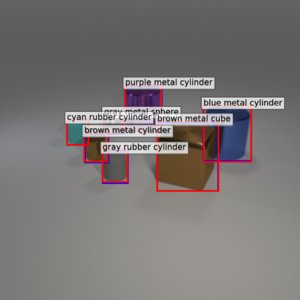}} \\
  
              \ldm{} &
              \adjustbox{valign=c}{\includegraphics[height=0.14\textwidth]{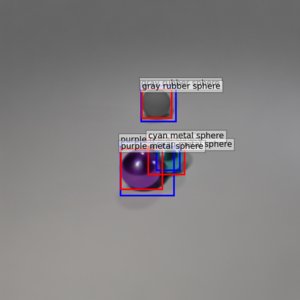}} &
              \adjustbox{valign=c}{\includegraphics[height=0.14\textwidth]{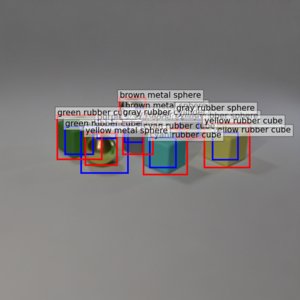}} &
              \adjustbox{valign=c}{\includegraphics[height=0.14\textwidth]{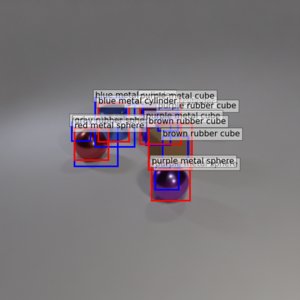}} &
              \adjustbox{valign=c}{\includegraphics[height=0.14\textwidth]{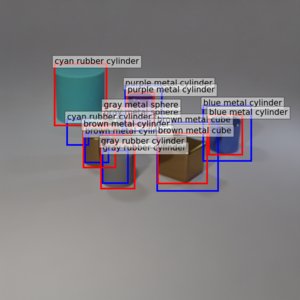}} \\
  
              \reco{} &
              \adjustbox{valign=c}{\includegraphics[height=0.14\textwidth]{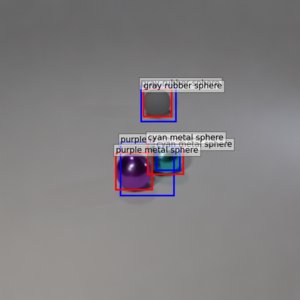}} &
              \adjustbox{valign=c}{\includegraphics[height=0.14\textwidth]{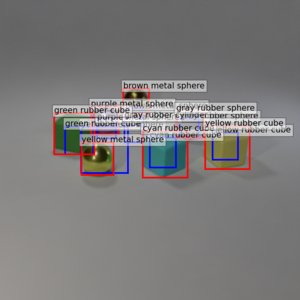}} &
              \adjustbox{valign=c}{\includegraphics[height=0.14\textwidth]{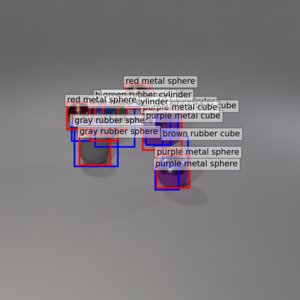}} &
              \adjustbox{valign=c}{\includegraphics[height=0.14\textwidth]{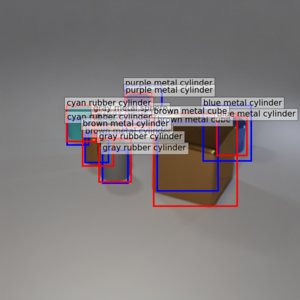}} \\
  
              {\method{} \\ (Ours)} &
              \adjustbox{valign=c}{\includegraphics[height=0.14\textwidth]{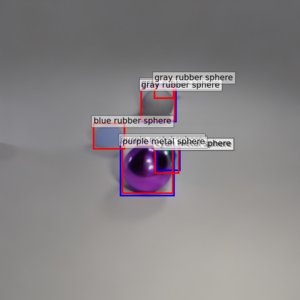}} &
              \adjustbox{valign=c}{\includegraphics[height=0.14\textwidth]{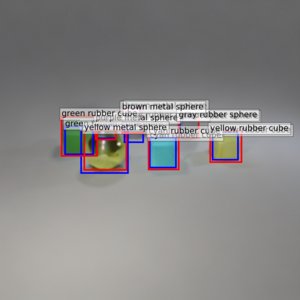}} &
              \adjustbox{valign=c}{\includegraphics[height=0.14\textwidth]{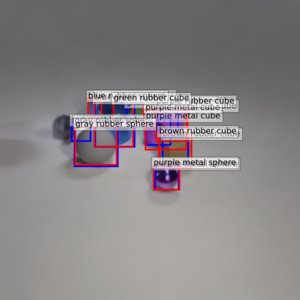}} &
              \adjustbox{valign=c}{\includegraphics[height=0.14\textwidth]{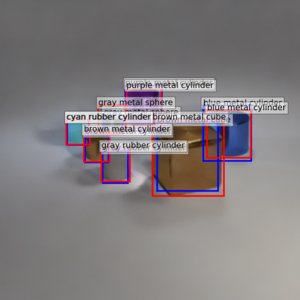}} \\
  
              \bottomrule
            \end{tblr}
        }
    \end{center}
    \caption{
    Additional \benchmark{} Position-center task samples.
    Top 4 rows: Images with GT boxes (in \textcolor{blue}{blue}).
    Bottom 4 rows: Images with GT boxes (in \textcolor{blue}{blue}) and object detection results (in \textcolor{red}{red}).
    }
    \label{tab:additional_layoutbench_position_center}
    \end{table*}
\begin{table*}[t]
    \begin{center}
        \resizebox{.75\linewidth}{!}{
          \begin{tblr}{l l l l l}
              \toprule
                
              GT &
              \adjustbox{valign=c}{\includegraphics[height=0.14\textwidth]{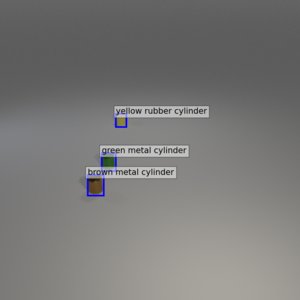}} &
              \adjustbox{valign=c}{\includegraphics[height=0.14\textwidth]{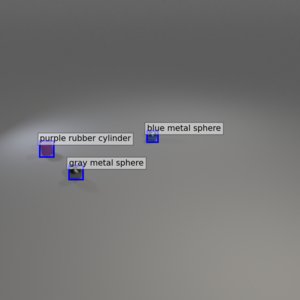}} &
              \adjustbox{valign=c}{\includegraphics[height=0.14\textwidth]{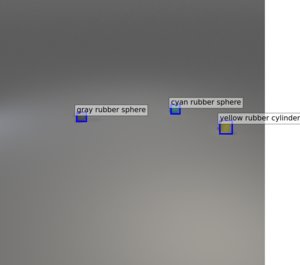}} &
              \adjustbox{valign=c}{\includegraphics[height=0.14\textwidth]{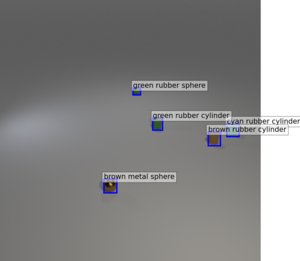}} \\
              
              \ldm{} &
              \adjustbox{valign=c}{\includegraphics[height=0.14\textwidth]{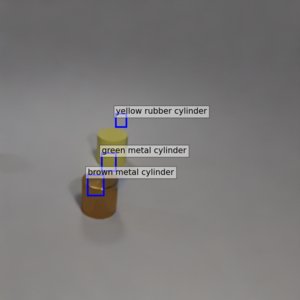}} &
              \adjustbox{valign=c}{\includegraphics[height=0.14\textwidth]{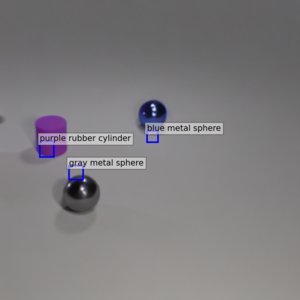}} &
              \adjustbox{valign=c}{\includegraphics[height=0.14\textwidth]{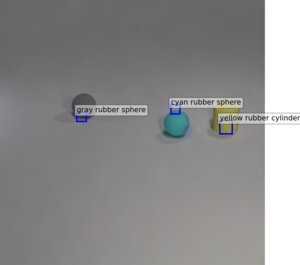}} &
              \adjustbox{valign=c}{\includegraphics[height=0.14\textwidth]{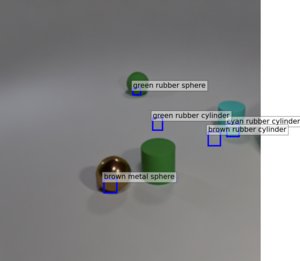}} \\
  
              \reco{} &
              \adjustbox{valign=c}{\includegraphics[height=0.14\textwidth]{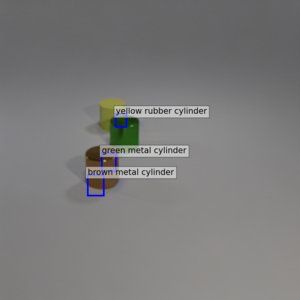}} &
              \adjustbox{valign=c}{\includegraphics[height=0.14\textwidth]{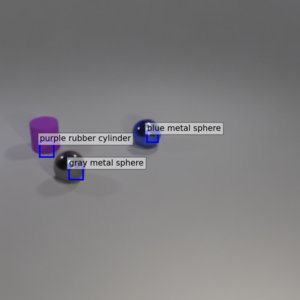}} &
              \adjustbox{valign=c}{\includegraphics[height=0.14\textwidth]{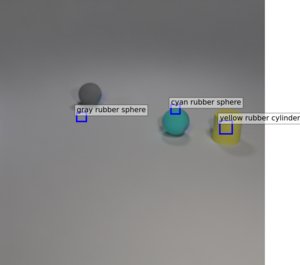}} &
              \adjustbox{valign=c}{\includegraphics[height=0.14\textwidth]{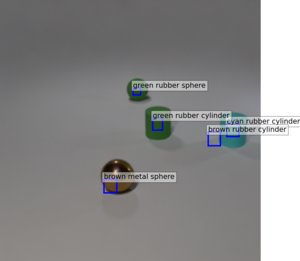}} \\
  
              {\method{} \\ (Ours)} & 
              \adjustbox{valign=c}{\includegraphics[height=0.14\textwidth]{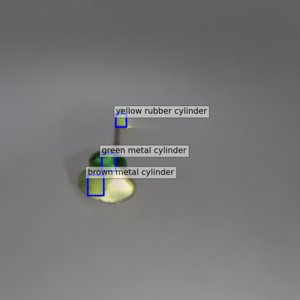}} &
              \adjustbox{valign=c}{\includegraphics[height=0.14\textwidth]{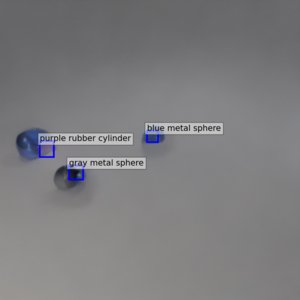}} &
              \adjustbox{valign=c}{\includegraphics[height=0.14\textwidth]{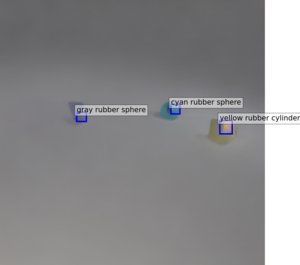}} &
              \adjustbox{valign=c}{\includegraphics[height=0.14\textwidth]{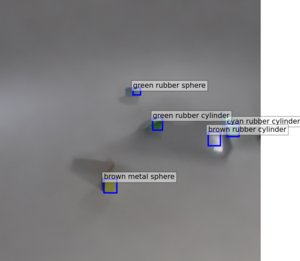}} \\
  
              \midrule
  
              GT &
              \adjustbox{valign=c}{\includegraphics[height=0.14\textwidth]{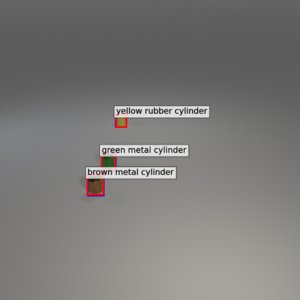}} &
              \adjustbox{valign=c}{\includegraphics[height=0.14\textwidth]{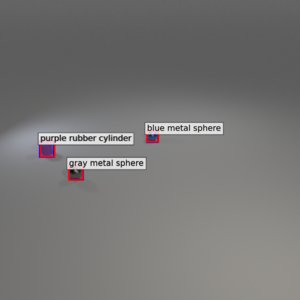}} &
              \adjustbox{valign=c}{\includegraphics[height=0.14\textwidth]{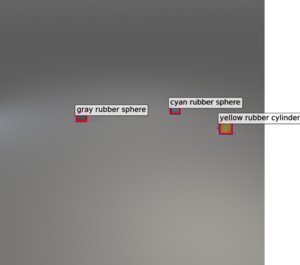}} &
              \adjustbox{valign=c}{\includegraphics[height=0.14\textwidth]{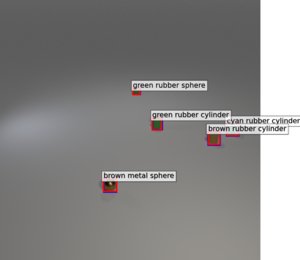}} \\
  
              \ldm{} &
              \adjustbox{valign=c}{\includegraphics[height=0.14\textwidth]{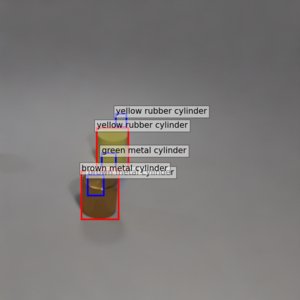}} &
              \adjustbox{valign=c}{\includegraphics[height=0.14\textwidth]{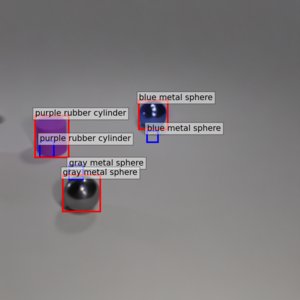}} &
              \adjustbox{valign=c}{\includegraphics[height=0.14\textwidth]{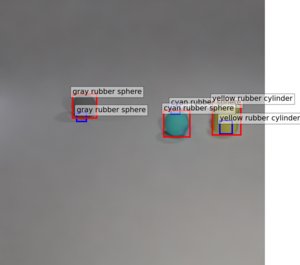}} &
              \adjustbox{valign=c}{\includegraphics[height=0.14\textwidth]{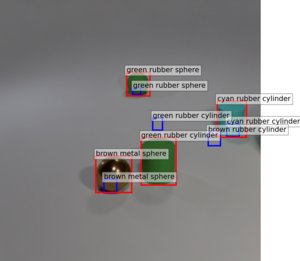}} \\
  
              \reco{} &
              \adjustbox{valign=c}{\includegraphics[height=0.14\textwidth]{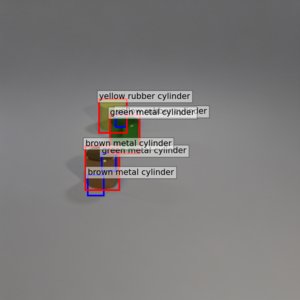}} &
              \adjustbox{valign=c}{\includegraphics[height=0.14\textwidth]{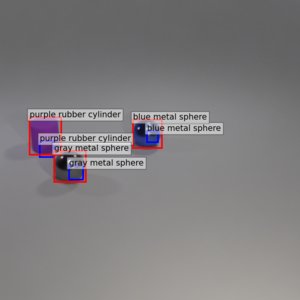}} &
              \adjustbox{valign=c}{\includegraphics[height=0.14\textwidth]{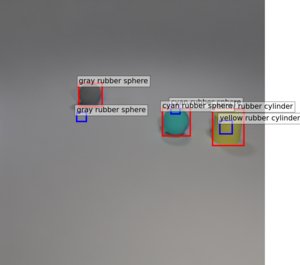}} &
              \adjustbox{valign=c}{\includegraphics[height=0.14\textwidth]{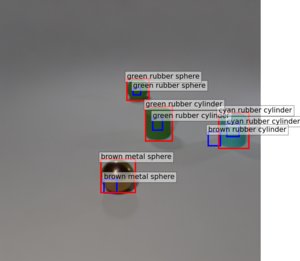}} \\
  
              {\method{} \\ (Ours)} &
              \adjustbox{valign=c}{\includegraphics[height=0.14\textwidth]{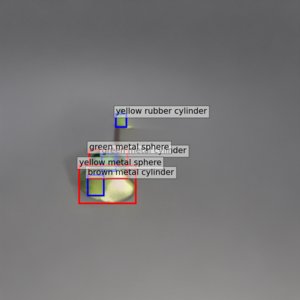}} &
              \adjustbox{valign=c}{\includegraphics[height=0.14\textwidth]{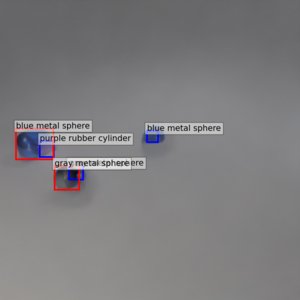}} &
              \adjustbox{valign=c}{\includegraphics[height=0.14\textwidth]{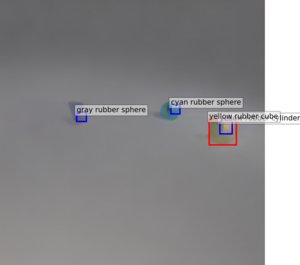}} &
              \adjustbox{valign=c}{\includegraphics[height=0.14\textwidth]{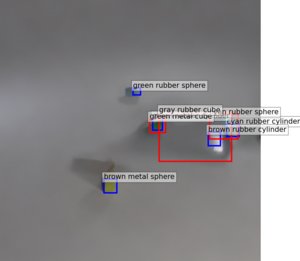}} \\
  
              \bottomrule
            \end{tblr}
        }
    \end{center}
    \caption{
    Additional \benchmark{} Size-tiny task samples.
    Top 4 rows: Images with GT boxes (in \textcolor{blue}{blue}).
    Bottom 4 rows: Images with GT boxes (in \textcolor{blue}{blue}) and object detection results (in \textcolor{red}{red}).
    }
    \label{tab:additional_layoutbench_size_tiny}
    \end{table*}
\begin{table*}[t]
    \begin{center}
        \resizebox{.75\linewidth}{!}{
          \begin{tblr}{l l l l l}
              \toprule
                
              GT &
              \adjustbox{valign=c}{\includegraphics[height=0.14\textwidth]{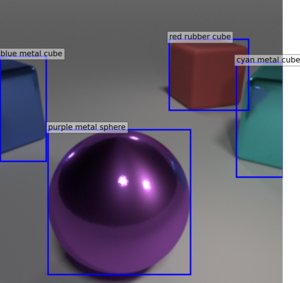}} &
              \adjustbox{valign=c}{\includegraphics[height=0.14\textwidth]{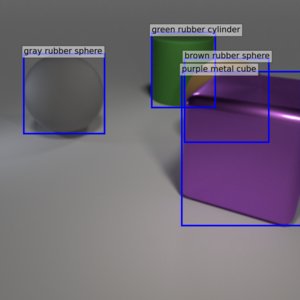}} &
              \adjustbox{valign=c}{\includegraphics[height=0.14\textwidth]{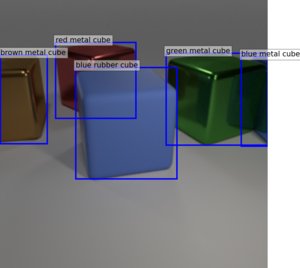}} &
              \adjustbox{valign=c}{\includegraphics[height=0.14\textwidth]{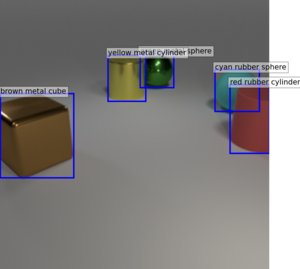}} \\
              
              \ldm{} &
              \adjustbox{valign=c}{\includegraphics[height=0.14\textwidth]{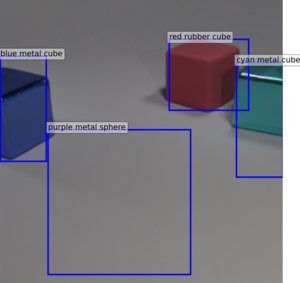}} &
              \adjustbox{valign=c}{\includegraphics[height=0.14\textwidth]{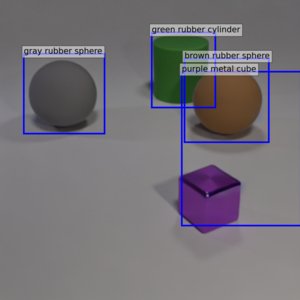}} &
              \adjustbox{valign=c}{\includegraphics[height=0.14\textwidth]{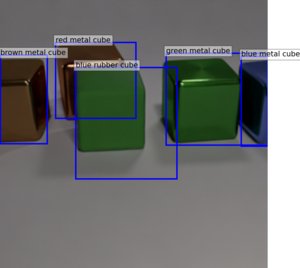}} &
              \adjustbox{valign=c}{\includegraphics[height=0.14\textwidth]{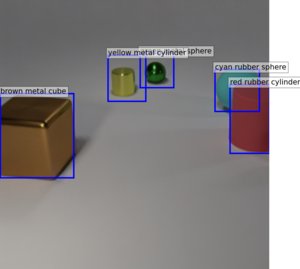}} \\
  
              \reco{} &
              \adjustbox{valign=c}{\includegraphics[height=0.14\textwidth]{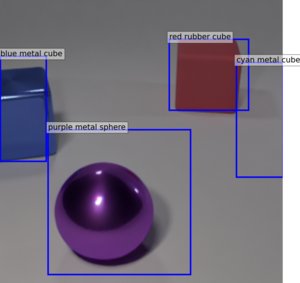}} &
              \adjustbox{valign=c}{\includegraphics[height=0.14\textwidth]{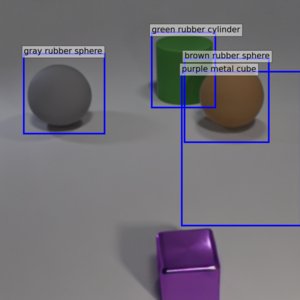}} &
              \adjustbox{valign=c}{\includegraphics[height=0.14\textwidth]{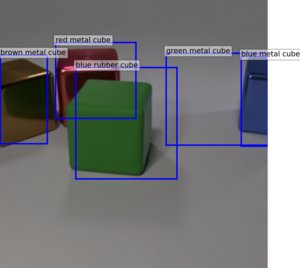}} &
              \adjustbox{valign=c}{\includegraphics[height=0.14\textwidth]{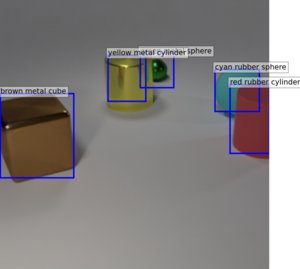}} \\
  
              {\method{} \\ (Ours)} & 
              \adjustbox{valign=c}{\includegraphics[height=0.14\textwidth]{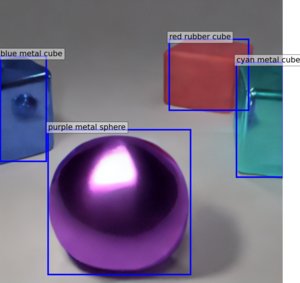}} &
              \adjustbox{valign=c}{\includegraphics[height=0.14\textwidth]{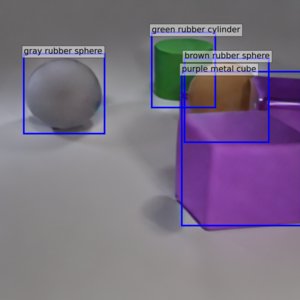}} &
              \adjustbox{valign=c}{\includegraphics[height=0.14\textwidth]{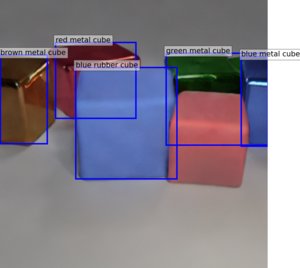}} &
              \adjustbox{valign=c}{\includegraphics[height=0.14\textwidth]{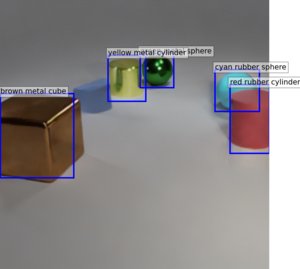}} \\
  
              \midrule
  
              GT &
              \adjustbox{valign=c}{\includegraphics[height=0.14\textwidth]{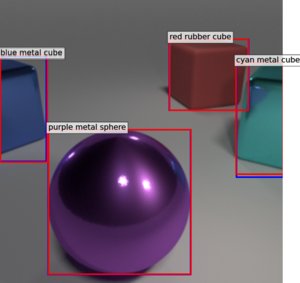}} &
              \adjustbox{valign=c}{\includegraphics[height=0.14\textwidth]{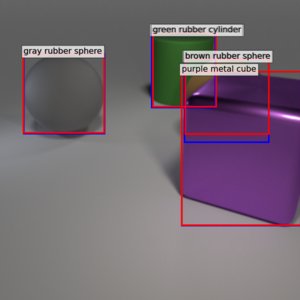}} &
              \adjustbox{valign=c}{\includegraphics[height=0.14\textwidth]{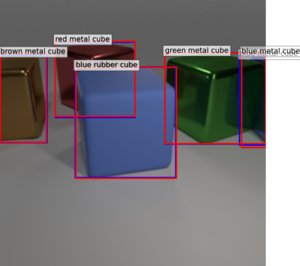}} &
              \adjustbox{valign=c}{\includegraphics[height=0.14\textwidth]{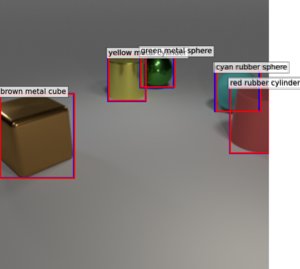}} \\
  
              \ldm{} &
              \adjustbox{valign=c}{\includegraphics[height=0.14\textwidth]{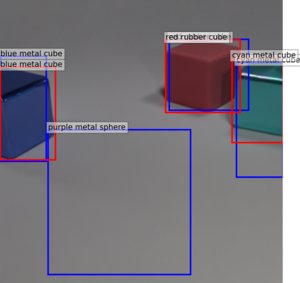}} &
              \adjustbox{valign=c}{\includegraphics[height=0.14\textwidth]{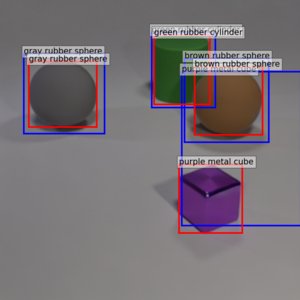}} &
              \adjustbox{valign=c}{\includegraphics[height=0.14\textwidth]{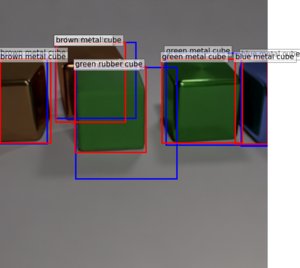}} &
              \adjustbox{valign=c}{\includegraphics[height=0.14\textwidth]{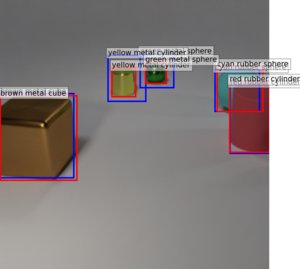}} \\
  
              \reco{} &
              \adjustbox{valign=c}{\includegraphics[height=0.14\textwidth]{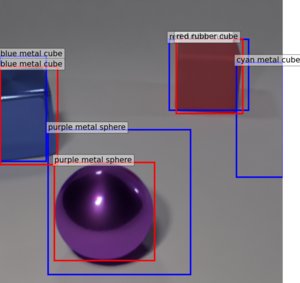}} &
              \adjustbox{valign=c}{\includegraphics[height=0.14\textwidth]{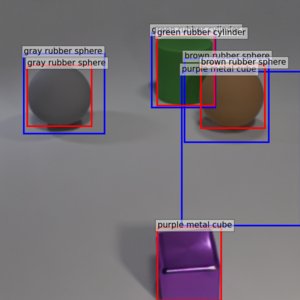}} &
              \adjustbox{valign=c}{\includegraphics[height=0.14\textwidth]{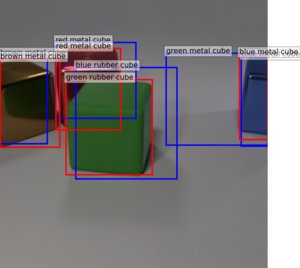}} &
              \adjustbox{valign=c}{\includegraphics[height=0.14\textwidth]{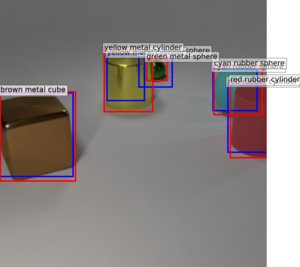}} \\
  
              {\method{} \\ (Ours)} &
              \adjustbox{valign=c}{\includegraphics[height=0.14\textwidth]{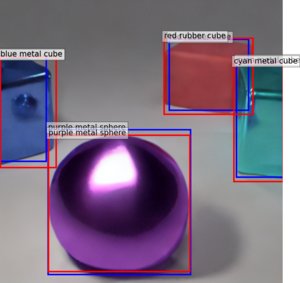}} &
              \adjustbox{valign=c}{\includegraphics[height=0.14\textwidth]{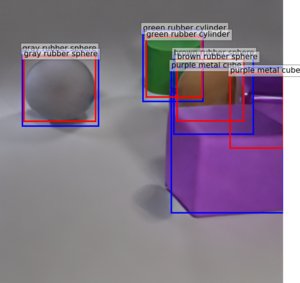}} &
              \adjustbox{valign=c}{\includegraphics[height=0.14\textwidth]{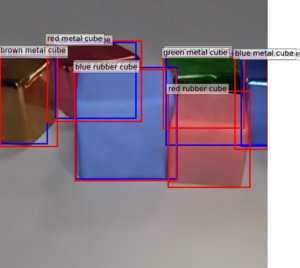}} &
              \adjustbox{valign=c}{\includegraphics[height=0.14\textwidth]{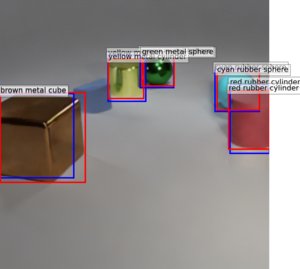}} \\
  
              \bottomrule
            \end{tblr}
        }
    \end{center}
    \caption{
    Additional \benchmark{} Size-large task samples.
    Top 4 rows: Images with GT boxes (in \textcolor{blue}{blue}).
    Bottom 4 rows: Images with GT boxes (in \textcolor{blue}{blue}) and object detection results (in \textcolor{red}{red}).
    }
    \label{tab:additional_layoutbench_size_large}
    \end{table*}
\begin{table*}[t]
    \begin{center}
        \resizebox{.75\linewidth}{!}{
          \begin{tblr}{l l l l l}
              \toprule
                
              GT &
              \adjustbox{valign=c}{\includegraphics[height=0.14\textwidth]{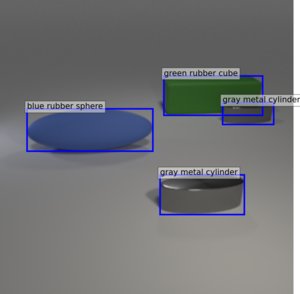}} &
              \adjustbox{valign=c}{\includegraphics[height=0.14\textwidth]{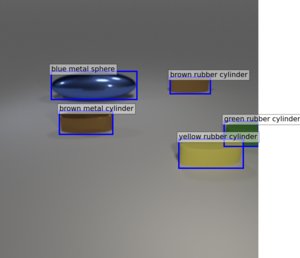}} &
              \adjustbox{valign=c}{\includegraphics[height=0.14\textwidth]{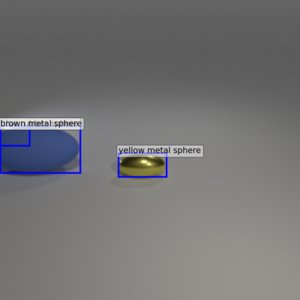}} &
              \adjustbox{valign=c}{\includegraphics[height=0.14\textwidth]{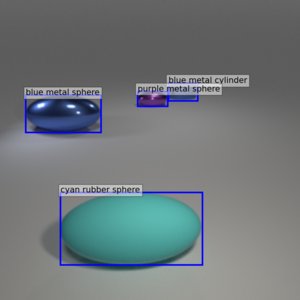}} \\
              
              \ldm{} &
              \adjustbox{valign=c}{\includegraphics[height=0.14\textwidth]{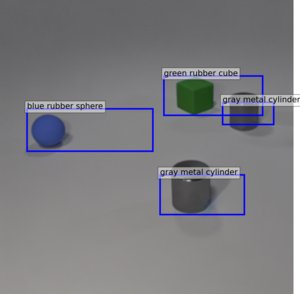}} &
              \adjustbox{valign=c}{\includegraphics[height=0.14\textwidth]{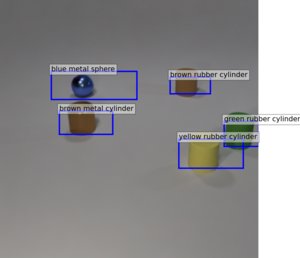}} &
              \adjustbox{valign=c}{\includegraphics[height=0.14\textwidth]{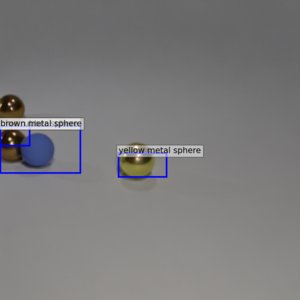}} &
              \adjustbox{valign=c}{\includegraphics[height=0.14\textwidth]{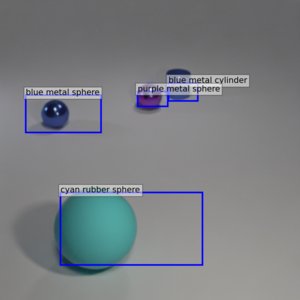}} \\
  
              \reco{} &
              \adjustbox{valign=c}{\includegraphics[height=0.14\textwidth]{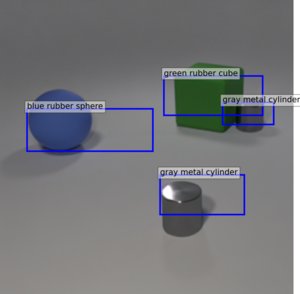}} &
              \adjustbox{valign=c}{\includegraphics[height=0.14\textwidth]{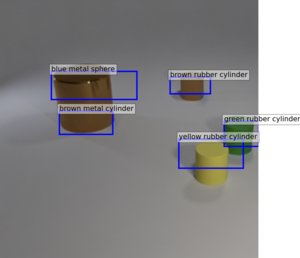}} &
              \adjustbox{valign=c}{\includegraphics[height=0.14\textwidth]{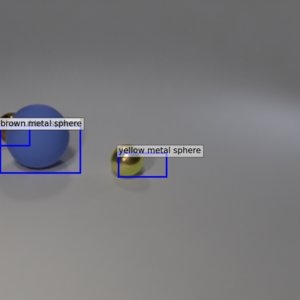}} &
              \adjustbox{valign=c}{\includegraphics[height=0.14\textwidth]{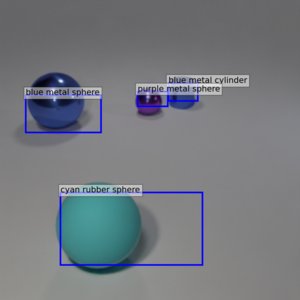}} \\
  
              {\method{} \\ (Ours)} & 
              \adjustbox{valign=c}{\includegraphics[height=0.14\textwidth]{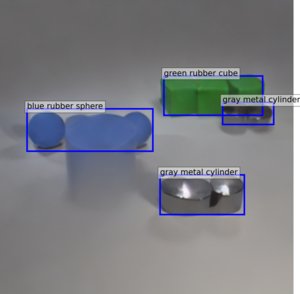}} &
              \adjustbox{valign=c}{\includegraphics[height=0.14\textwidth]{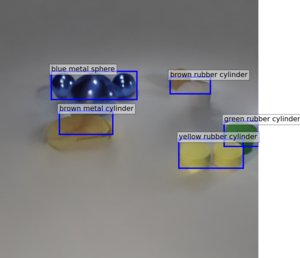}} &
              \adjustbox{valign=c}{\includegraphics[height=0.14\textwidth]{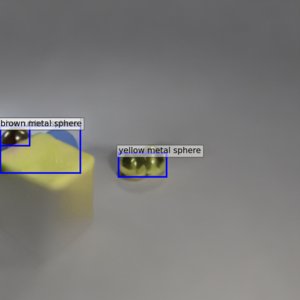}} &
              \adjustbox{valign=c}{\includegraphics[height=0.14\textwidth]{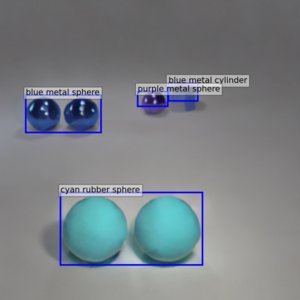}} \\
  
              \midrule
  
              GT &
              \adjustbox{valign=c}{\includegraphics[height=0.14\textwidth]{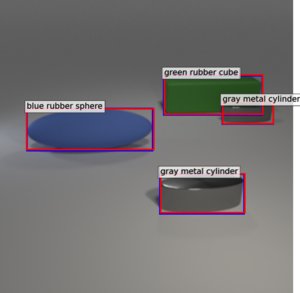}} &
              \adjustbox{valign=c}{\includegraphics[height=0.14\textwidth]{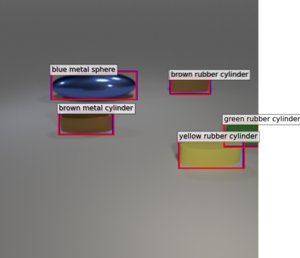}} &
              \adjustbox{valign=c}{\includegraphics[height=0.14\textwidth]{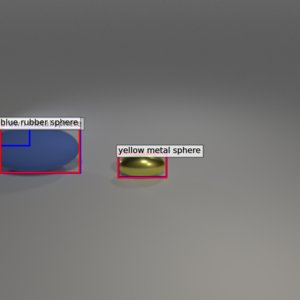}} &
              \adjustbox{valign=c}{\includegraphics[height=0.14\textwidth]{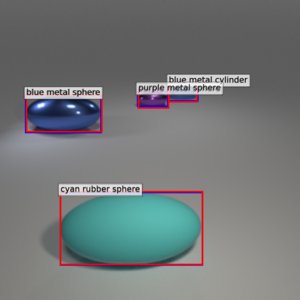}} \\
  
              \ldm{} &
              \adjustbox{valign=c}{\includegraphics[height=0.14\textwidth]{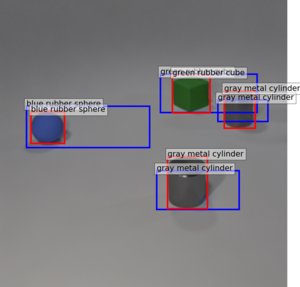}} &
              \adjustbox{valign=c}{\includegraphics[height=0.14\textwidth]{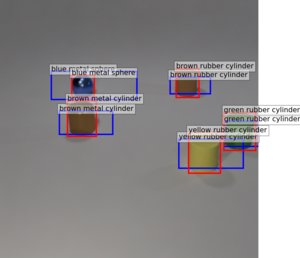}} &
              \adjustbox{valign=c}{\includegraphics[height=0.14\textwidth]{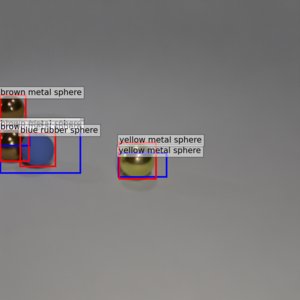}} &
              \adjustbox{valign=c}{\includegraphics[height=0.14\textwidth]{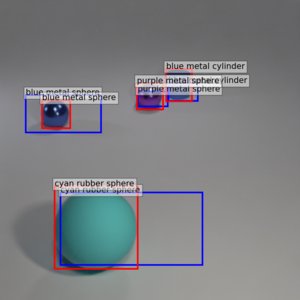}} \\
  
              \reco{} &
              \adjustbox{valign=c}{\includegraphics[height=0.14\textwidth]{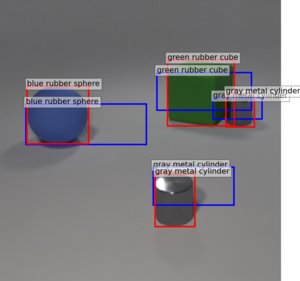}} &
              \adjustbox{valign=c}{\includegraphics[height=0.14\textwidth]{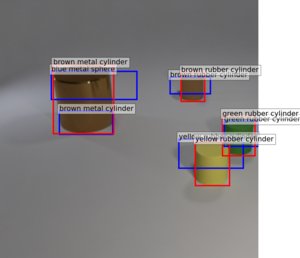}} &
              \adjustbox{valign=c}{\includegraphics[height=0.14\textwidth]{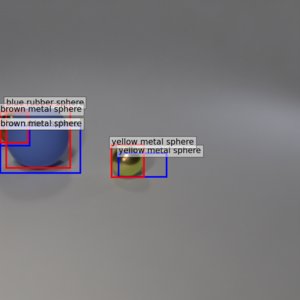}} &
              \adjustbox{valign=c}{\includegraphics[height=0.14\textwidth]{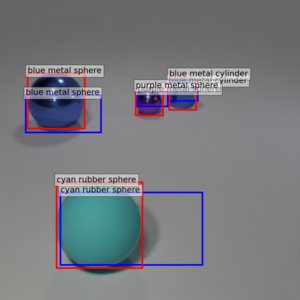}} \\
  
              {\method{} \\ (Ours)} &
              \adjustbox{valign=c}{\includegraphics[height=0.14\textwidth]{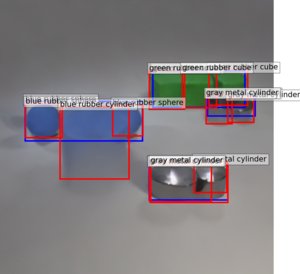}} &
              \adjustbox{valign=c}{\includegraphics[height=0.14\textwidth]{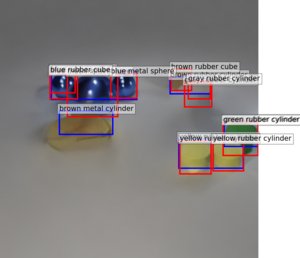}} &
              \adjustbox{valign=c}{\includegraphics[height=0.14\textwidth]{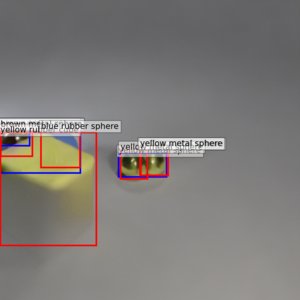}} &
              \adjustbox{valign=c}{\includegraphics[height=0.14\textwidth]{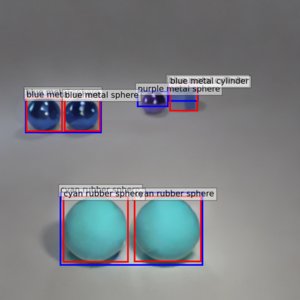}} \\
  
              \bottomrule
            \end{tblr}
        }
    \end{center}
    \caption{
    Additional \benchmark{} Shape-horizontal task samples.
    Top 4 rows: Images with GT boxes (in \textcolor{blue}{blue}).
    Bottom 4 rows: Images with GT boxes (in \textcolor{blue}{blue}) and object detection results (in \textcolor{red}{red}).
    }
    \label{tab:additional_layoutbench_shape_horizontal}
    \end{table*}
\begin{table*}[t]
    \begin{center}
        \resizebox{.75\linewidth}{!}{
          \begin{tblr}{l l l l l}
              \toprule
                
              GT &
              \adjustbox{valign=c}{\includegraphics[height=0.14\textwidth]{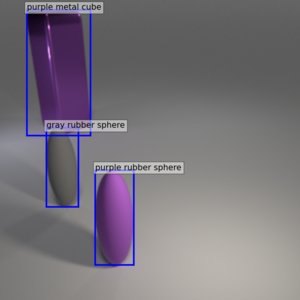}} &
              \adjustbox{valign=c}{\includegraphics[height=0.14\textwidth]{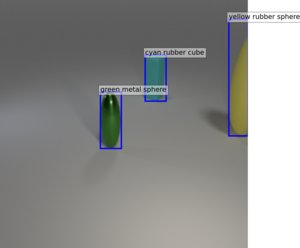}} &
              \adjustbox{valign=c}{\includegraphics[height=0.14\textwidth]{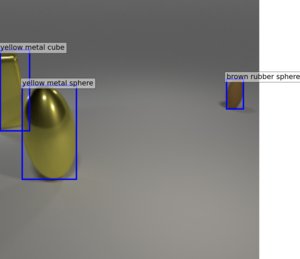}} &
              \adjustbox{valign=c}{\includegraphics[height=0.14\textwidth]{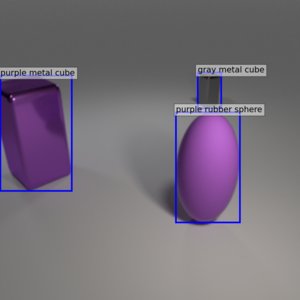}} \\
              
              \ldm{} &
              \adjustbox{valign=c}{\includegraphics[height=0.14\textwidth]{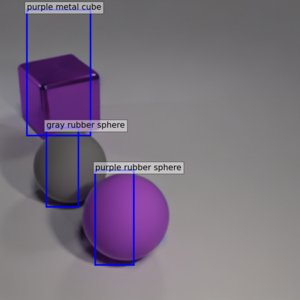}} &
              \adjustbox{valign=c}{\includegraphics[height=0.14\textwidth]{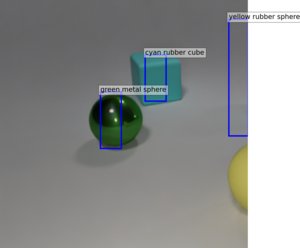}} &
              \adjustbox{valign=c}{\includegraphics[height=0.14\textwidth]{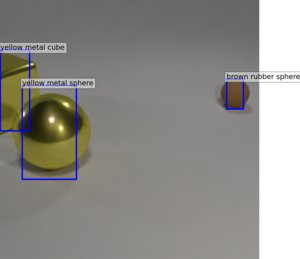}} &
              \adjustbox{valign=c}{\includegraphics[height=0.14\textwidth]{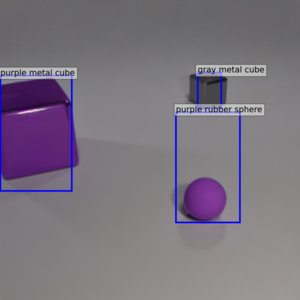}} \\
  
              \reco{} &
              \adjustbox{valign=c}{\includegraphics[height=0.14\textwidth]{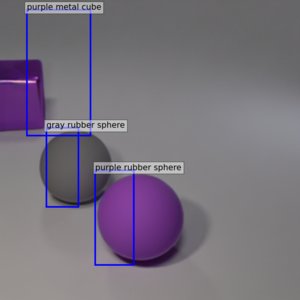}} &
              \adjustbox{valign=c}{\includegraphics[height=0.14\textwidth]{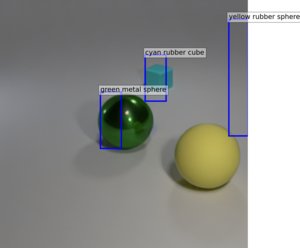}} &
              \adjustbox{valign=c}{\includegraphics[height=0.14\textwidth]{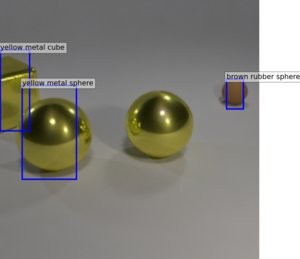}} &
              \adjustbox{valign=c}{\includegraphics[height=0.14\textwidth]{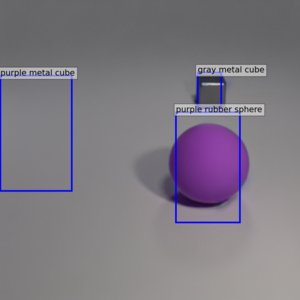}} \\
  
              {\method{} \\ (Ours)} & 
              \adjustbox{valign=c}{\includegraphics[height=0.14\textwidth]{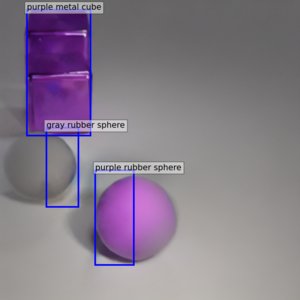}} &
              \adjustbox{valign=c}{\includegraphics[height=0.14\textwidth]{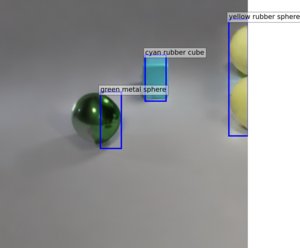}} &
              \adjustbox{valign=c}{\includegraphics[height=0.14\textwidth]{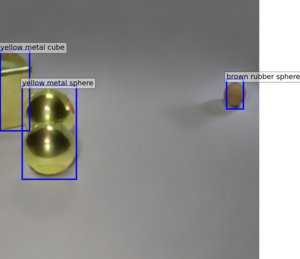}} &
              \adjustbox{valign=c}{\includegraphics[height=0.14\textwidth]{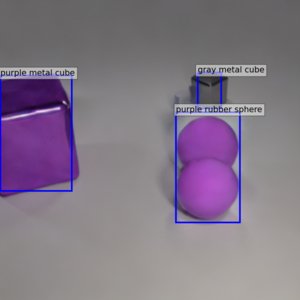}} \\
  
              \midrule
  
              GT &
              \adjustbox{valign=c}{\includegraphics[height=0.14\textwidth]{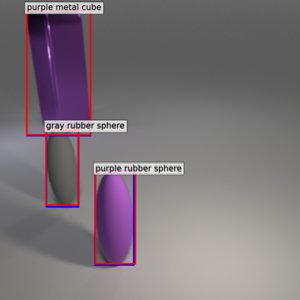}} &
              \adjustbox{valign=c}{\includegraphics[height=0.14\textwidth]{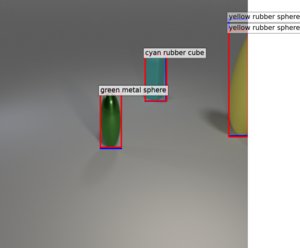}} &
              \adjustbox{valign=c}{\includegraphics[height=0.14\textwidth]{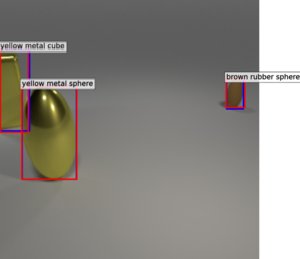}} &
              \adjustbox{valign=c}{\includegraphics[height=0.14\textwidth]{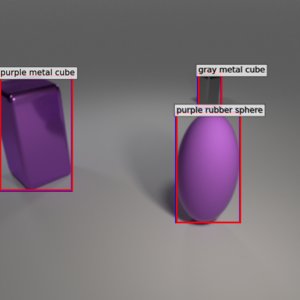}} \\
  
              \ldm{} &
              \adjustbox{valign=c}{\includegraphics[height=0.14\textwidth]{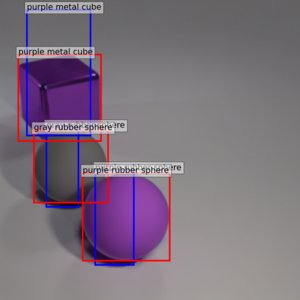}} &
              \adjustbox{valign=c}{\includegraphics[height=0.14\textwidth]{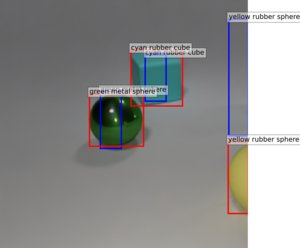}} &
              \adjustbox{valign=c}{\includegraphics[height=0.14\textwidth]{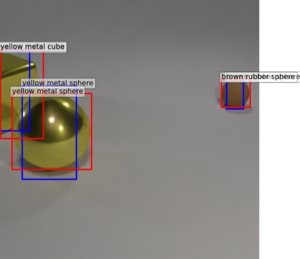}} &
              \adjustbox{valign=c}{\includegraphics[height=0.14\textwidth]{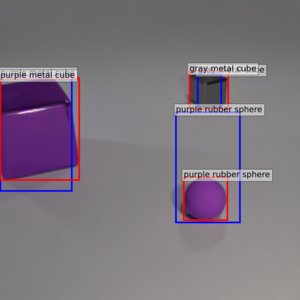}} \\
  
              \reco{} &
              \adjustbox{valign=c}{\includegraphics[height=0.14\textwidth]{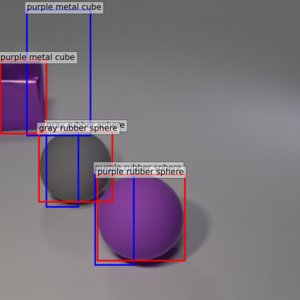}} &
              \adjustbox{valign=c}{\includegraphics[height=0.14\textwidth]{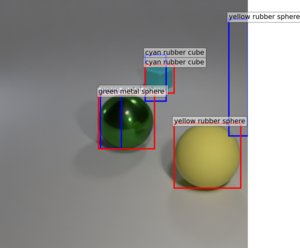}} &
              \adjustbox{valign=c}{\includegraphics[height=0.14\textwidth]{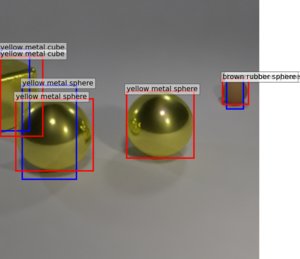}} &
              \adjustbox{valign=c}{\includegraphics[height=0.14\textwidth]{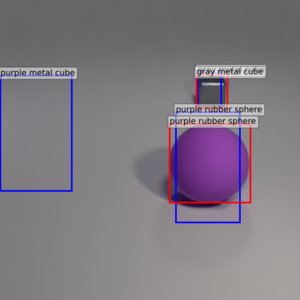}} \\
  
              {\method{} \\ (Ours)} &
              \adjustbox{valign=c}{\includegraphics[height=0.14\textwidth]{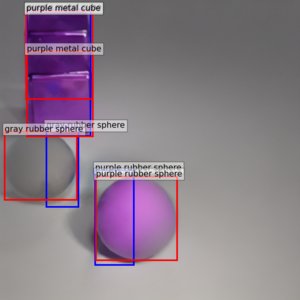}} &
              \adjustbox{valign=c}{\includegraphics[height=0.14\textwidth]{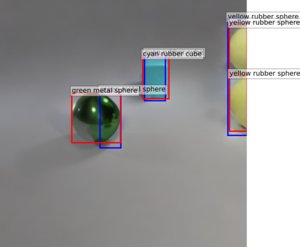}} &
              \adjustbox{valign=c}{\includegraphics[height=0.14\textwidth]{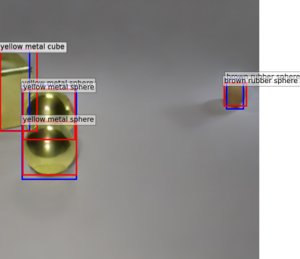}} &
              \adjustbox{valign=c}{\includegraphics[height=0.14\textwidth]{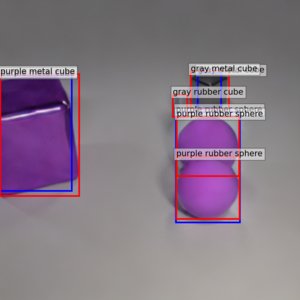}} \\
  
              \bottomrule
            \end{tblr}
        }
    \end{center}
    \caption{
    Additional \benchmark{} Shape-vertical task samples.
    Top 4 rows: Images with GT boxes (in \textcolor{blue}{blue}).
    Bottom 4 rows: Images with GT boxes (in \textcolor{blue}{blue}) and object detection results (in \textcolor{red}{red}).
    }
    \label{tab:additional_layoutbench_shape_vertical}
    \end{table*}

\end{document}